%% file: root-IEEE.tex
\documentclass[conference]{IEEEtran}
\IEEEoverridecommandlockouts    

\usepackage{multirow}
\usepackage{lipsum}
\usepackage{amsmath}
\usepackage{amssymb}
\usepackage[ruled,vlined]{algorithm2e}
\usepackage{graphicx}
\usepackage{xcolor}
\usepackage[utf8]{inputenc}
\usepackage{textcomp}
\graphicspath{{./Figures/}}

\usepackage{xspace}
\usepackage{xfrac}
\usepackage{pifont}
\usepackage{adjustbox}
\usepackage{float}
\usepackage{comment}
\usepackage{multicol}
\usepackage{textcomp}
\usepackage{algorithmic}
\usepackage{cite}

\usepackage{booktabs}
\usepackage{multirow}
\usepackage[table]{xcolor}
\usepackage{tabularx}
\usepackage{array}
\usepackage{makecell}
\usepackage{colortbl}
\definecolor{lightgray}{gray}{0.95}
\definecolor{lightgreen}{RGB}{220,250,220}
\definecolor{lightblue}{RGB}{220,235,250}
\definecolor{lightyellow}{RGB}{255,250,205}

\usepackage[colorlinks,pagebackref=true,citecolor=blue,bookmarks=false,hypertexnames=true]{hyperref} 

\title{\LARGE \bf Multi-Modal Traffic Sign Detection with Semantic Attributes \\ for Autonomous Driving}

\author{
	\parbox{\textwidth}{%
		\centering
		Meda Lazar$^{*}$$^{1}$, Sourab Sridhar$^{*}$$^{2}$, Shashwata Gupta$^{3}$, Alexandra Tripcea$^{1}$, Varun Ravi$^{4}$, Senthil Yogamani$^{4}$
	}%
    \thanks{$^{*}$Co-first authors with equal contribution}%
	\thanks{$^{1}$Arriver System Software S.r.l., Romania
		{\tt\small mlazar@qti.qualcomm.com}}%
	\thanks{$^{2}$Qualcomm Auto Ltd Sweden Filial 
		{\tt\small soursrid@qti.qualcomm.com}}%
	\thanks{$^{3}$Qualcomm Auto Ltd., UK
		{\tt\small shagup@qti.qualcomm.com}%
        }     
	\thanks{$^{4}$Automated Driving, Qualcomm Technologies, Inc
        }
}

\begin{document}
	
	\maketitle
	\thispagestyle{empty}
	\pagestyle{empty}
	
	\begin{abstract}
    Reliable traffic sign detection is a prerequisite for the global deployment of autonomous driving systems, where regulatory compliance and road safety depend on perceiving signs correctly across regions, ranges, and weather conditions. Despite recent progress, vision-based methods continue to face three fundamental limitations: poor cross-regional generalization due to high diversity across countries, degraded performance on small-object detection at long ranges (traffic signs occupy as little as $10\texorpdfstring{\times}{x}10$ pixels at 200\,m), and fragile temporal tracking under the strongly non-linear perspective distortion that occurs as a vehicle approaches a sign. In this paper, we address the problem of robust, long-range, region-agnostic traffic sign perception by combining camera and Light Detection and Ranging (LiDAR) sensing. We present a multi-modal detection framework whose Intensity-Aware Deformable Fusion module aligns retro-reflective LiDAR cues with camera features, anchoring detection on geometric invariants rather than region-specific visual appearance. We further introduce a dual motion-model tracker that explicitly accounts for non-linear perspective transformations during vehicle approach, substantially improving temporal consistency over linear motion assumptions. Additionally, we develop a semantic attribute classification pipeline that estimates occlusion level, readability, sign embeddedness, and road relevance, providing actionable context to downstream planning. Extensive evaluation on our dataset, spanning 60+ countries and 2{,}500+ hours of driving data, shows that the proposed pipeline achieves an Object Miss Ratio (OMR) of 0.49\% across 221{,}068 evaluation sequences, demonstrating globally generalizable traffic sign perception in commercial-grade autonomous driving systems.
    \end{abstract}

    \input{ieee_sections/sec_introduction}
    \input{ieee_sections/sec_related_work}
    \input{ieee_sections/sec_materials_and_methods}
    \input{ieee_sections/sec_detector_and_tracker}
    \input{ieee_sections/sec_semantic_attributes}
    \input{ieee_sections/sec_results}
    \input{ieee_sections/sec_conclusions}
	\bibliographystyle{IEEEtran}
	\bibliography{root} 
	
\end{document}

%% file: ieee_sections/sec_introduction.tex
\section{Introduction} 
\label{sec:introduction}
Traffic-sign detection is a critical component of modern Advanced Driver Assistance Systems (ADAS) and autonomous vehicles, ensuring regulatory compliance and road safety~\cite{joseph2021autonomous}. While core perception tasks like object detection~\cite{sistu2019neurall}, semantic segmentation~\cite{Chennupativisapp19}, and depth estimation~\cite{kumar2018near} have matured, traffic-sign detection remains uniquely challenging due to the requirement of high-fidelity classification across diverse sign categories at high detection ranges and difficult environmental conditions~\cite{wali2019vision}.

Recent research has prioritized optimizing vision-only architectures for real-time edge inference, yielding lightweight detectors and improved attention mechanisms while still leaving traffic-sign perception susceptible to adverse weather, motion blur, and varying spectral conditions. We defer a detailed discussion of these methods, public benchmarks, and prior multi-modal and tracking approaches to Section~\ref{sec:relatedwork}. In the rest of this section, we focus on the limitations that motivate our design and on the contributions of this paper.

\begin{figure}[!t]
    \centering
    \includegraphics[width=\columnwidth]{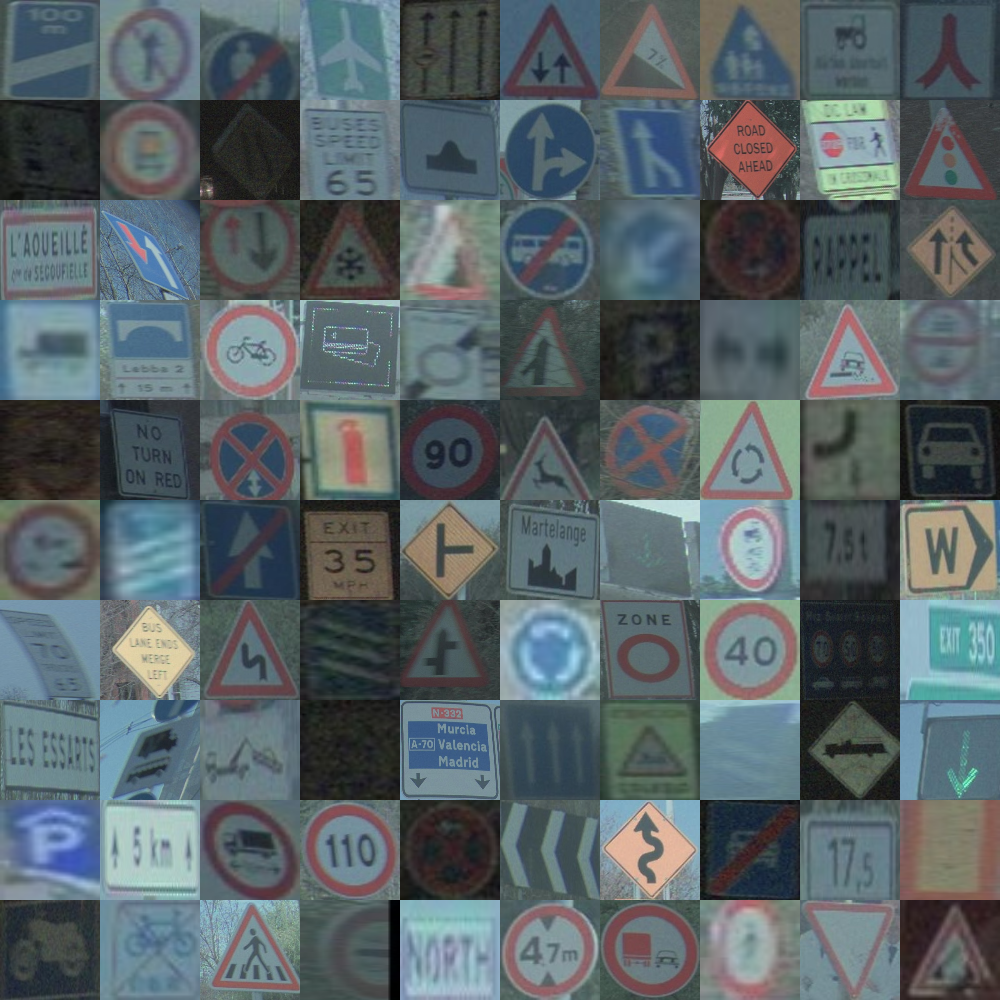}
    \caption{Traffic sign samples from the Qualcomm internal dataset. The samples highlight the geographic diversity and the wide range of regulatory standards captured by our dataset.}
    \label{fig:diverse_traffic_signs}
\end{figure}

Despite these strides, traffic-sign detection remains a comparatively underexplored task: multi-modal approaches are particularly scarce, public benchmarks are geographically narrow and fail to capture the full diversity of global regulatory standards (as illustrated in Figure ~\ref{fig:diverse_traffic_signs}), and the community lacks large-scale datasets that reflect real-world deployment conditions. These gaps manifest as three fundamental limitations that hinder global deployment of traffic-sign detection systems:

\begin{enumerate}
    \item \textbf{Geographic Generalization and Appearance Variability} (addressed in Sections~\ref{sec:materialsandmethods} and~\ref{sec:proposeddetectorandtracker}): Models trained on region-specific datasets often fail in unseen geographic regions with different regulatory standards~\cite{babic2021analysis}, and even large-scale image-only datasets are prone to appearance-based overfitting. While the visual appearance of signs varies globally, their physical structure, typically planar surfaces at standardized heights, is a geometric invariant. Camera-only systems are forced to rely on volatile visual signatures that differ across regions, whereas LiDAR captures this structural invariant directly through 3D point geometry. We address this through a large-scale, multi-continental dataset and a LiDAR-camera fusion strategy that anchors detection on geometry rather than region-specific visual appearance.

    \item \textbf{Sensor Fragility under Real-World Conditions} (addressed in Section~\ref{sec:proposeddetectorandtracker}): Camera-only systems exhibit three coupled failure modes that limit their deployability. First, monocular vision suffers from an inherent mathematical ambiguity between an object's physical size and its distance from the camera, which cannot be resolved from a single image alone. LiDAR directly measures metric depth per point and thus eliminates this scale uncertainty. Second, passive cameras degrade severely under adverse weather, glare, adversarial attacks, and near-total darkness \cite{yahiaoui2019overview, sobh2021adversarial}. In contrast, LiDAR, as an active sensor, is immune to ambient lighting conditions, and traffic signs are manufactured with retro-reflective coatings that produce distinctively high LiDAR intensity returns. This provides a modality-specific cue that remains robust precisely when cameras degrade. Third, reliance on 2D image features leads to failures at extreme ranges: at distances of 200\,m, visual features are often lost due to neural network downsampling, and the resulting pixel footprint is insufficient for standard convolutional kernels to extract discriminative features. LiDAR point clouds, being resolution-independent, preserve geometric structure at long range and provide complementary spatial hypotheses that guide camera-based detectors toward low-pixel-count targets.

    \item \textbf{Temporal Instability in Tracking} (addressed in Section~\ref{sec:proposeddetectorandtracker}): Standard tracking algorithms often assume linear motion, failing to account for the non-linear perspective distortions and rapid scale changes that occur as a vehicle approaches a sign. We address this with a dual motion model that explicitly handles both linear and non-linear regimes, improving temporal stability and reducing track fragmentation.
\end{enumerate}

Although multi-modal sensor fusion has matured for general autonomous-driving perception (see Section~\ref{sec:relatedwork}), its application to traffic-sign detection remains comparatively unexplored, in part because public datasets that combine geographic diversity, long-range annotations, and synchronized 2D/3D labels are scarce. We therefore validate our approach on a large-scale Qualcomm internal dataset collected from a fleet of data collection vehicles across multiple continents (Table~\ref{tab:dataset_comparison}).

Our contributions are summarized as follows:
\begin{itemize}
   \item \textbf{Geometrically Robust Multi-modal Detection}: We present a novel detection pipeline that integrates LiDAR geometric depth and reflectance with camera features. This fusion exploits physical consistency to achieve region-invariant performance and extends the effective detection range to 200\,m.
   \item \textbf{Dual Motion Model for Consistent Tracking}: We propose a tracking algorithm that explicitly models non-linear perspective transformations. This substantially improves temporal stability and reduces track fragmentation compared to standard linear motion models.
   \item \textbf{Semantic Attribute Classification}: We develop a comprehensive pipeline to assess occlusion, readability, relevance, and embeddedness. This allows the stack to filter non-actionable detections, improving the quality of downstream planning. 
   \item \textbf{Large-Scale Validation}: We demonstrate the efficacy of our pipeline on a large-scale Qualcomm internal dataset, benchmarking its performance against manual ground-truth annotations across diverse global regions, achieving an Object Miss Ratio of 0.49\% across 2{,}500+ hours of driving data.
\end{itemize}

The remainder of this paper is organized as follows. Section~\ref{sec:relatedwork} reviews related work in vision-based traffic-sign detection, LiDAR-camera fusion, and multi-object tracking, and identifies the gap that motivates our design. Section~\ref{sec:materialsandmethods} describes the materials and methods, including the large-scale Qualcomm internal dataset, sensor specifications, multi-modal data pre-processing, and the experimental setup. Section~\ref{sec:proposeddetectorandtracker} presents the proposed technical framework, comprising the multi-modal detector, the dual motion-model tracker, and the intensity-aware deformable fusion module. Section~\ref{sec:semanticattributeclassification} introduces a semantic attribute classification pipeline for estimating occlusion, readability, embeddedness, and relevance. Section~\ref{sec:results} reports a comprehensive evaluation of the proposed system, and Section~\ref{sec:conclusions} concludes the paper.

%% file: ieee_sections/sec_related_work.tex
\section{Related Work}
\label{sec:relatedwork}

\subsection{Vision-Based Traffic Sign Detection and Benchmarks}
The dominant paradigm for traffic-sign detection is single-image, camera-only inference, and the field has matured around increasingly efficient real-time detectors. Recent work has emphasized lightweight architectures that target edge deployment: Wang~et~al.~\cite{wang2023improved} extend YOLOv5 with multi-scale enhancements for traffic-sign-specific scales, while Zhang~et~al.~\cite{zhang2025lr} introduce LR-DETR as a lightweight transformer-based detector targeting real-time performance. Other work has focused on small-object handling, either through specialized attention mechanisms~\cite{zhang2024learning} or vehicle-mounted detectors that explicitly target small-sized signs across operating conditions~\cite{wang2023vehicle}; recent surveys~\cite{wali2019vision, suresha2024recent, zhu2022traffic} document the resulting body of work and its persistent failure modes under adverse weather, motion blur, and varying spectral conditions~\cite{temel2019traffic, michaelis2020benchmarkingrobustnessobjectdetection}.

The development of these methods has been shaped, and constrained, by the available benchmarks. GTSDB~\cite{Stallkamp-IJCNN-2011}, TT100K~\cite{Zhe_2016_CVPR}, and BelgiumTS~\cite{Timofte-WACV-2009} each cover a single country or region; broader datasets such as BDD100K~\cite{bdd100k} and Mapillary~\cite{ertler2020mapillarytrafficsigndataset} expand geographic coverage but remain image-only. The Zenseact Open Dataset~\cite{alibeigi2023zenseact} is, to our knowledge, the only public driving benchmark that pairs synchronized 2D and 3D annotations with multi-modal sensing for traffic signs, but its geographic coverage is limited to 14 European countries and its sign annotation density is two orders of magnitude smaller than what we use in this work (Section~\ref{sec:materialsandmethods}). Vision-only multi-modal extensions such as CLFNet~\cite{liu2024clfnet} fuse multiple visual feature streams but do not exploit physical sensing modalities beyond the camera.

\subsection{LiDAR-Camera Fusion for Object Detection}
Multi-modal fusion of camera and LiDAR has matured rapidly for general 3D object detection in autonomous driving, and recent surveys~\cite{9000872} catalog the design space. Three lines of work are particularly relevant to ours. PointPainting~\cite{vora2020pointpaintingsequentialfusion3d} pioneered the projection-based fusion paradigm by decorating LiDAR points with semantic features from a camera segmentation network, demonstrating that even simple cross-modal cues materially improve detection. TransFusion~\cite{bai2022transfusionrobustlidarcamerafusion} addresses the robustness of this projection by introducing a transformer-based fusion that is more tolerant to camera-LiDAR misalignment than rigid concatenation. BEVFusion~\cite{liu2024bevfusionmultitaskmultisensorfusion} unifies the two modalities in a shared bird's-eye-view representation, providing a strong reference point for multi-task multi-sensor perception. The DAL framework~\cite{huang2023detectinglabelingrethinkinglidarcamera} reframes 3D fusion as a labeling problem and is the architecture from which our initial 3D-centric experiments were derived (Section~\ref{sec:proposeddetectorandtracker}). In a different sensing combination, BEVCar~\cite{schramm2024bevcar} demonstrates the value of camera-radar fusion for segmentation-style tasks. Despite this progress, almost all of these methods are evaluated on dense object classes (vehicles, pedestrians, cyclists), and their behavior on extreme small-object targets such as traffic signs at 200\,m, where each LiDAR return carries disproportionate weight, has received comparatively little attention.

\subsection{Tracking-by-Detection with Non-Linear Motion}
The tracking-by-detection paradigm, established by SORT~\cite{7533003} and extended with appearance features by DeepSORT~\cite{8296962}, remains the dominant approach for online multi-object tracking. Subsequent work has improved either the data association step, as in OC-SORT~\cite{cao2023observationcentricsortrethinkingsort} which leverages observation-centric reasoning to recover from short occlusions, or the use of low-confidence detections, as in ByteTrack~\cite{DBLP:journals/corr/abs-2110-06864} and StrongSORT~\cite{du2023strongsortmakedeepsortgreat} which trade off precision and recall through richer matching cascades. The BoostTrack family~\cite{stanojevic2024boosttrack, stanojevic2024boosttrack++} further refines similarity measures and tracklet-level reasoning. A common assumption shared across these methods is that target motion in the image plane is well approximated by a constant-velocity Kalman filter, with corrections applied at the association level. This assumption is reasonable for pedestrians, vehicles, and cyclists, whose image-plane trajectories are dominated by their own motion. It is materially less appropriate for traffic signs, which are stationary in 3D but exhibit strongly non-linear scale and position dynamics in the image plane as the ego-vehicle approaches them, especially in the final tens of meters before passing.

\subsection{Key Differences}
Our work differs from each of the threads above along a specific axis. Relative to vision-only TSR~\cite{wang2023improved, zhang2025lr, zhang2024learning, wang2023vehicle}, we anchor detection on a sensing modality (LiDAR) whose output is invariant to regional appearance variation and remains informative under conditions where cameras degrade. Relative to public benchmarks~\cite{Stallkamp-IJCNN-2011, Zhe_2016_CVPR, bdd100k, ertler2020mapillarytrafficsigndataset, alibeigi2023zenseact}, we evaluate on a substantially larger and more geographically diverse dataset that includes synchronized 2D and 3D annotations out to 200\,m. Relative to general-purpose LiDAR-camera fusion~\cite{vora2020pointpaintingsequentialfusion3d, bai2022transfusionrobustlidarcamerafusion, liu2024bevfusionmultitaskmultisensorfusion, huang2023detectinglabelingrethinkinglidarcamera}, we adopt a 2D-centric fusion strategy specifically chosen for its tolerance to long-range cross-sensor misalignment, and we exploit the retro-reflective signature of traffic signs as a class-specific prior through our Intensity-Aware Deformable Fusion module. Relative to constant-velocity trackers~\cite{7533003, 8296962, cao2023observationcentricsortrethinkingsort, DBLP:journals/corr/abs-2110-06864, du2023strongsortmakedeepsortgreat, stanojevic2024boosttrack, stanojevic2024boosttrack++}, we replace the single-motion-model assumption with a parallel dual motion model (constant-acceleration and constant-jerk) that captures the higher-order perspective dynamics specific to vehicle approach. Finally, beyond detection and tracking, we introduce a semantic attribute classification stage (occlusion, readability, embeddedness, relevance) that, to the best of our knowledge, is not jointly modeled in any prior public traffic sign perception system.

%% file: ieee_sections/sec_materials_and_methods.tex
\section{Materials and Methods}
\label{sec:materialsandmethods}

\subsection{Dataset Description and Comparison}
To validate the proposed method's efficacy in real-world autonomous driving scenarios, we perform a multi-stage evaluation leveraging a large-scale, proprietary Qualcomm dataset. While existing academic benchmarks provide foundational metrics, they often lack the geographic diversity and multi-modal primitives required for robust global deployment \cite{uricar2019challenges}.

\begin{figure*}[t]
    \centering
    \includegraphics[width=0.8\textwidth]{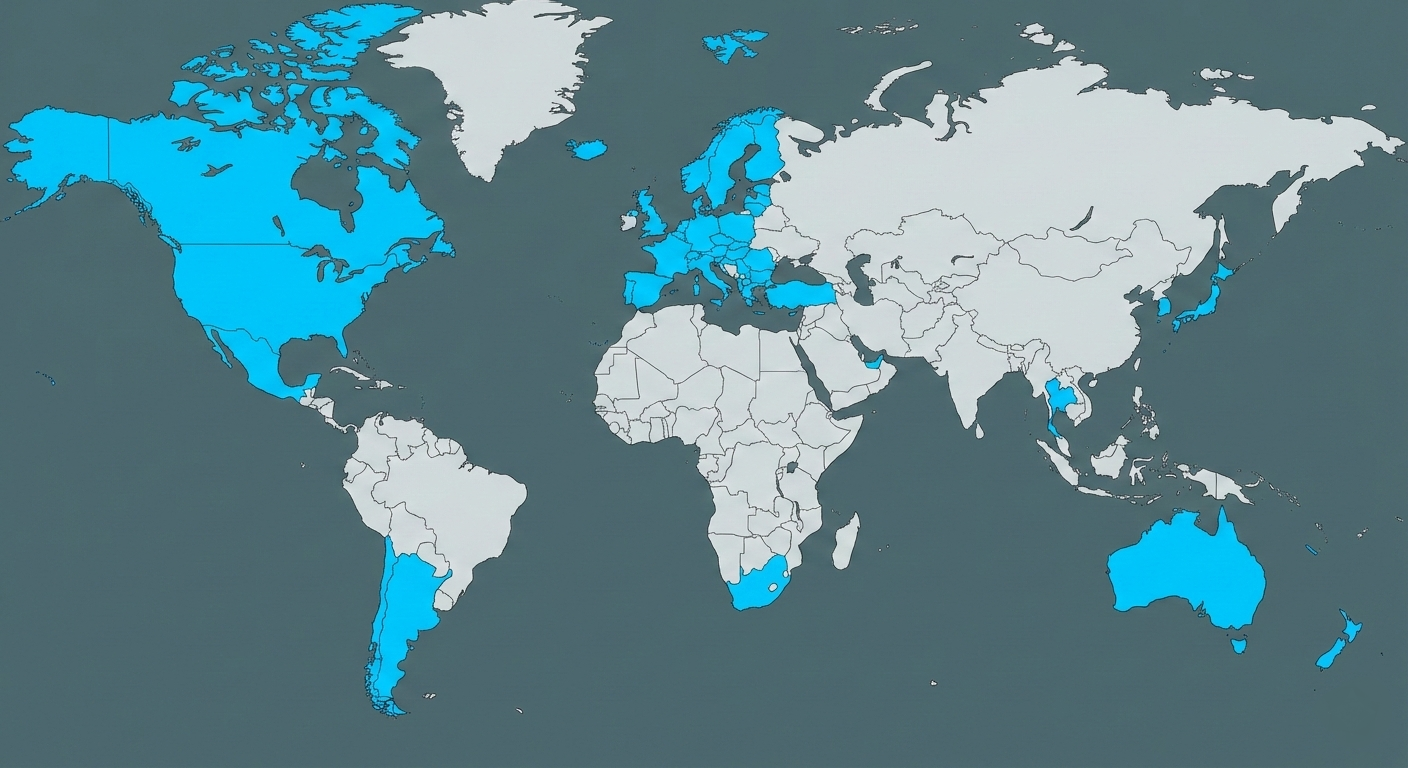}
    \caption{Global geographic coverage of the Qualcomm dataset used in our development.}
    \label{fig:countries}
\end{figure*}

\begin{table*}[!t]
\caption{Overview of traffic sign datasets and comparative attributes. ZOD: Zenseact Open Dataset; MTSD: Mapillary Traffic Sign Dataset; TT100K: Tsinghua-Tencent 100K; GTSDB: German Traffic-Sign Detection Benchmark.}
\label{tab:dataset_comparison}
\centering
\small
\resizebox{0.8\linewidth}{!}
{
\begin{tabular}{lcccc}
\toprule
\rowcolor{lightgreen}
\textbf{Dataset} & \textbf{Annotated Objects} & \textbf{Annotation Type} & \textbf{Region} & \textbf{Modality} \\
\midrule
Qualcomm & 142{,}000{,}000 & 2D/3D Bounding Box & 60+ Countries & Image + LiDAR \\
Zenseact (ZOD) \cite{alibeigi2023zenseact} & 446{,}000 & 2D/3D Bounding Box & 14 (Europe) & Image + LiDAR \\
Mapillary (MTSD) \cite{ertler2020mapillarytrafficsigndataset} & 300{,}000 & 2D Bounding Box & 60 Countries & Image Only \\
TT100K \cite{Zhe_2016_CVPR} & 26{,}000 & 2D Bounding Box & China & Image Only \\
GTSDB \cite{Stallkamp-IJCNN-2011} & 852 & 2D Bounding Box & Germany & Image Only \\
\bottomrule
\end{tabular}
}
\end{table*}

\subsubsection{The Qualcomm Dataset: Scale and Advantages}
The Qualcomm dataset represents a substantial increase in data volume compared to traditional benchmarks. Containing over 142 million annotated objects, it is several orders of magnitude larger than existing open-source alternatives.
This scale is critical for addressing the long-tail distribution of traffic-sign detection~\cite{Sun_2017_ICCV}, ensuring that models are exposed to rare edge cases, adverse weather conditions, and diverse illumination profiles that are statistically underrepresented in smaller datasets~\cite{6335478}. The country distribution is presented in Figure \ref{fig:countries}. Countries where data is available in our dataset are shown in blue.

The decision to utilize this internal dataset over external benchmarks is driven by three primary factors:
\begin{itemize}
    \item \textbf{Global Geographic Breadth:} The data spans 60+ countries across multiple continents, facilitating the development of region-invariant features that generalize across different international signage standards (see Figure ~\ref{fig:countries}).
    \item \textbf{Multi-Modal High-Fidelity Annotations:} The Qualcomm dataset provides synchronized 2D and 3D bounding boxes, which are unavailable in vision-only benchmarks such as Mapillary, TT100K, and GTSDB.
    \item \textbf{Extended Detection Horizon:} The dataset includes high-fidelity 3D annotations for objects at ranges up to 200\,m, allowing for supervised long-range depth estimation and localization.
\end{itemize}

A comprehensive comparison of the Qualcomm dataset against common academic benchmarks is provided in Table~\ref{tab:dataset_comparison}.

\subsection{Evaluation Process and Metrics}
The evaluation of the proposed detector follows the standard COCO protocol to ensure comparability and rigor. The performance of the detector, tracker, and semantic attribute classification is assessed using the following metrics.

\subsubsection{Detection Metrics}
We utilize the Average Precision (AP) metric as defined by the Microsoft COCO benchmark~\cite{lin2015microsoftcococommonobjects}. AP is calculated by averaging results over multiple Intersection over Union (IoU) thresholds:
$\text{IoU} \in \{0.50, 0.55, \dots, 0.95\}$ with a step size of $0.05$.
Additionally, we report:
\begin{itemize}
    \item \textbf{AP$_{50}$ and AP$_{75}$:} Precision at specific IoU thresholds of $0.50$ and $0.75$.
    \item \textbf{Scale-Specific AP:} $\text{AP}_s$, $\text{AP}_m$, and $\text{AP}_l$ to analyze performance across small ($<32^2$ px), medium ($32^2$--$96^2$ px), and large ($>96^2$ px) object scales.
\end{itemize}

\subsubsection{Tracking and Attribute Metrics}
The tracker and semantic attributes are assessed via standard statistical measures to determine the reliability of the system's temporal and classification heads:
\begin{equation}
\text{Precision} = \frac{TP}{TP + FP}
\end{equation}
\begin{equation}
\text{Recall} = \frac{TP}{TP + FN}
\end{equation}
\begin{equation}
\text{Accuracy} = \frac{TP + TN}{TP + FP + TN + FN}
\end{equation}
where $TP$, $TN$, $FP$, and $FN$ represent True Positives, True Negatives, False Positives, and False Negatives, respectively.

\subsubsection{End-to-End Efficacy}
To assess performance against human-level ground truth, we define the Object Miss Ratio (OMR), representing the ratio of undetected signs relative to total manual annotations:
\begin{equation}
\text{Object Miss Ratio} = \frac{FN}{TP + FN}.
\end{equation}
The metric takes values in $[0\%, 100\%]$, where $0\%$ corresponds to no missed objects relative to the manually verified set and $100\%$ corresponds to a complete miss; lower values are therefore better.

%% file: ieee_sections/sec_detector_and_tracker.tex
\section{Proposed Multi-Modal Detection and Tracking}
\label{sec:proposeddetectorandtracker}

The task of long-range traffic-sign detection requires a balance between high-resolution semantic features and precise geometric localization. Our initial research explored a 3D-centric detection paradigm; however, empirical results revealed significant scalability and reliability issues, leading to the development of our current 2D multi-modal framework.

\subsection{Initial 3D Detection Approach and Limitations}
Initially, we sought to leverage a 3D detection framework to exploit the spatial depth of LiDAR data directly. We employed the DAL-Large~\cite{huang2023detectinglabelingrethinkinglidarcamera} architecture, utilizing a ResNet-50~\cite{he2015deepresiduallearningimage} camera encoder and Lk3D~\cite{chen2023largekernel3dscalingkernels3d} as the LiDAR encoder. The model was trained on the Zenseact dataset for 24 epochs using 10-frame temporal point cloud sweeps to increase density for objects beyond 50\,m. While this approach provided accurate localization for large obstacles (e.g., vehicles and pedestrians), it failed to scale for traffic-sign detection due to two primary factors:
\begin{itemize}
    \item \textbf{Point Density vs. Object Scale:} Traffic signs are physically small. At ranges exceeding 100\,m, even a multi-sweep point cloud provides only a handful of returns on a sign surface, making it nearly impossible for a 3D backbone to regress a stable 3D bounding box without significant visual support.
    \item \textbf{Sensitivity to Calibration and Synchronization:} 3D detectors rely on a rigid spatial correspondence between the point cloud and the image. In real-world deployment on the Qualcomm dataset, we observed that sub-degree extrinsic calibration drift or millisecond-level sensor desynchronization caused the 3D LiDAR proposals to ``detach'' from their visual counterparts.
\end{itemize}

When the sensor streams are not perfectly synchronized, the ego-vehicle's motion creates a spatial shear between the depth map and the RGB pixels. For 3D detectors, this misalignment results in the fusion of ``empty'' 3D space with sign imagery, leading to a substantial drop in precision. These findings necessitated a transition to a 2D multi-modal detector that utilizes LiDAR as a supplemental feature rather than a rigid geometric constraint. From the driving application perspective, a 2D detector augmented with a depth attribute obtained directly from LiDAR is sufficient for all our use cases.

\begin{figure}[!t]
    \centering
    \includegraphics[width=\columnwidth]{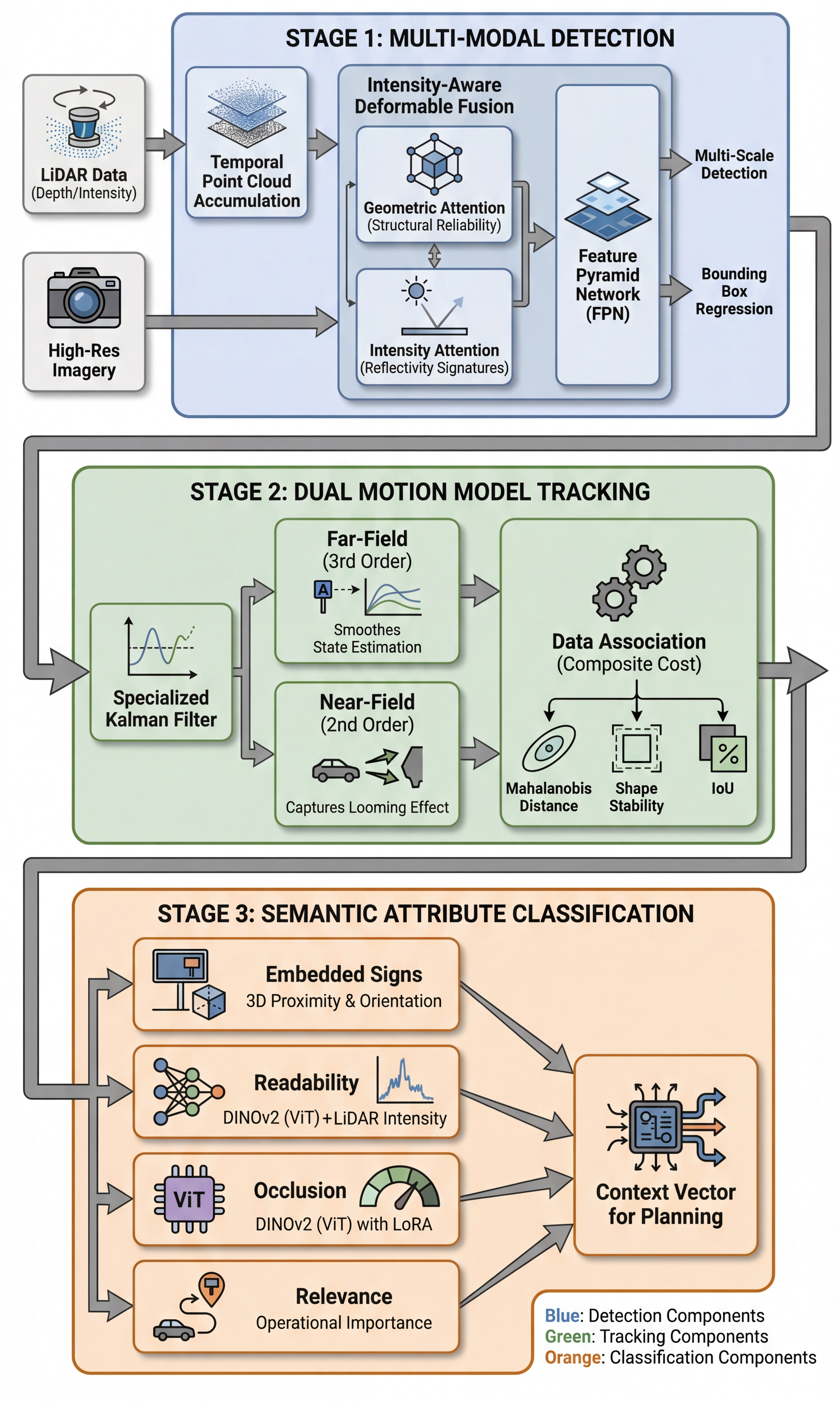}
    \caption{\textbf{Proposed architecture.} The pipeline has three stages. (i) A multi-modal detector fuses image features with depth and intensity maps, generated from a temporally accumulated LiDAR point cloud, through an Intensity-Aware Deformable Fusion module, and produces detections via a multi-scale detection head. (ii) Detections are linked into tracked bounding boxes by a dual motion-model tracker. (iii) Tracked boxes are post-processed and passed through an attribute classifier to produce semantic attributes. Architectural details are provided in Section~\ref{sec:proposeddetectorandtracker}.}
    \label{fig:architecture}
\end{figure}

\subsection{Proposed 2D Multi-Modal Detection Framework}
Our proposed architecture integrates LiDAR depth and intensity with high-resolution imagery to overcome the limitations of vision-only models at extreme ranges, where small pixel footprints and degraded visual features hinder detection~\cite{wang2023improved, zhang2025lr}. Unlike 3D-first models, our approach uses a 2D-centric fusion strategy that is more resilient to the alignment failures discussed above. The architecture of our proposed method is presented in Figure \ref{fig:architecture}.

\subsubsection{Multi-Modal Depth and Intensity Map Generation}
To address LiDAR sparsity, we employ a physically grounded temporal point cloud accumulation. We aggregate sequential frames into the current camera coordinate system using ego-vehicle odometry. This increases point density, which is critical for detecting small signs at 200\,m where visual pixels are degraded. The accumulated points are projected onto the image plane to generate dense depth and intensity maps. Median filtering is applied to depth maps for noise suppression, while max-pooling is used on intensity maps to preserve peak retro-reflectivity, providing a geometric ``anchor'' for the visual head.

\subsubsection{Intensity-Aware Deformable Fusion}
Traditional fusion methods often utilize rigid concatenation, which fails under calibration drift. We introduce \textit{Intensity-Aware Deformable Fusion} to achieve dynamic alignment between modalities. The core novelty lies in intensity-constrained offset prediction. Unlike standard deformable convolutions that predict sampling offsets solely from visual features, our method guides the sampling process using a joint geometric-intensity representation.

The network learns to ``pull'' RGB features toward high-reflectivity regions, creating a self-correcting alignment mechanism. To further refine this fusion, we employ a dual attention mechanism:
\begin{itemize}
    \item \textbf{Geometric Attention:} Assesses the structural reliability of LiDAR data to verify planar consistency.
    \item \textbf{Intensity Attention:} Emphasizes the specific reflective signatures of traffic signs.
\end{itemize}
By cross-verifying these streams, the system ensures that high weights are only assigned to features that are both structurally consistent and semantically relevant, preventing distractions from non-sign reflective objects such as metallic barriers~\cite{zhu2022traffic}.

\subsubsection{Backbone and Multi-Scale Detection Head}
\begin{figure*}[ht]
\centering
\includegraphics[width=.32\textwidth]{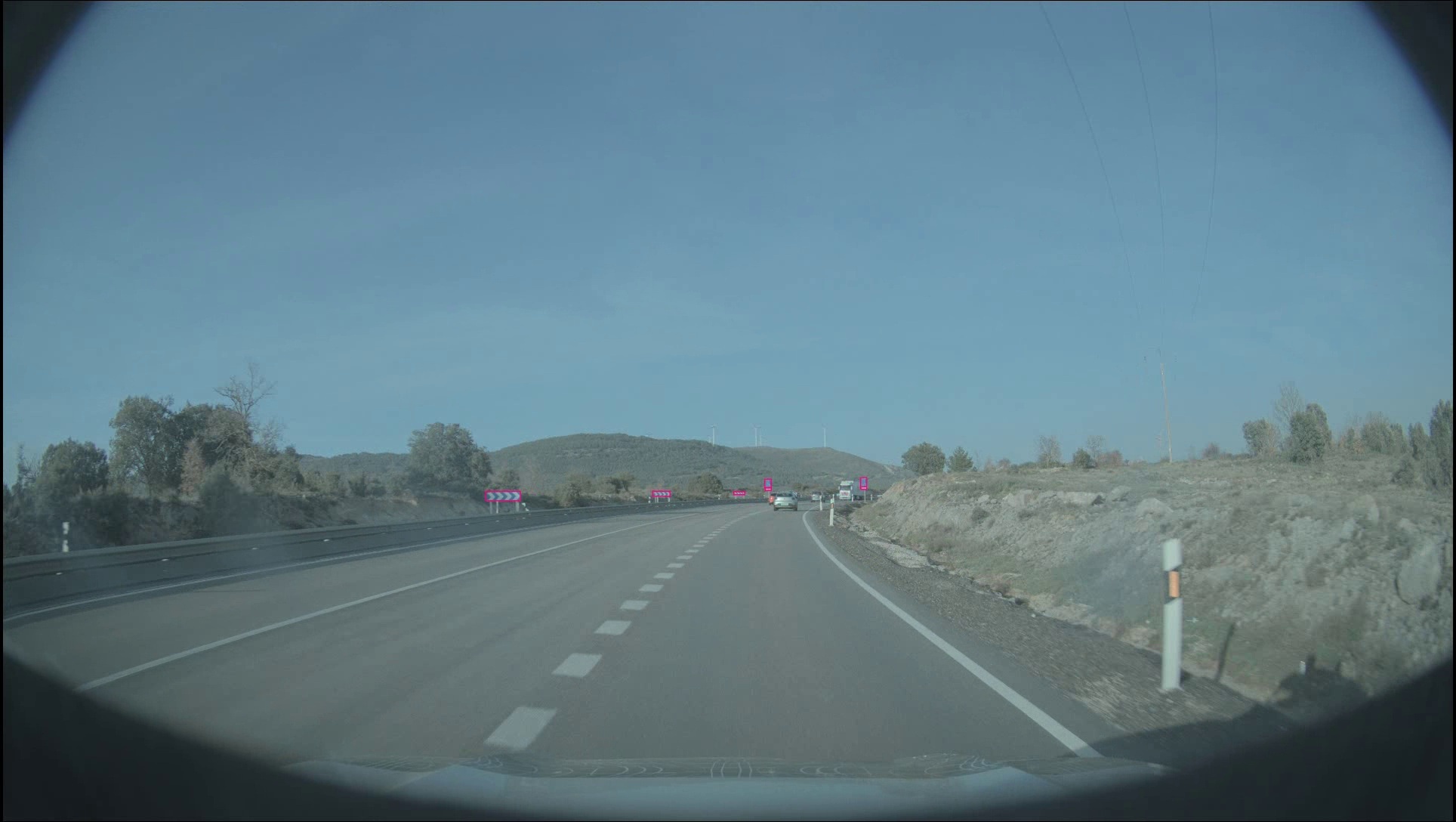}
\includegraphics[width=.32\textwidth]{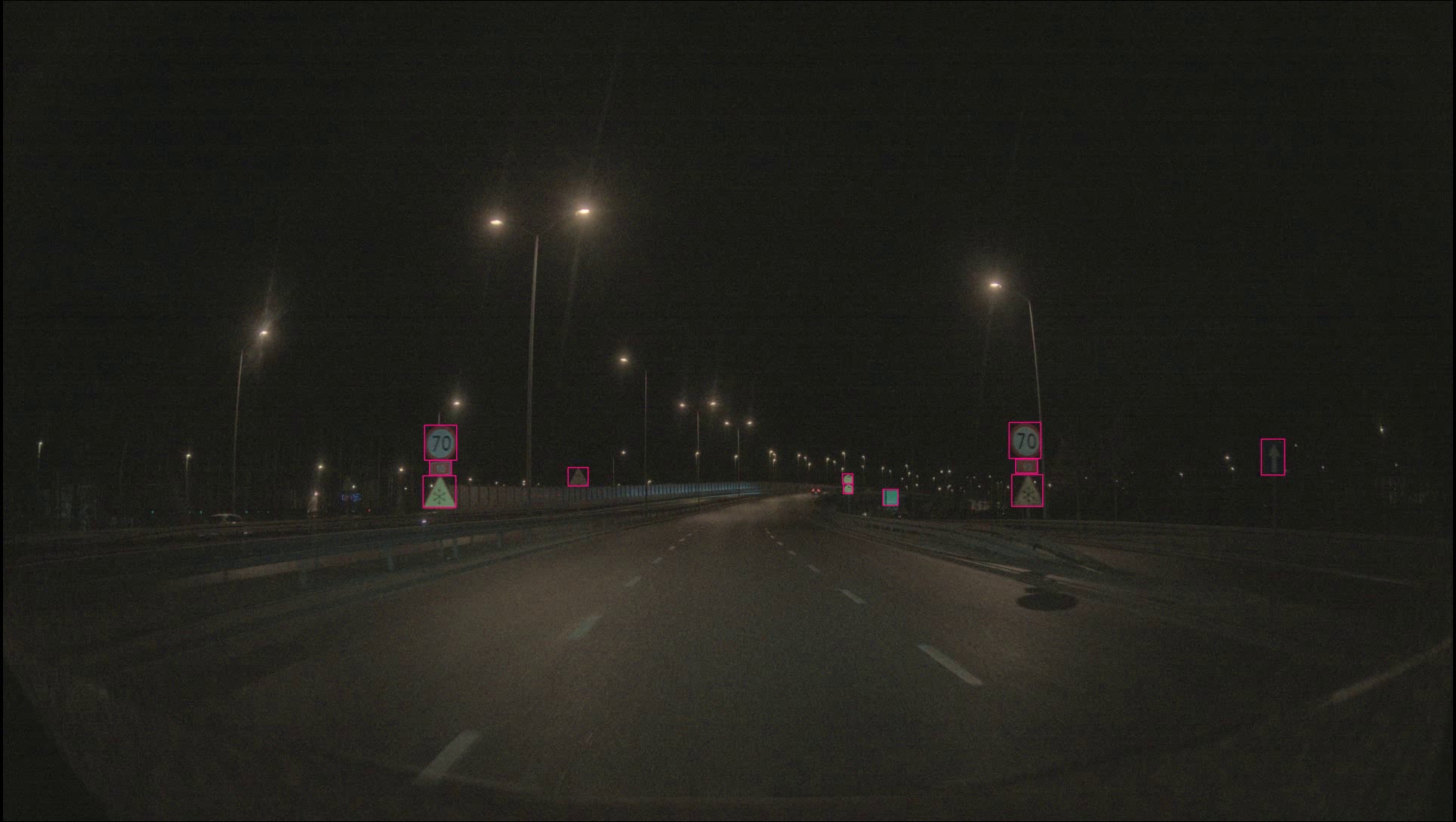}
\includegraphics[width=.32\textwidth]{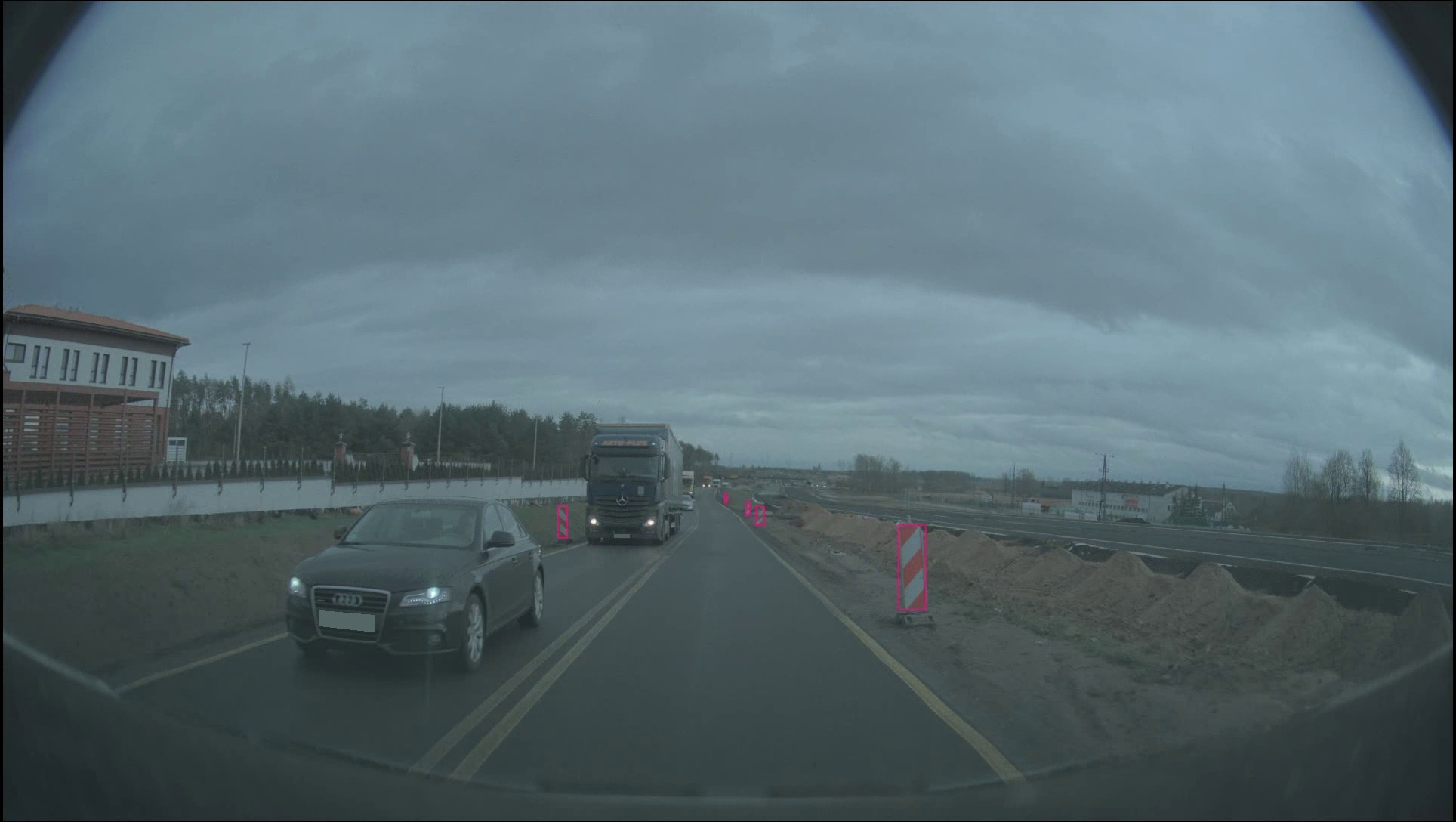}
\\[\smallskipamount]
\includegraphics[width=.32\textwidth]{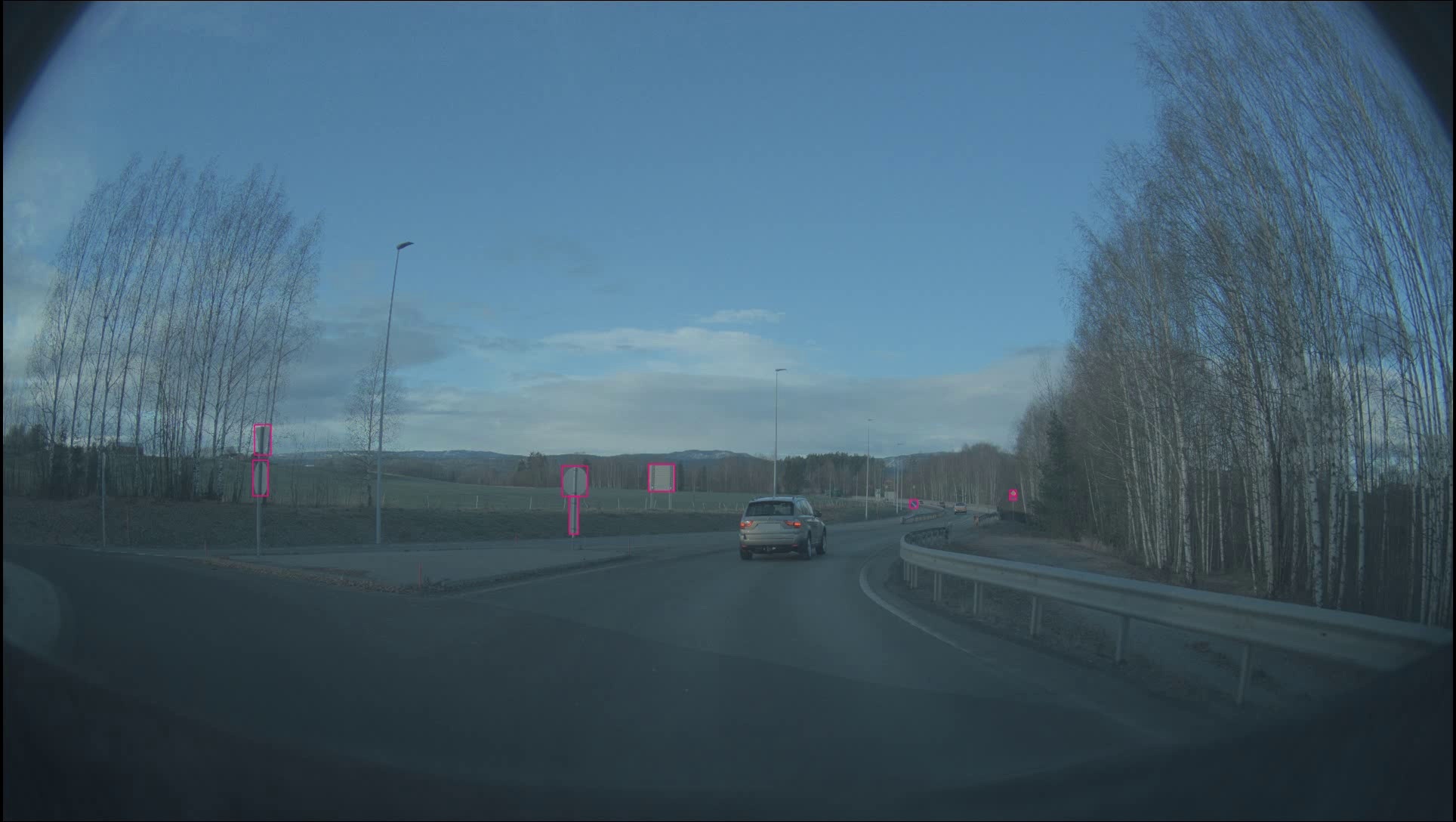}
\includegraphics[width=.32\textwidth]{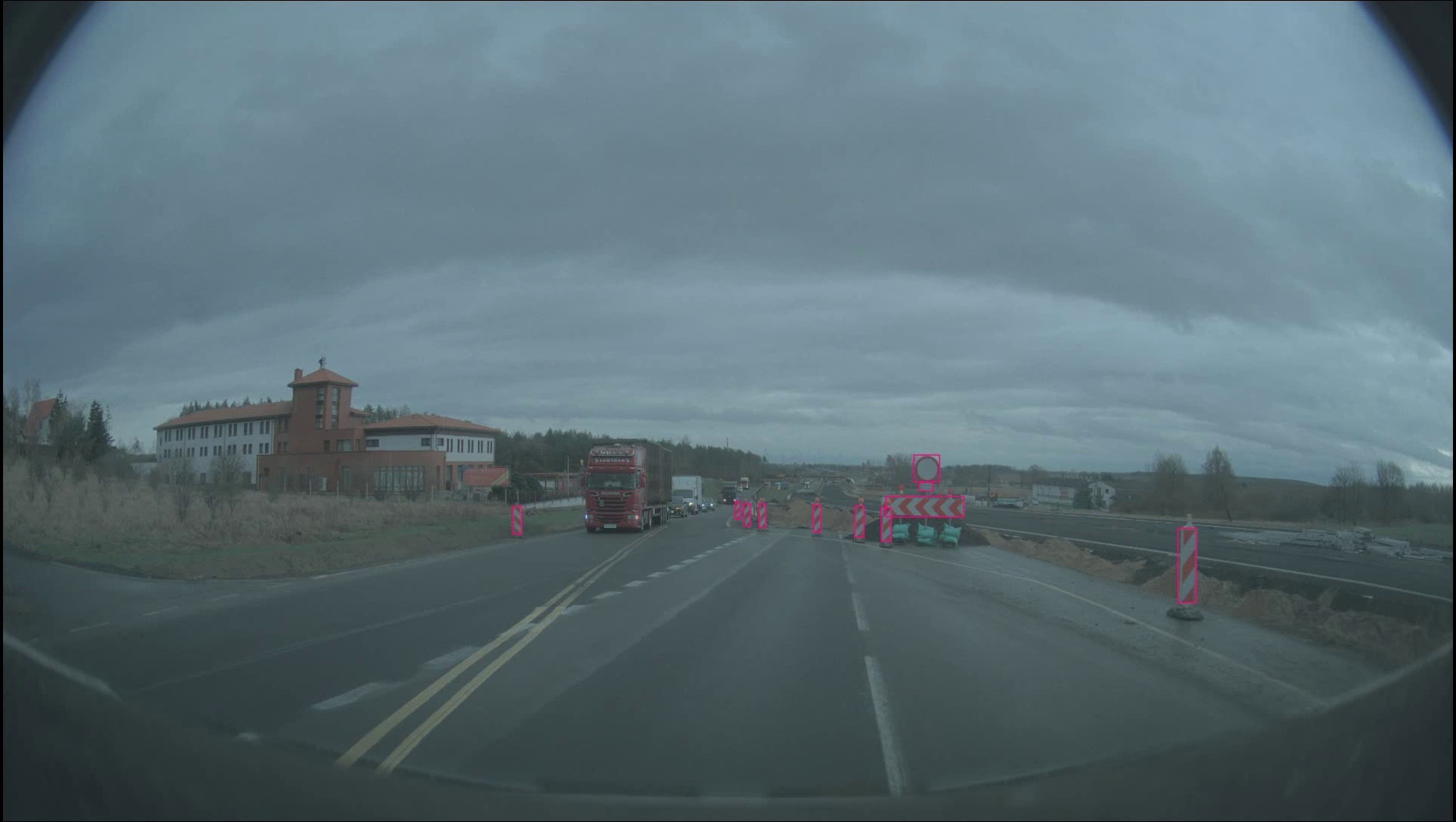}
\includegraphics[width=.32\textwidth]{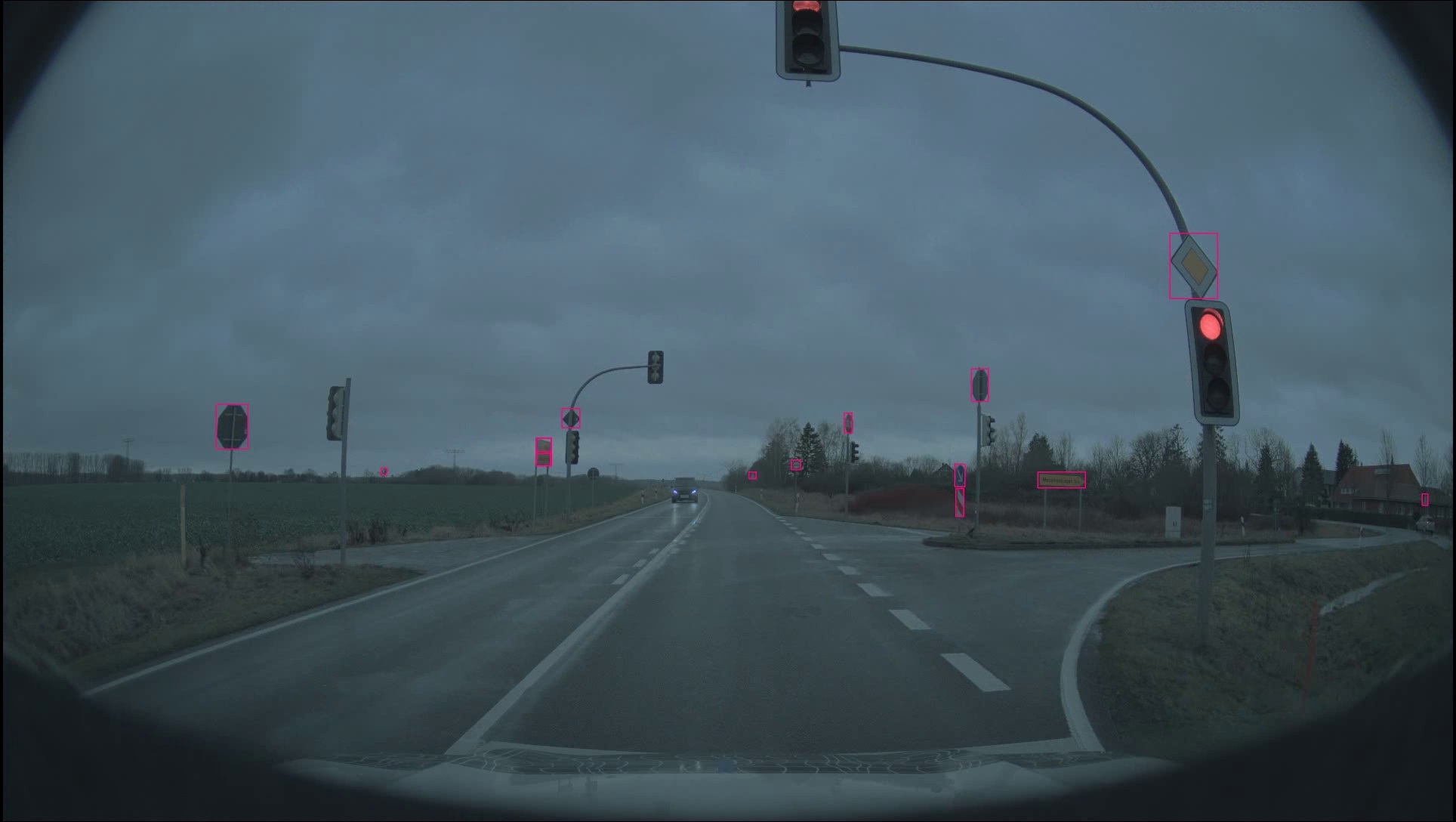}
\\[\smallskipamount]
\includegraphics[width=.32\textwidth]{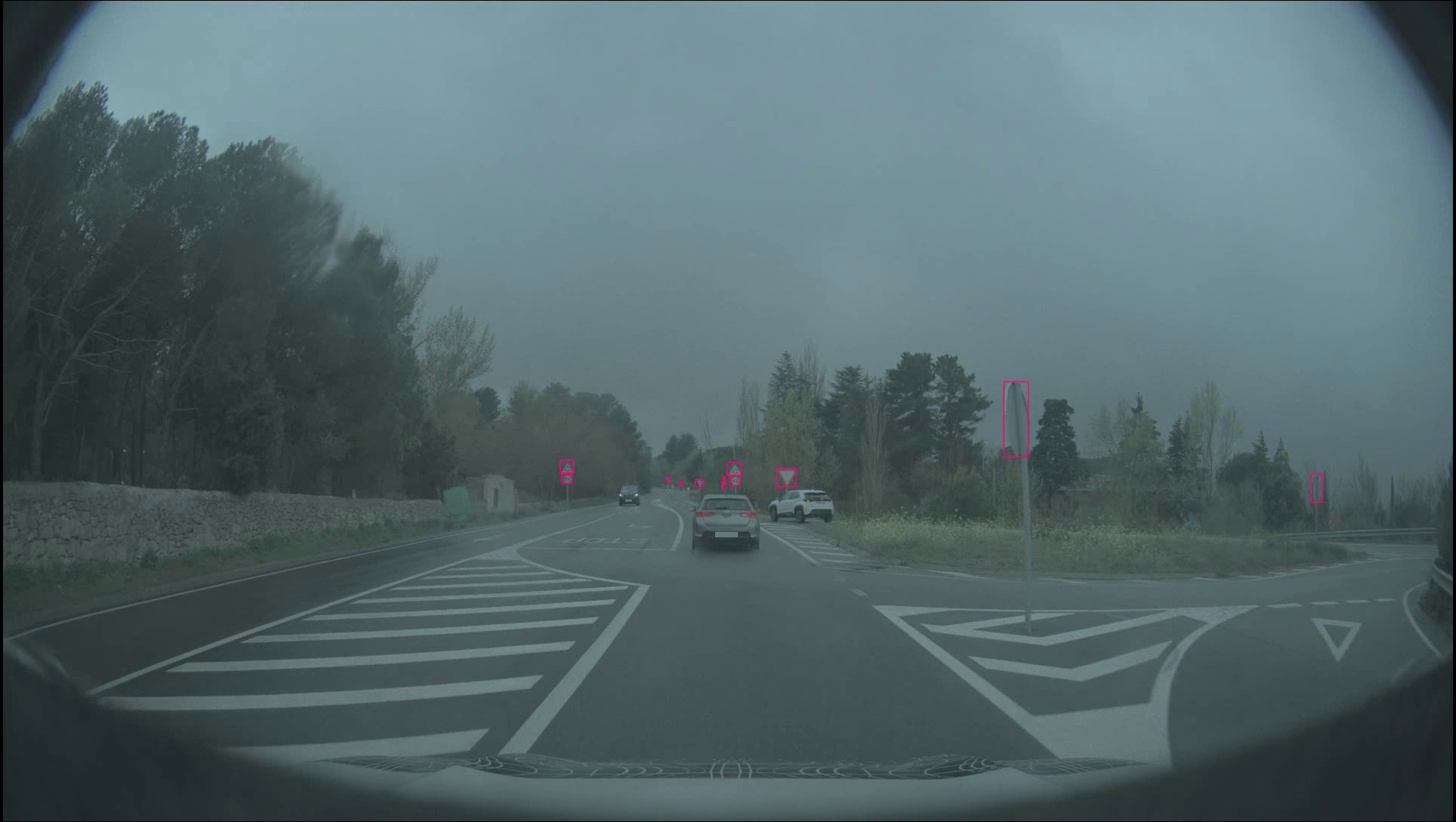}
\includegraphics[width=.32\textwidth]{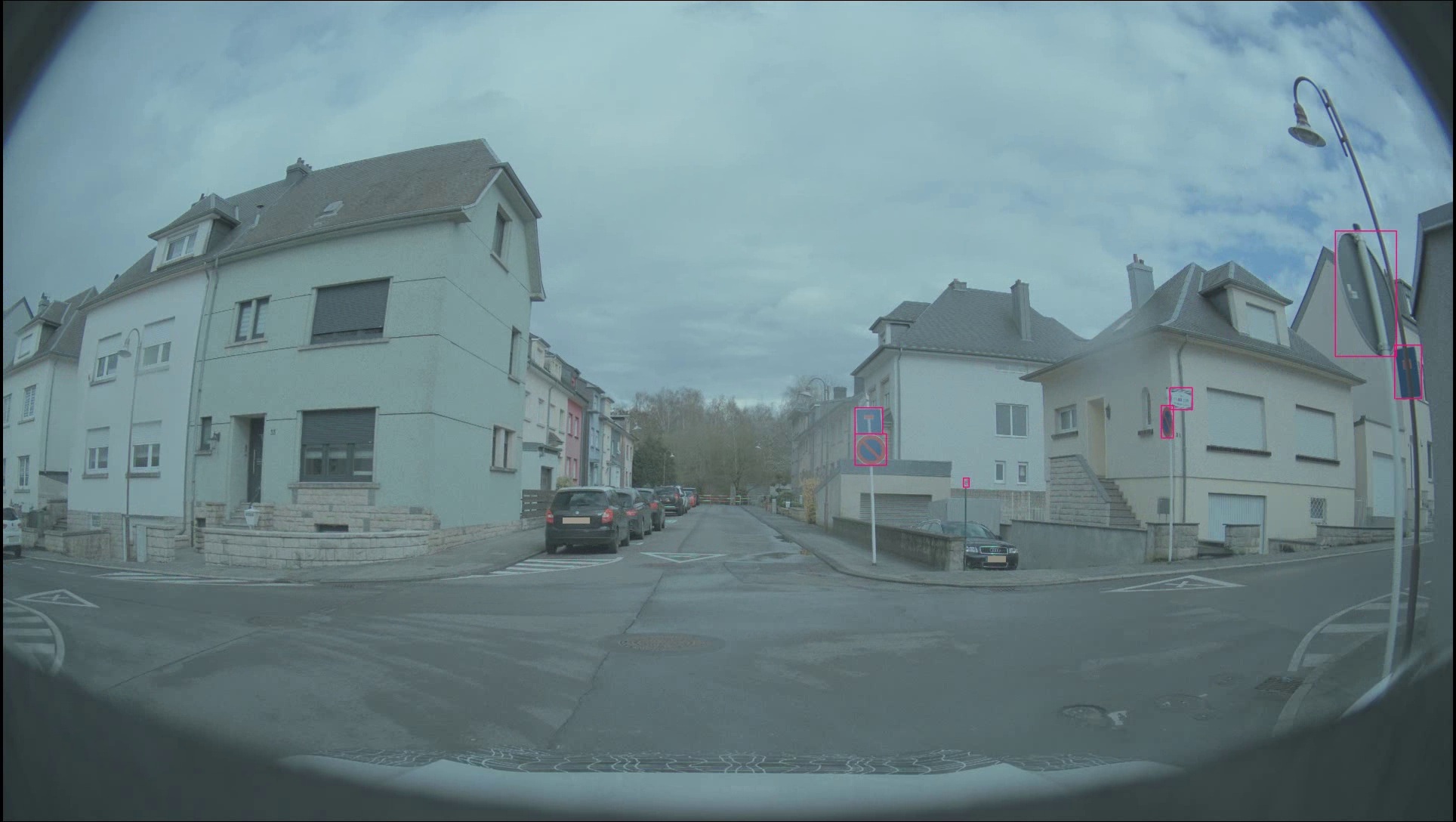}
\includegraphics[width=.32\textwidth]{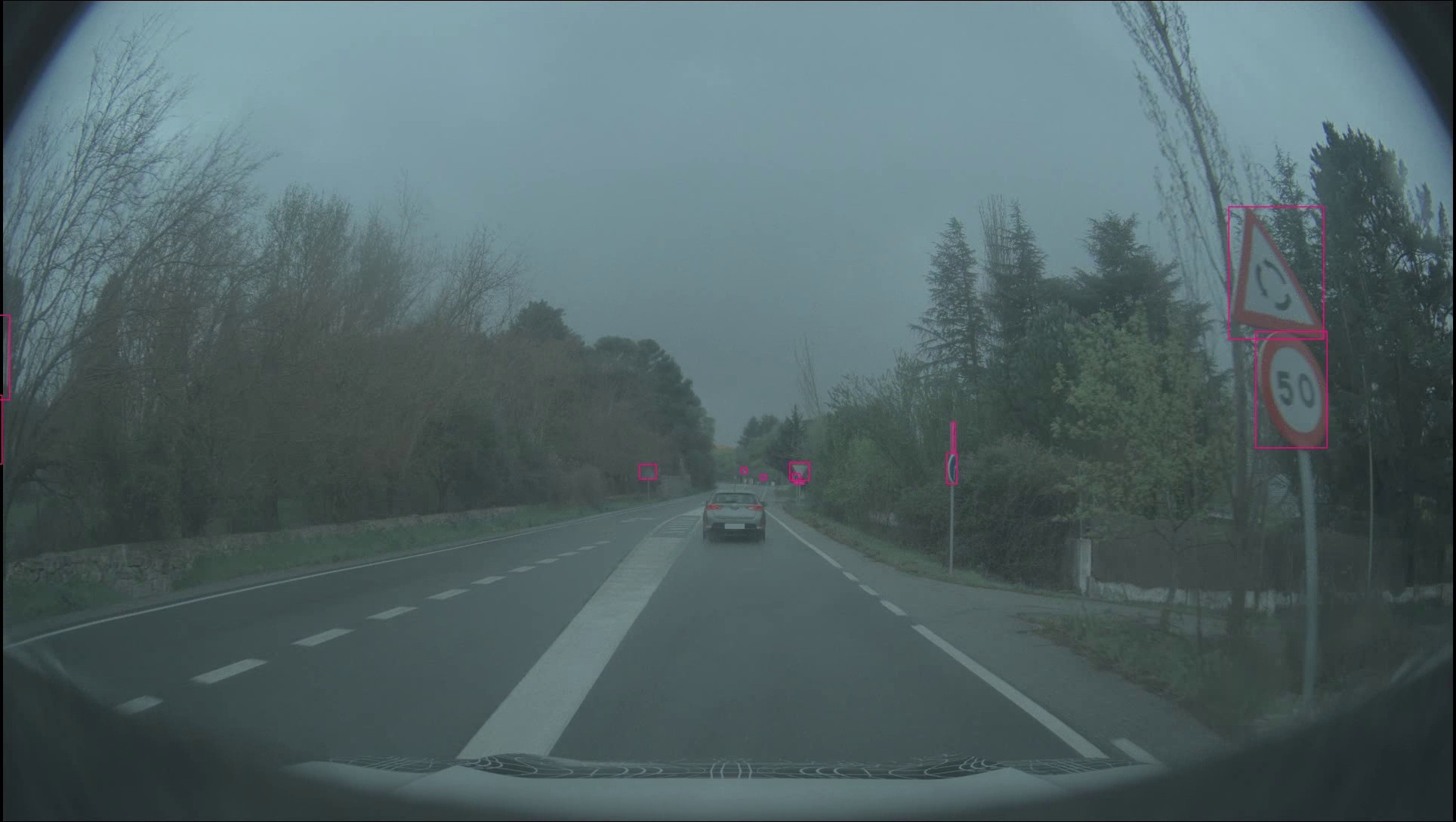}
\\[\smallskipamount]
\includegraphics[width=.32\textwidth]{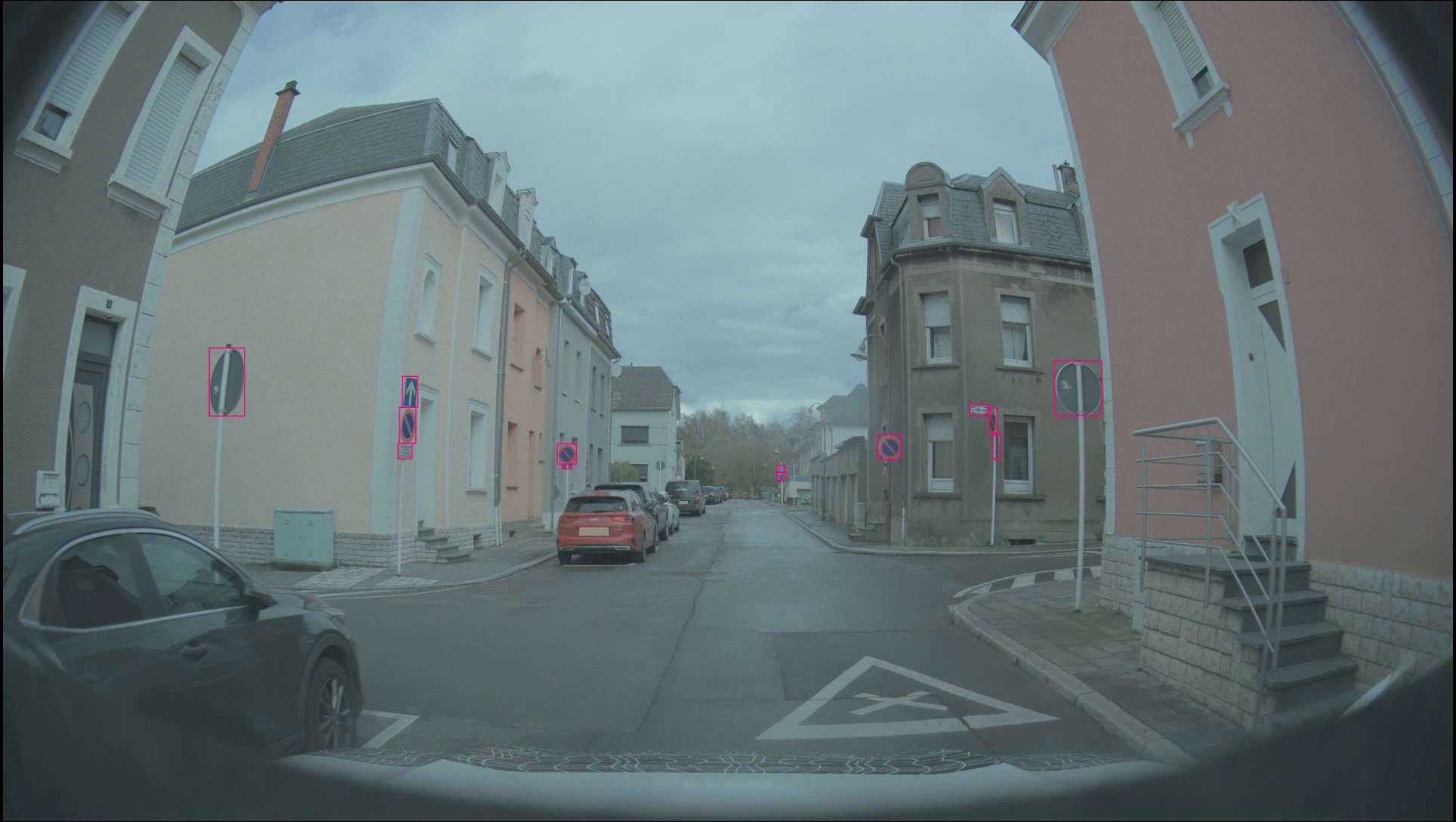}
\includegraphics[width=.32\textwidth]{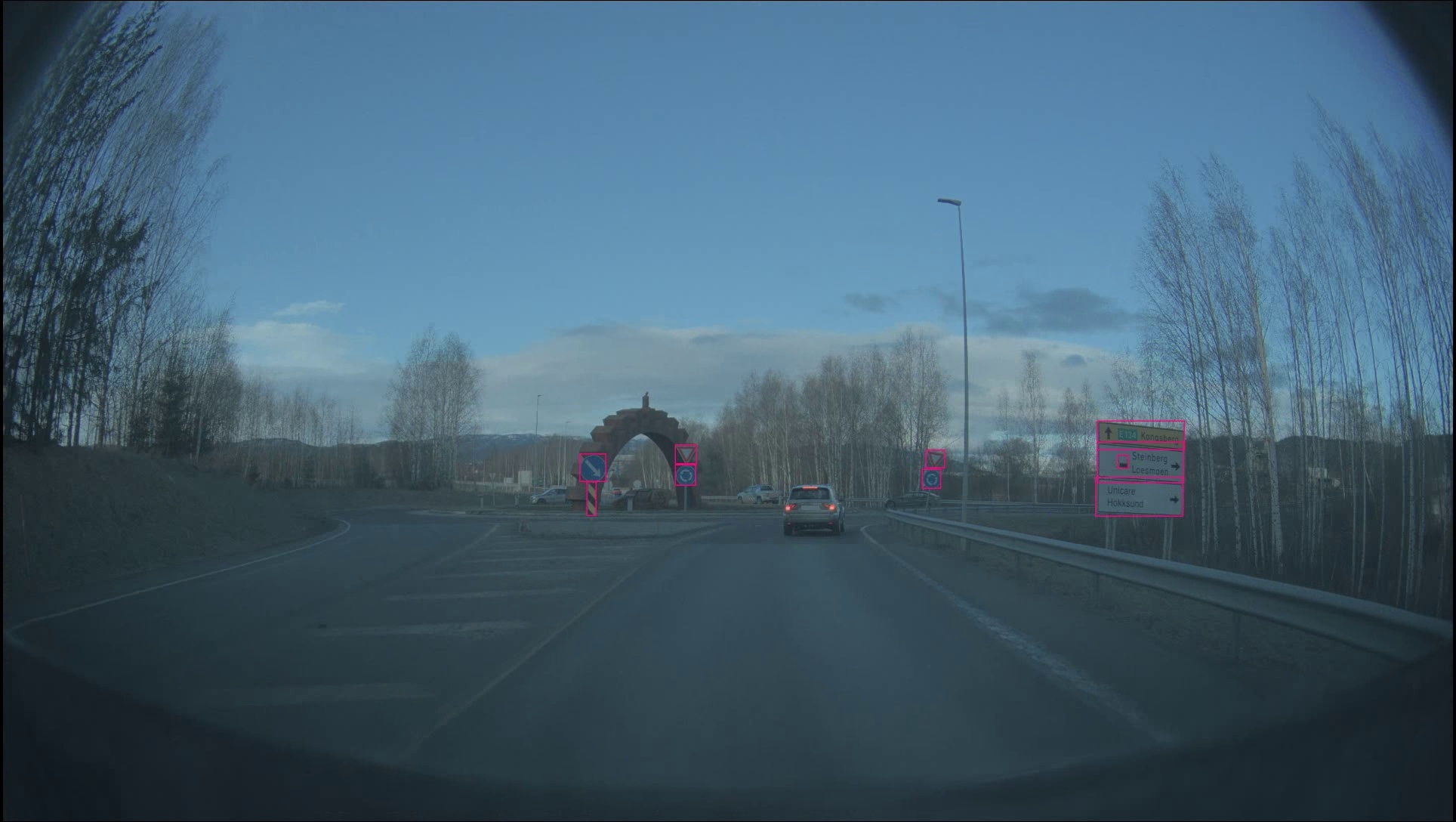}
\includegraphics[width=.32\textwidth]{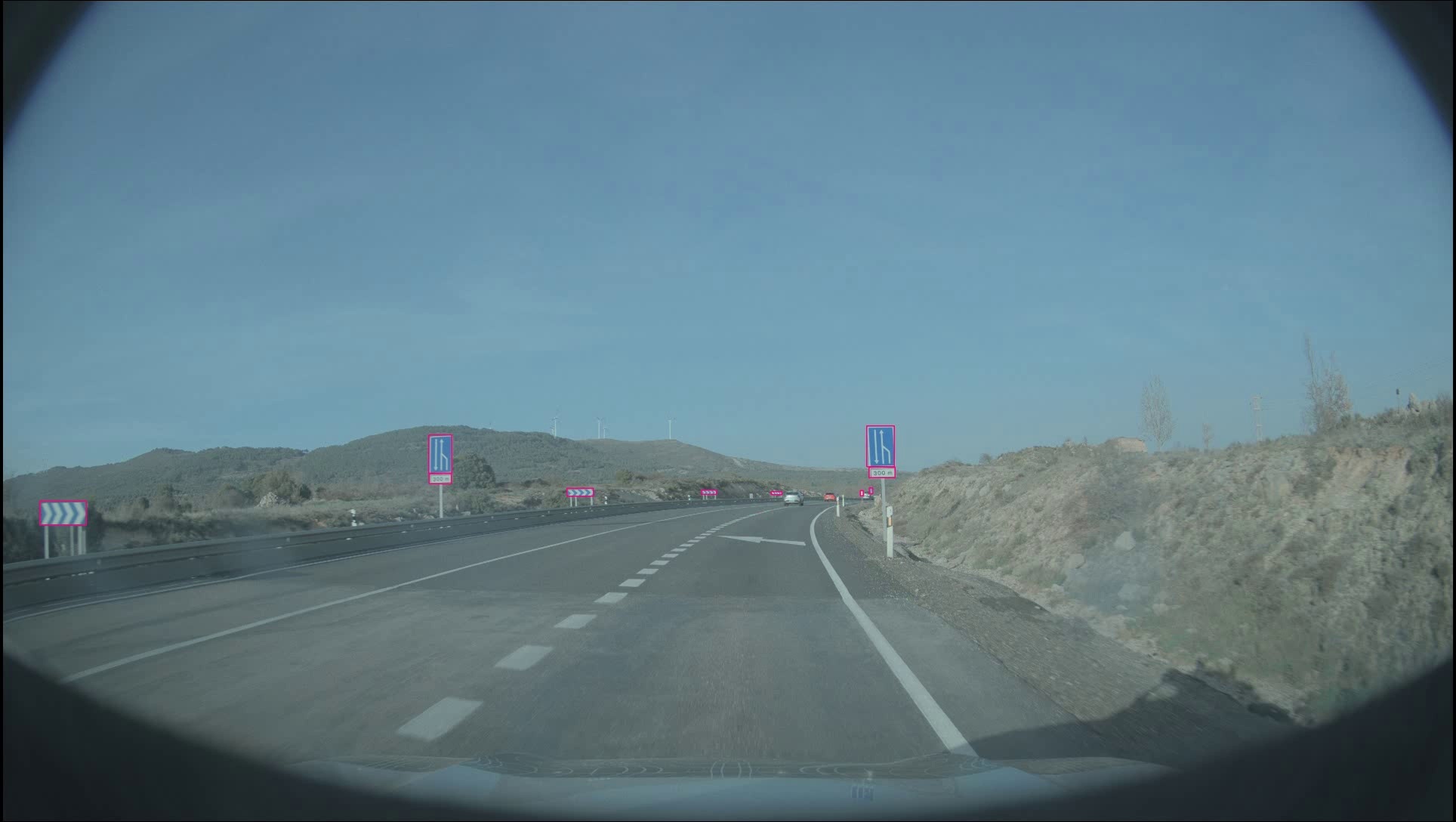}
\caption{Examples of detections generated by our detection framework on different scenarios including roundabouts, construction sites, and cities (best viewed by zooming in on the PDF image).}\label{fig:detector_comparison}
\end{figure*}

The fused multi-modal features are encoded by a SwinTransformer backbone~\cite{liu2021swintransformerhierarchicalvision}, whose hierarchical, window-based self-attention is well suited to capturing both fine-grained sign features and broader contextual cues. The resulting multi-scale feature pyramid is then passed to a transformer-based, anchor-driven detection head trained with the Co-DETR-style collaborative hybrid assignment scheme~\cite{zong2023detrscollaborativehybridassignments}. Operating on fused multi-modal features allows precise bounding box regression at extreme distances. This enriched feature set directly addresses the small-object deficiency prevalent in current general-purpose detectors, providing the robustness required for high-speed autonomous driving environments. In Figure \ref{fig:detector_comparison}, we present detections generated by our 2D detector in different scenarios.

\subsection{Tracking Algorithm}
To ensure identity consistency, we implement a robust 2D object tracking module that addresses the limitations of state-of-the-art trackers such as BoostTrack~\cite{stanojevic2024boosttrack, stanojevic2024boosttrack++} and OC-SORT~\cite{cao2023observationcentricsortrethinkingsort}. While these methods often rely on linear motion or constant-velocity assumptions~\cite{7533003, 8296962}, they struggle with the unique challenges of traffic sign tracking, specifically the non-linear scale and perspective changes that occur as the ego-vehicle rapidly approaches a sign. To address these limitations, we propose a dual motion model Kalman filter paired with a multi-stage data association strategy.

\subsubsection{Dual Motion Models}
The motion of a traffic sign in the image plane is governed by the ego-vehicle's motion and by the laws of perspective, both of which produce strongly non-linear scale and position changes as the vehicle approaches a sign. A single linear or constant-velocity model, as used in~\cite{stanojevic2024boosttrack++} or in the virtual trajectories of OC-SORT~\cite{cao2023observationcentricsortrethinkingsort}, cannot capture this behavior across the full operating range. Instead of selecting a single motion model, we run two complementary higher-order Kalman filters in parallel on every track and fuse their predictions, as detailed in Section~\ref{subsubsec:fusion_trajectory_synthesis}. Each filter is constructed to be sensitive to a different regime of perspective dynamics:

\begin{itemize}
    \item \textbf{Third-order kinematic model:} A constant-jerk filter that captures the smooth, higher-order curvature in the trajectory of distant signs, where small ego-motion changes induce subtle but accelerating perspective shifts. The state vector includes the bounding box center position, velocity, acceleration, and jerk, together with the width and height and their first-order rates of change, so that scale dynamics arising from the looming effect are explicitly tracked:
    \[
    \mathbf{x} = [x,\, \dot{x},\, \ddot{x},\, \dddot{x},\, y,\, \dot{y},\, \ddot{y},\, \dddot{y},\, w,\, \dot{w},\, h,\, \dot{h}]^T
    \]
    The state transition matrix $F_{\text{jerk}}$ is defined as:
    \[
    F_{\text{jerk}} =
    \begin{bmatrix}
    1 & \Delta t & \frac{\Delta t^2}{2} & \frac{\Delta t^3}{6} & 0 & \dots & 0 \\
    0 & 1 & \Delta t & \frac{\Delta t^2}{2} & 0 & \dots & 0 \\
    0 & 0 & 1 & \Delta t & 0 & \dots & 0 \\
    0 & 0 & 0 & 1 & 0 & \dots & 0 \\
    \vdots & & & & \ddots & \\
    0 & 0 & 0 & 0 & 0 & \dots & 1
    \end{bmatrix}
    \]
    \item \textbf{Second-order kinematic model:} A constant-acceleration filter that captures the rapid scale expansion and large pixel displacements characteristic of close-range signs, where the looming effect dominates and is poorly approximated by constant-velocity assumptions used in previous works. The transition matrix $F_{\text{acc}}$ is defined as:
    \[
    F_{\text{acc}} =
    \begin{bmatrix}
    1 & \Delta t & \frac{\Delta t^2}{2} & 0 & 0 & \dots & 0 \\
    0 & 1 & \Delta t & 0 & 0 & \dots & 0 \\
    0 & 0 & 1 & 0 & 0 & \dots & 0 \\
    \vdots & & & \ddots & \\
    0 & 0 & 0 & 0 & 0 & \dots & 1
    \end{bmatrix}
    \]
\end{itemize}

Because both filters operate on every track, the two models contribute complementary predictions throughout the lifetime of a sign rather than being activated only in a specific distance regime. This design is consistent with the cumulative ablation reported in Table~\ref{tab:tracker_ablation} of Section~\ref{sec:results}, where enabling the second-order filter and then additionally enabling the third-order filter each yield incremental improvements over the baseline tracker.

\subsubsection{Multi-Stage Data Association}
When traditional IoU metrics fail due to motion blur or the minute footprint of distant targets, we utilize a composite matching cost $S$ to improve association robustness:
\begin{equation}
\label{eq:matching_cost}
    S = w_1 \, d_M + w_2 \, S_{\text{shape}} + w_3 \, \text{IoU},
\end{equation}
where $d_M$ is the Mahalanobis distance for spatial consistency and $S_{\text{shape}}$ is a shape-similarity term that penalizes unrealistic aspect-ratio fluctuations between a track prediction with width--height $(w_t, h_t)$ and a candidate detection with width--height $(w_d, h_d)$:
\begin{equation}
\label{eq:shape_term}
    S_{\text{shape}} = \left| \log \frac{w_d}{w_t} - \log \frac{h_d}{h_t} \right|.
\end{equation}
The weights $w_1$, $w_2$, and $w_3$ are tuned on a held-out validation split via grid search to maximize tracking F1; the values used in our experiments are $w_1{=}0.25$, $w_2{=}0.25$, and $w_3{=}0.5$. This multi-metric approach ensures stable tracking where detection confidence alone~\cite{stanojevic2024boosttrack} may be insufficient. 

\subsubsection{Fusion and Trajectory Synthesis}
\label{subsubsec:fusion_trajectory_synthesis}
\begin{figure*}[!t]
\centering
\includegraphics[width=.32\textwidth]{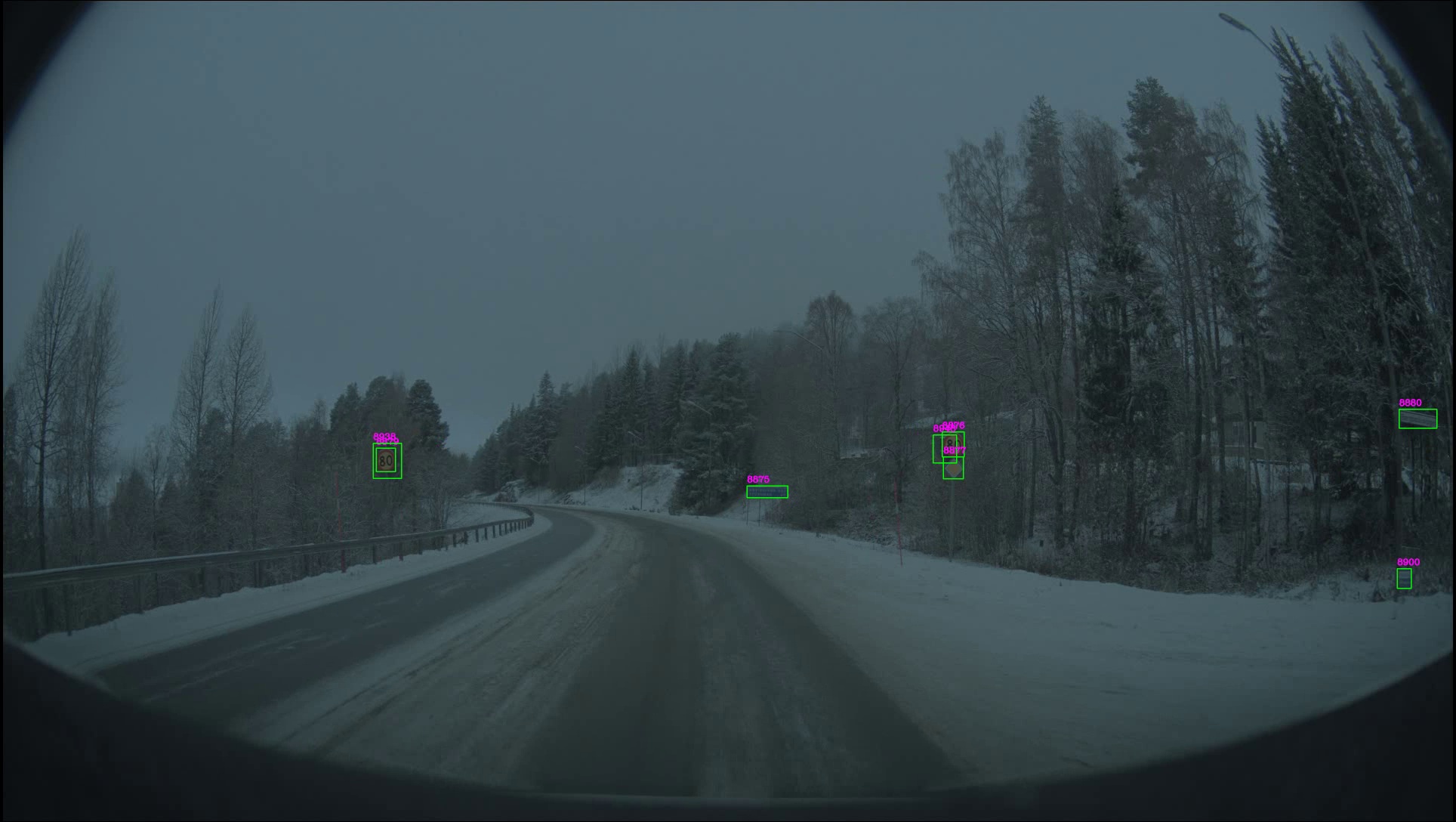}
\includegraphics[width=.32\textwidth]{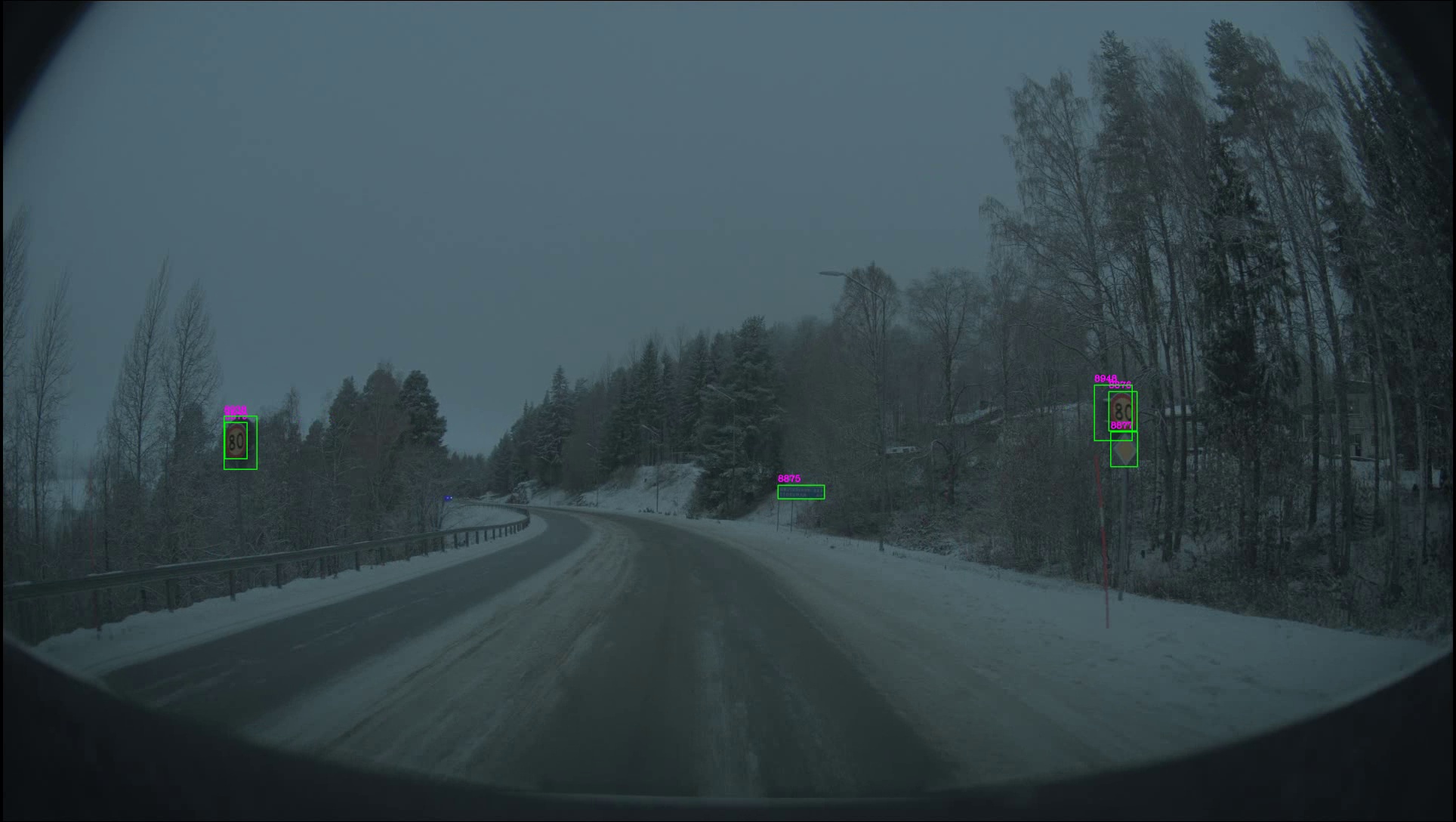}
\includegraphics[width=.32\textwidth]{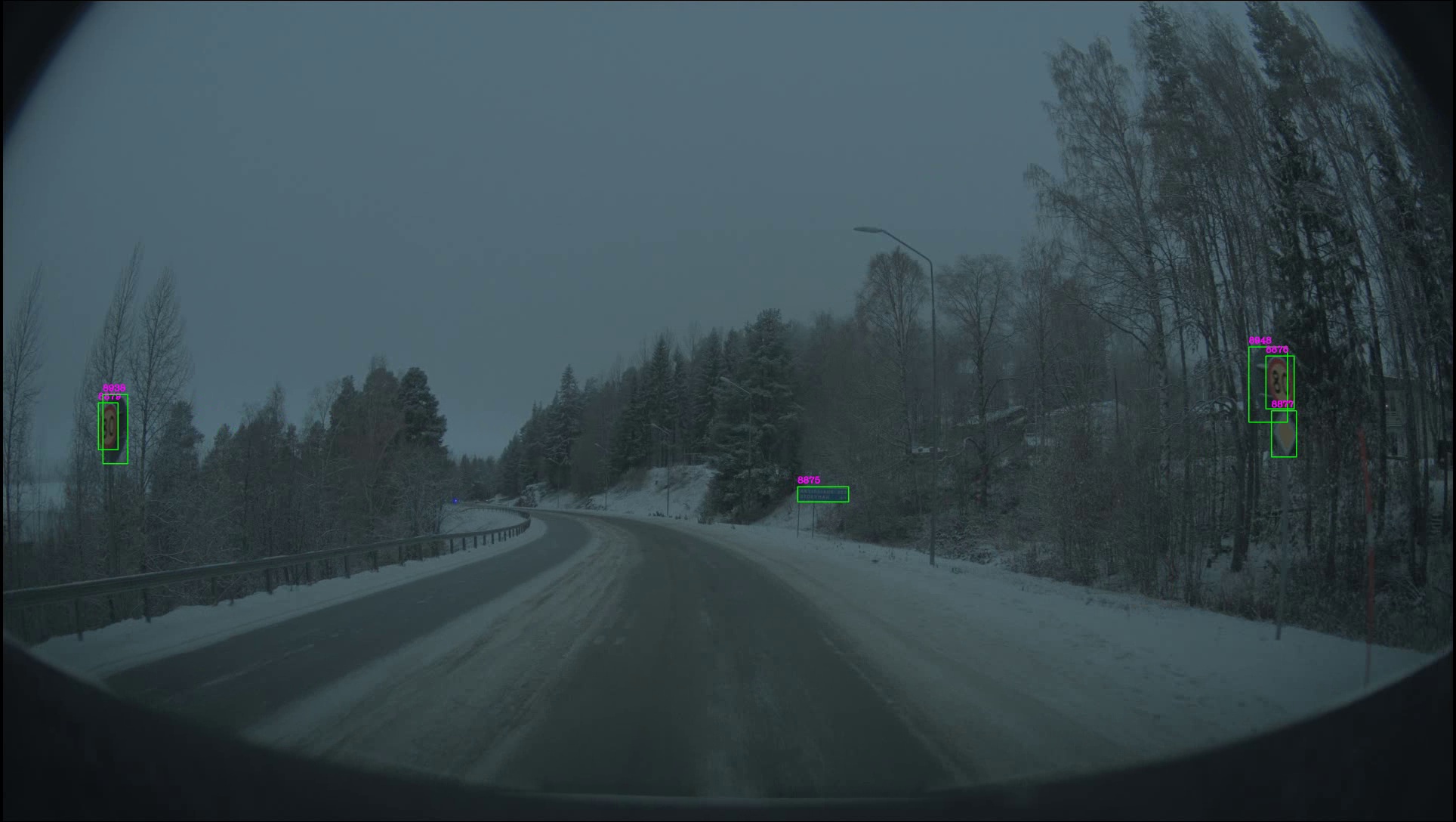}
\\[\smallskipamount]
\includegraphics[width=.32\textwidth]{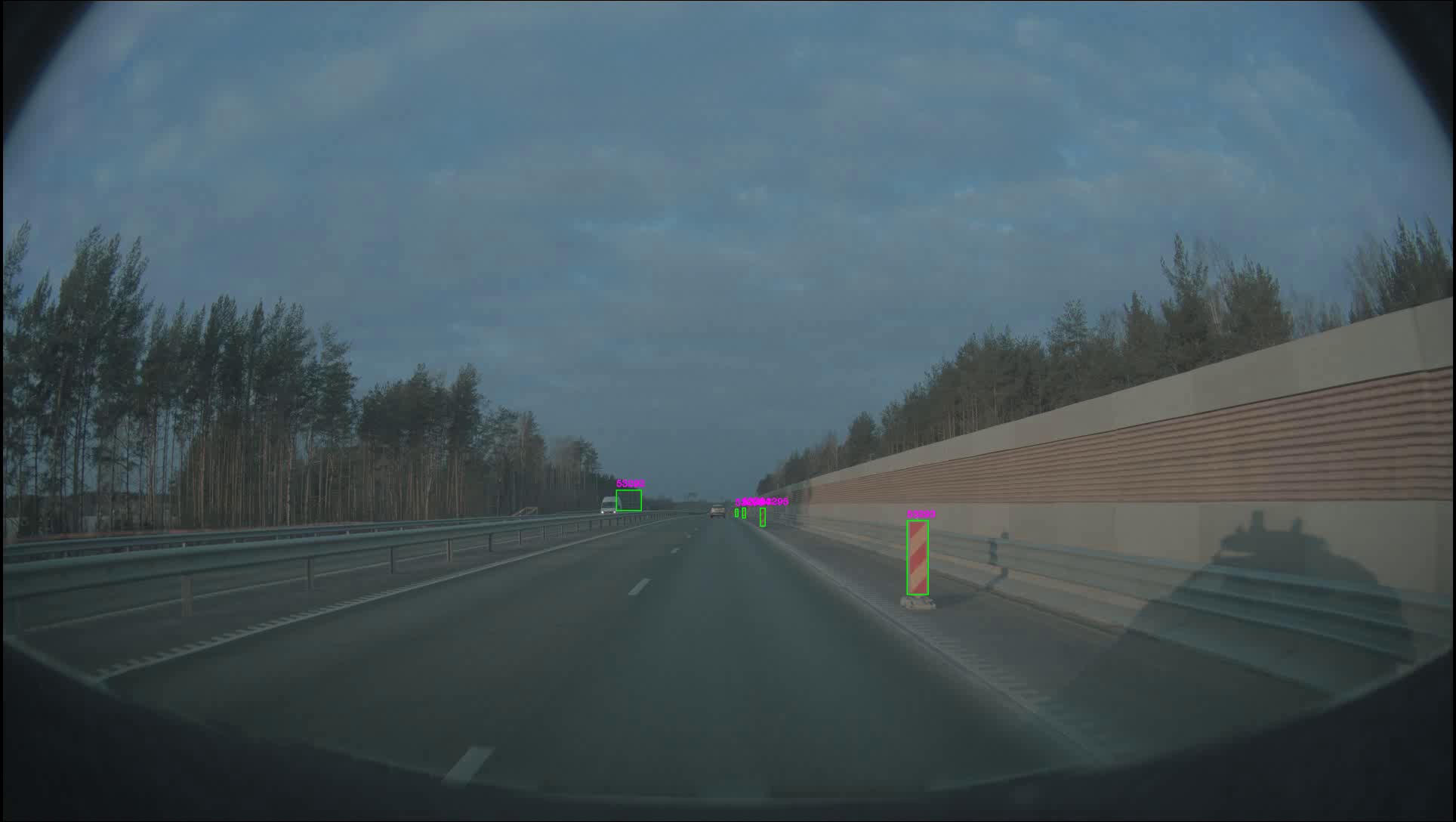}
\includegraphics[width=.32\textwidth]{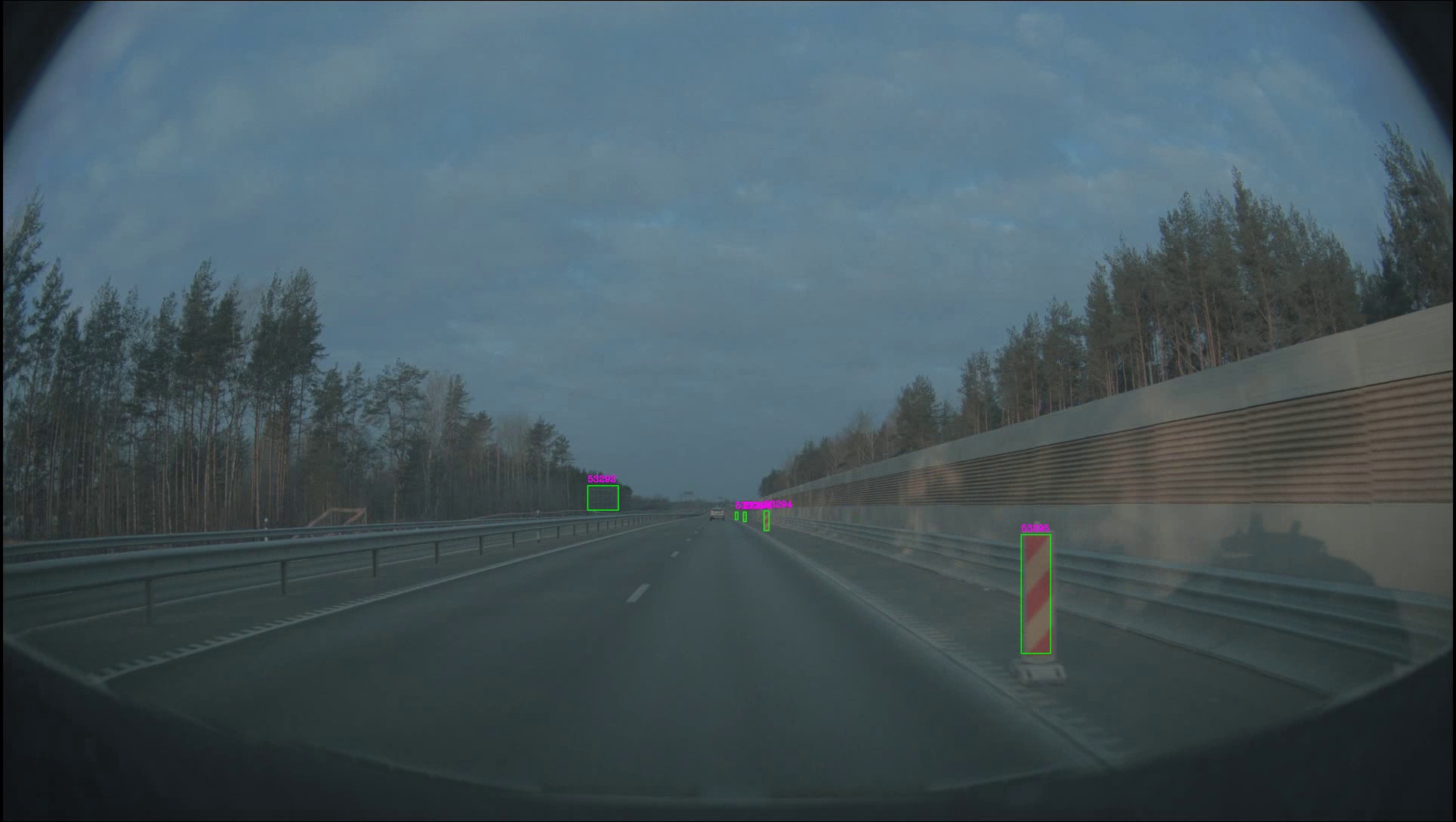}
\includegraphics[width=.32\textwidth]{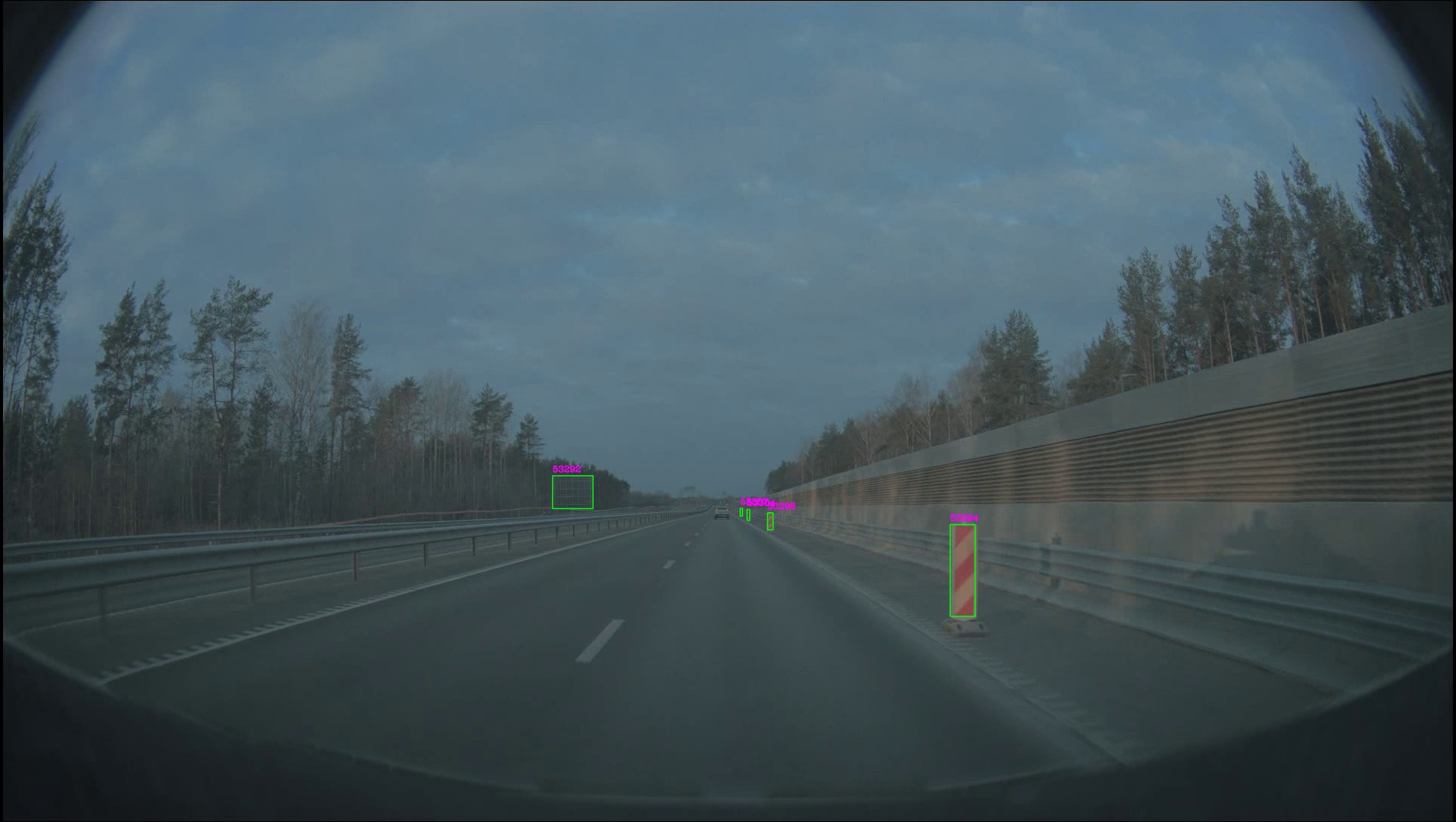}
\\[\smallskipamount]
\includegraphics[width=.32\textwidth]{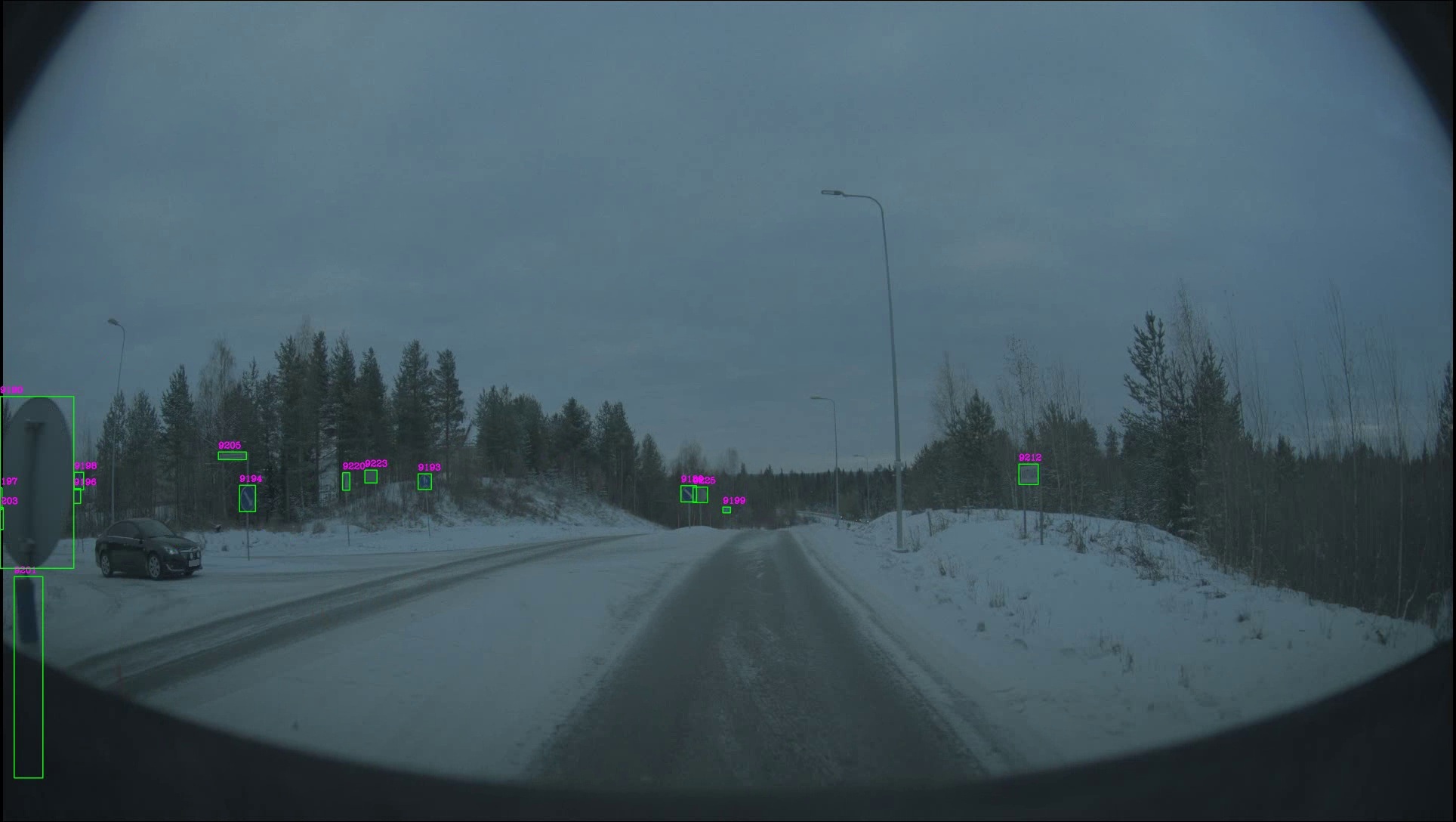}
\includegraphics[width=.32\textwidth]{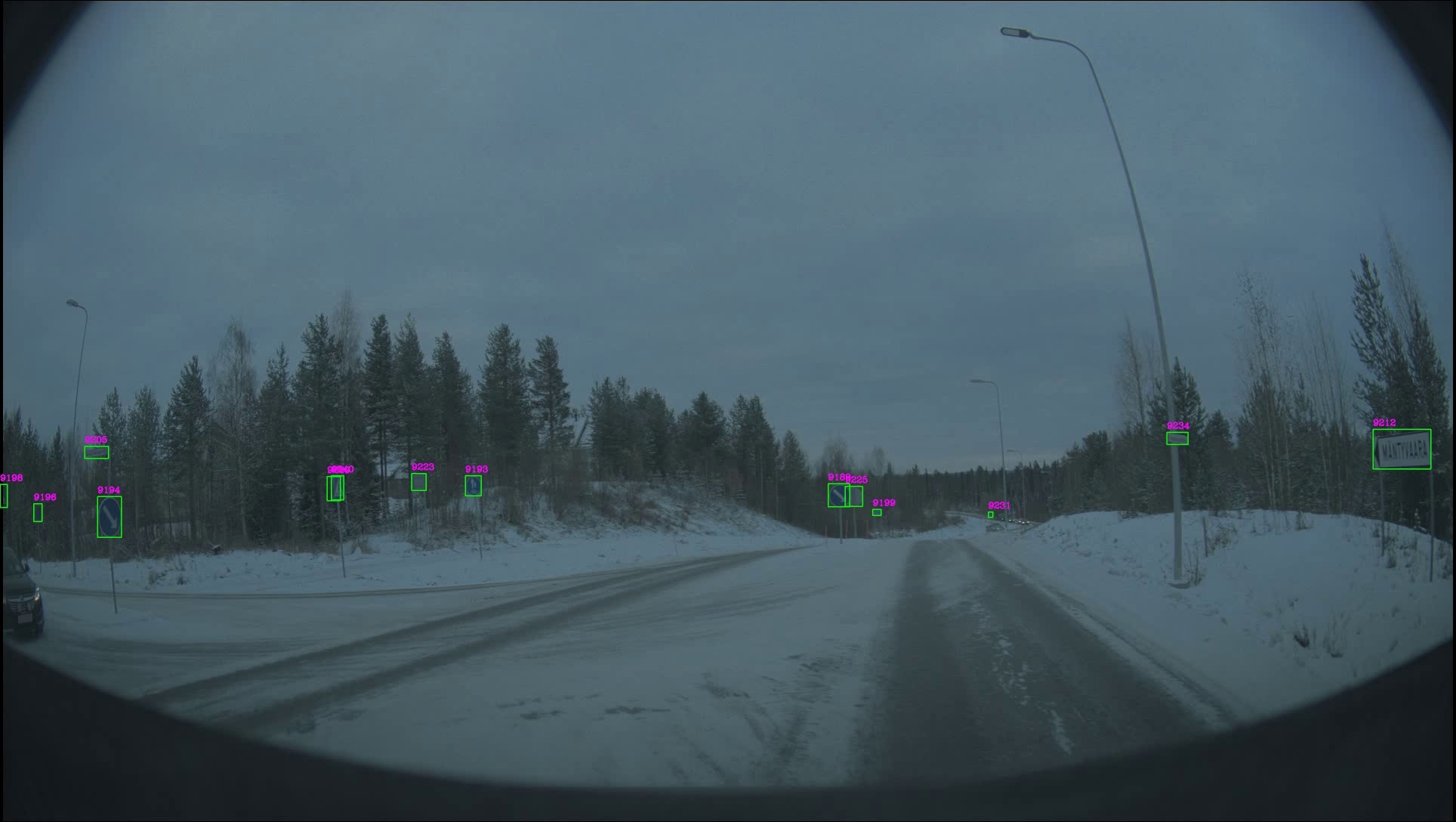}
\includegraphics[width=.32\textwidth]{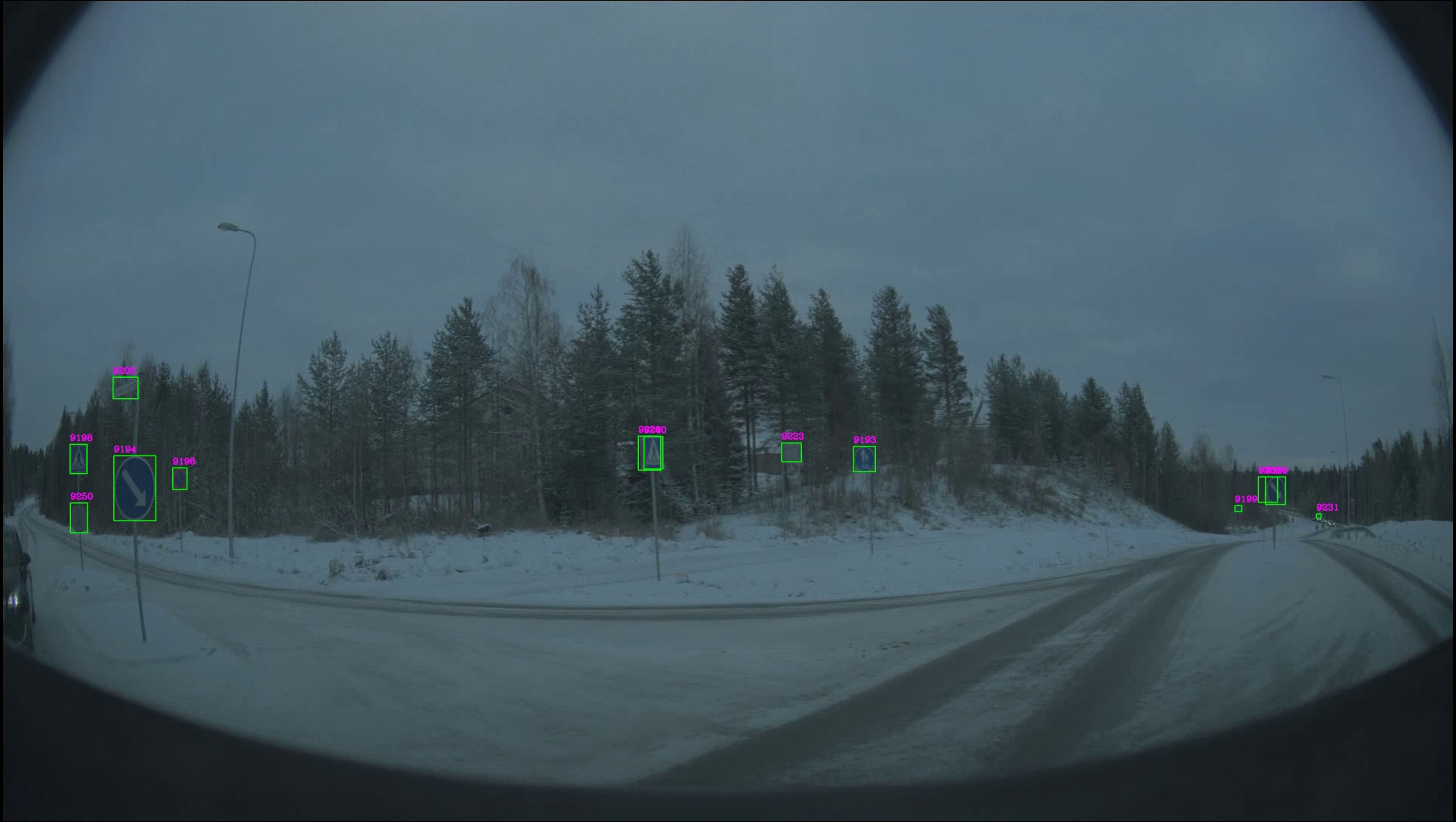}
\\[\smallskipamount]
\includegraphics[width=.32\textwidth]{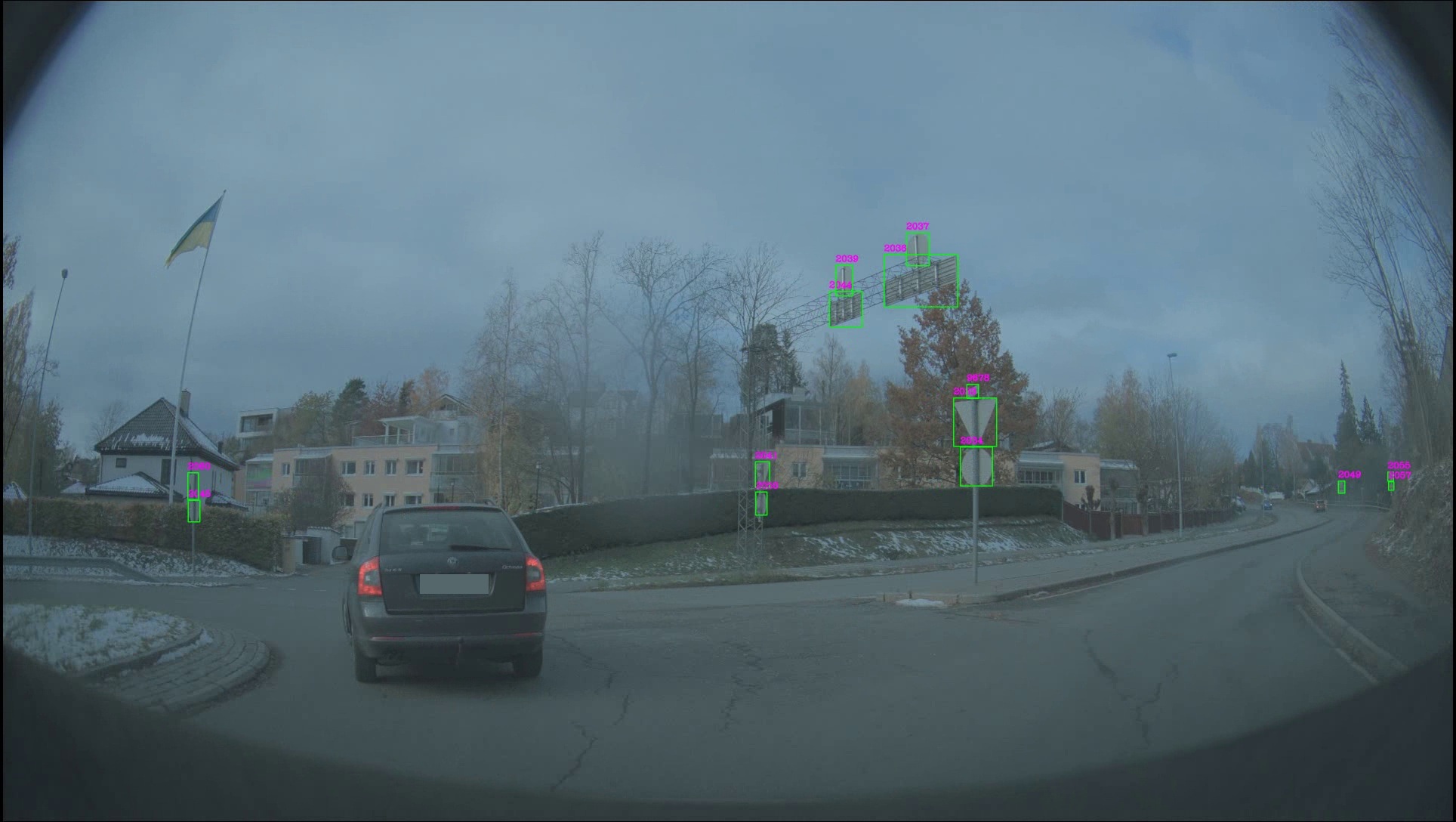}
\includegraphics[width=.32\textwidth]{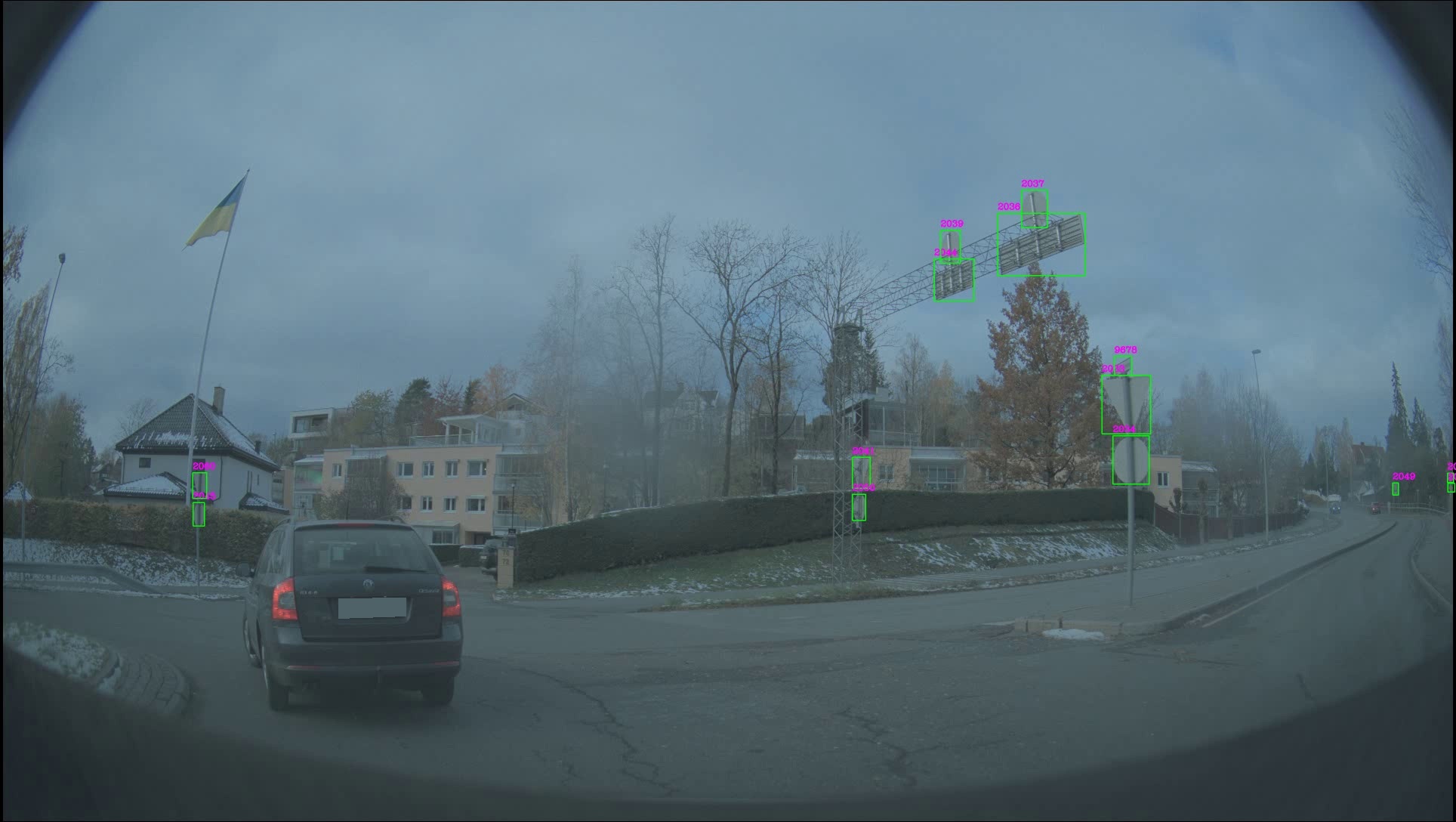}
\includegraphics[width=.32\textwidth]{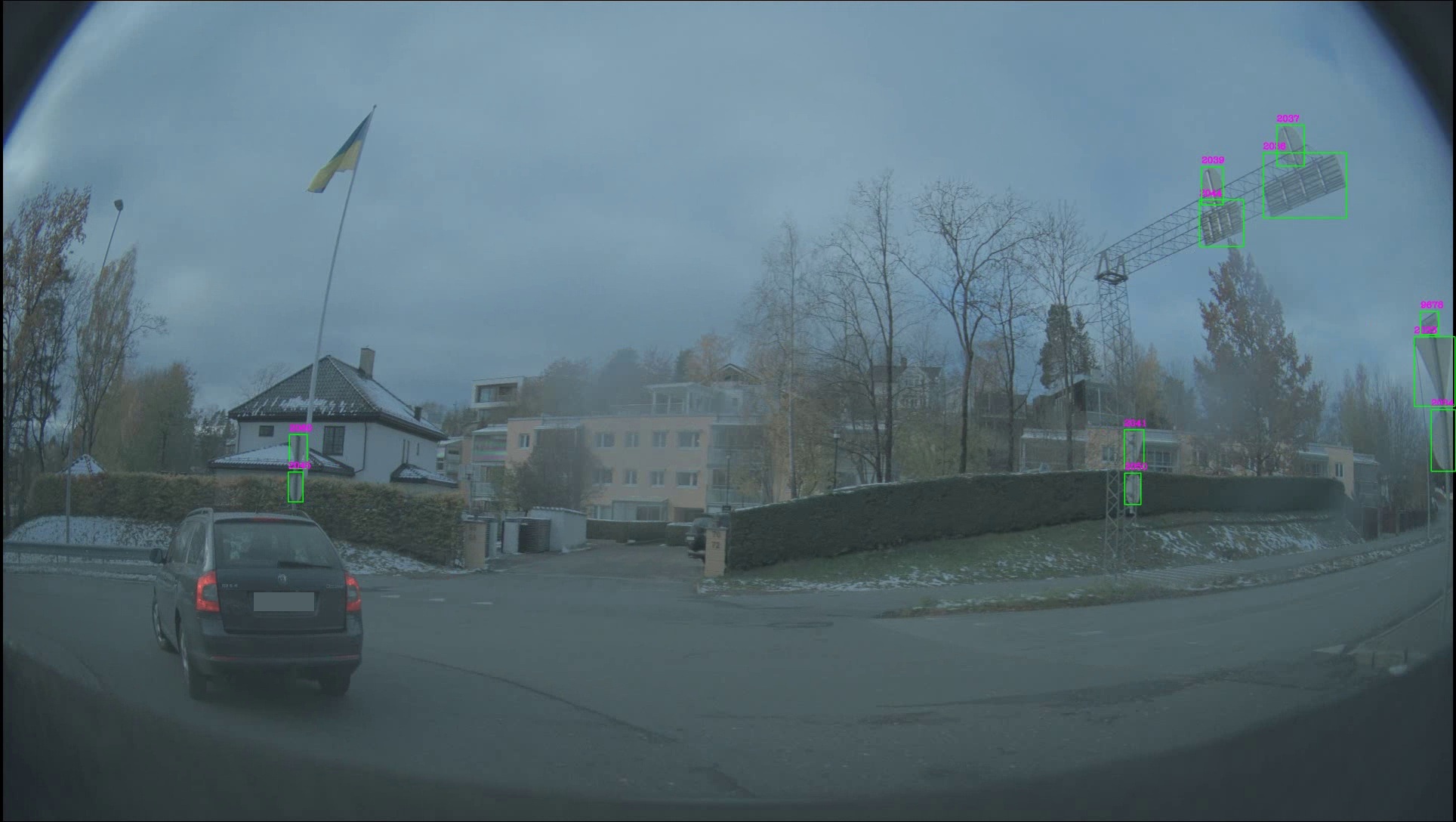}
\caption{Examples of our proposed tracking framework on edge cases such as adverse weather conditions, roundabout driving, and occluded objects (best viewed by zooming in on the PDF image).}\label{fig:comparison_tracker}
\end{figure*}

At each frame, the second-order and third-order filters each produce an updated state estimate for every active track. The two predictions are reconciled through a duplicate suppression mechanism that retains the estimate with the lower posterior covariance trace (interpreted as the more confident estimate) when the predicted boxes overlap, and otherwise treats them as independent observations of the same physical sign. The resulting unified, temporally stable track is then propagated forward, facilitating the consistent assignment of the semantic attributes described in Section~\ref{sec:semanticattributeclassification}.
In Figure \ref{fig:comparison_tracker}, we present tracklets generated by our proposed tracker. Green bounding boxes denote detected traffic signs, while pink labels indicate the corresponding tracking IDs. Three representative frames from four distinct sequences are shown, each capturing challenging edge cases including occlusions, distant signs and significant perspective variations across frames.

%% file: ieee_sections/sec_semantic_attributes.tex
\section{Semantic Attribute Classification}
\label{sec:semanticattributeclassification}

Unlike existing traffic-sign detection frameworks that focus solely on 2D localization and classification, our pipeline introduces a semantic attribute classification layer to provide operational context, a feature rarely modeled jointly in current public datasets. While prior work treats all detections as uniform inputs, our system decouples spatial localization from a composite attribute vector $\mathcal{A}$ to facilitate deeper environmental understanding.

Each detection $d$ is defined by the tuple $(\mathbf{B}, \mathcal{A})$, where:

\begin{itemize}
    \item \textbf{Spatial Component ($\mathbf{B}$):} A 2D bounding box localized within the image plane.
    \item \textbf{Semantic Attributes ($\mathcal{A}$):} A composite vector encoding sign-level properties, including:
    \begin{itemize}
        \item \textbf{Occlusion}: Physical obstruction status.
        \item \textbf{Readability}: Visual interpretability of sign content for downstream tasks.
        \item \textbf{Embedded} ($A_{\text{emb}}$): Hierarchical relationship to the parent signs.       
        \item \textbf{Relevance}: Operational significance relative to the ego-vehicle's path.
    \end{itemize}
\end{itemize}

This structured representation provides a critical filtering mechanism for downstream planning. By classifying these specific attributes, the stack can prioritize actionable detections and ignore irrelevant or unreadable signs. This capability, not found in traditional vision-only systems, significantly reduces the computational burden on the downstream motion planner by pruning the input space to only operationally significant entities.

\subsection{Occlusion Attribute}
Traffic signs in urban environments are frequently obstructed by dynamic agents or static infrastructure. To systematically handle these visibility constraints, we introduce \textbf{occlusion attribute}. This attribute serves as a critical filter, distinguishing between algorithmic failures and cases where detection is physically impossible due to environmental obstructions.

To achieve robust feature extraction under partial visibility, we utilize a \textbf{DINOv2} backbone. Leveraging this self-supervised Vision Transformer allows the system to benefit from strong object-centric representations that generalize well to occluded geometries. To maintain computational efficiency while adapting this large-scale foundation model, we implement \textbf{Low-Rank Adaptation (LoRA)}. The classification process follows a three-stage pipeline:
\begin{enumerate}
    \item \textbf{RoI Extraction:} 2D image patches are cropped from the tracked bounding boxes and resized to the transformer's input resolution.
    \item \textbf{Feature Adaptation:} Trainable rank decomposition matrices are injected into the frozen DINOv2 layers, allowing the model to learn occlusion-specific nuances without full-parameter fine-tuning.
    \item \textbf{Classification:} The model classifies images into five occlusion levels ranging from 0\% to 100\% in 25\% intervals.
\end{enumerate}

The integration of the occlusion attribute enhances the integrity of our performance metrics. By filtering out instances where the sign's defining features are compromised, we ensure that our 2D object detection benchmarks reflect actual model performance rather than visibility limitations. This provides a more accurate assessment of the system's reliability in complex, cluttered scenes.

\subsection{Readability Attribute}
The \textbf{readability attribute} provides a semantic filter to determine if the textual or symbolic content of a traffic sign is visually interpretable. By explicitly encoding the quality of usable information, the system can prioritize high-confidence detections for Optical Character Recognition (OCR) and filter out semantically non-critical instances in cluttered environments.

We define the readability state within a categorical set $\mathcal{R} = \{ r_{\text{readable}}, r_{\text{unreadable}}, r_{\text{back}} \}$:

\begin{itemize}
    \item \textbf{Readable ($r_{\text{readable}}$):} Symbols and borders are distinct and unobstructed. These instances are prioritized for downstream fine-grained classification.
    \item \textbf{Unreadable ($r_{\text{unreadable}}$):} The sign face is visible but compromised by motion blur, severe weathering, or partial occlusion. These are flagged to prevent the propagation of low-confidence features.
    \item \textbf{Back of Sign ($r_{\text{back}}$):} The detection represents the reverse side of a sign. While spatially valid, these are excluded from the set of actionable signs to ensure performance metrics reflect the system's ability to interpret navigationally relevant information.
\end{itemize}

To minimize computational overhead, we employ a patch-based classification pipeline rather than full-frame inference. Detected sign regions are processed through a DINOv2 backbone with LoRA adapters. This architecture is chosen for its robustness against illumination changes and environmental artifacts. To ensure objective performance across diverse lighting and weather conditions, we utilize a balanced sampling strategy during training across all three readability classes. Additionally, we utilize the LiDAR intensity to convert wrongly classified front-facing signs to back of sign, since the back of a sign does not produce high LiDAR intensity values. 

\subsection{Embedded Attribute}
To address cases where signs are physically integrated within larger panels (e.g., speed-limit disks inside informational boards), we introduce the \textbf{embedded attribute} ($A_{\text{emb}} \in \{0,1\}$). Standard detectors often localize these nested objects independently, leading to redundant detections or misrouted downstream logic. By explicitly modeling this hierarchical relationship, we stabilize the scene representation and reduce high-priority evaluation overhead.

We define an embedded relationship $\mathsf{Embedded}(b,B)$ between a candidate sign $b$ and a parent sign $B$ through a conjunctive predicate. A sign is classified as embedded if and only if it satisfies three distinct physical constraints:

\begin{itemize}
    \item \textbf{2D Containment:} The candidate's 2D bounding box must lie entirely within the parent's box extent ($b_c \subset B$).
    \item \textbf{3D Proximity:} The Euclidean distance between the 3D centroids must be below an adaptive threshold ($\| p_b - p_B \| < \tau$), ensuring physical contact rather than perspective overlap.
    \item \textbf{Orientation Consistency:} The yaw difference between objects must fall within a narrow tolerance band to ensure the signs share a common viewing angle.
\end{itemize}

The embedded attribute is treated as a static, frame-invariant property of the object's track. Once the predicate $\mathsf{Embedded}(b,B)$ is triggered in any single frame, the attribute $A_{\text{emb}} = 1$ is propagated across the entire track sequence. This temporal anchoring prevents ``flickering'' attributes caused by occlusion or noise and provides a stable signal for downstream motion planning and mapping. 

\subsection{Relevance Attribute}
The relevance attribute distinguishes between traffic signs that directly influence the ego-vehicle's driving task and those that are operationally inert. To prevent downstream decision-making modules from processing irrelevant noise (e.g., signs on adjacent frontage roads or distant background signs), we implement a hierarchical filtering pipeline that integrates spatial, temporal, and semantic constraints.

The determination of relevance follows a multi-stage pruning process:    
\begin{enumerate}
    \item \textbf{Spatial Drivable-Area Check:} Utilizing lane geometry (from OSM or HD maps), we define a drivable area $\mathcal{D}$. A sign $d$ is considered potentially relevant only if it lies within $\mathcal{D}$ or a proximity threshold $\epsilon$, filtering out signs intended for secondary roads or pedestrian areas.
    \item \textbf{Operational Horizon Distance:} Detections exceeding a 100\,m threshold are classified as \textit{Irrelevant}. The relevance horizon is deliberately tighter than the 200\,m detection horizon: the detector continues to report all signs out to 200\,m so that downstream modules (such as mapping or perception buffering) can use them, while the relevance attribute marks only those signs within the actionable planning horizon at typical highway speeds. At 30\,m/s ($\approx$108\,km/h), a 100\,m horizon corresponds to roughly 3.3 seconds of look-ahead, which is consistent with the planning latency required to react to a directive sign such as a speed-limit change.
    \item \textbf{Geometric Orientation:} Using the output of the readability classifier, any detection identified as a \textit{back-facing} sign ($r_{\text{back}}$) is immediately discarded as irrelevant to the ego-vehicle's direction of travel.
\end{enumerate}

Signs passing the spatial and geometric filters undergo a final \textbf{Content-Based Validation}. A dedicated classifier assesses whether the sign's semantic directive (e.g., a lane-specific speed limit) applies to the ego-vehicle's current trajectory. This ensures that only navigationally significant signs reach the motion planning stack, improving the robustness of the system in dense, multi-lane urban environments.

%% file: ieee_sections/sec_results.tex
\section{Results}
\label{sec:results}

In this section, we evaluate the individual performance of each architectural component. All manual annotations are first labeled in the LiDAR coordinate frame and subsequently projected onto the image plane. Additionally, all objects are evaluated up to the 200\,m boundary without any filtering.

\subsection{Detector} For training the proposed detector, we create a subsampled split using metadata tags, ensuring coverage across all road and weather conditions. The resulting dataset contains over 16{,}000 sequences for training and nearly 7{,}000 sequences for testing. We adopt SwinTransformer~\cite{liu2021swintransformerhierarchicalvision} as the feature extractor with a transformer-based, anchor-driven detection head~\cite{zong2023detrscollaborativehybridassignments}. The network is optimized using AdamW~\cite{loshchilov2019decoupledweightdecayregularization} with a learning rate of $1\times10^{-4}$.
Training is performed in two stages: first, the model is trained on the Zenseact~\cite{alibeigi2023zenseact} dataset for 16 epochs; this checkpoint constitutes our baseline. 
Second, the Zenseact-trained checkpoint is used to initialize training on the Qualcomm dataset, which continues for an additional 16 epochs using the same optimizer settings. We note that the Zenseact and Qualcomm datasets share an identical sensor configuration, so the observed performance difference between the two reflects a data distribution gap rather than a sensor domain gap; this motivates conducting all subsequent experiments on the Qualcomm dataset.
On the Zenseact dataset, our method achieves an AP of 0.55, AP$_{50}$ of 0.78, and AP$_{75}$ of 0.60 (Table~\ref{tab:results_merged}), indicating strong overall performance at both relaxed and strict localization thresholds. When the checkpoint is applied directly to inference on the Qualcomm dataset, performance degrades (AP drops to 0.38) relative to the Zenseact dataset; we treat this inference as the baseline model.
On the Qualcomm dataset, our Qualcomm-trained method substantially outperforms the baseline across all COCO metrics, improving AP by 0.27 absolute points (from 0.38 to 0.65) and AP$_{75}$ by 0.34 points (from 0.39 to 0.73), with particularly strong gains on small and medium objects (AP$_s$ +0.24, from 0.27 to 0.51; AP$_m$ +0.26, from 0.54 to 0.80), indicating improved localization accuracy and robustness across object scales.

\begin{table*}[!t]
    \centering
    \caption{\textbf{Multi-modal 2D object detection results.} Best values per dataset are highlighted in bold. AP follows the COCO protocol; AP$_{50}$ and AP$_{75}$ are evaluated at IoU thresholds 0.50 and 0.75, respectively; AP$_s$, AP$_m$, AP$_l$ correspond to small ($<32^2$ px), medium ($32^2$--$96^2$ px), and large ($>96^2$ px) objects.}    
    \renewcommand{\arraystretch}{1.15}
    \setlength{\tabcolsep}{10pt}
    \small
    \begin{tabular}{lcccccc}
    \toprule
    \rowcolor{lightgreen}
    \textbf{Method} & \textbf{AP} & \textbf{AP$_{50}$} & \textbf{AP$_{75}$} & \textbf{AP$_s$} & \textbf{AP$_m$} & \textbf{AP$_l$} \\
    \midrule

    \rowcolor{lightgray}
    \multicolumn{7}{c}{\textbf{\textit{Zenseact dataset}}} \\
    Model trained on ZOD & 0.55 & 0.78 & 0.60 & 0.44 & 0.80 & 0.86 \\
    \midrule

    \rowcolor{lightgray}
    \multicolumn{7}{c}{\textbf{\textit{Qualcomm dataset}}} \\
    \makecell[l]{Model trained on ZOD 
    }
    &  0.38 & 0.63 & 0.39 & 0.27 & 0.54 & 0.71 \\
    \makecell[l]{Model trained on Qualcomm 
    }
    & \textbf{0.65} & \textbf{0.86} & \textbf{0.73} & \textbf{0.51} & \textbf{0.80} & \textbf{0.90} \\
    \bottomrule
    \end{tabular}
    \label{tab:results_merged}
\end{table*}

\subsection{Tracker} We subsample a separate portion of our dataset to fine-tune the tracker. Unlike the detector split, which emphasizes diversity in road types and weather conditions, the tracking split targets challenging driving conditions, including scenes with varying traffic density, rapid viewpoint changes for traffic signs, and frequent occlusions. Ground-truth tracks are associated with detector outputs using an IoU matching threshold of 0.5. We first establish a detection baseline on this subsampled split using our detector, which yields a recall of 0.73 and precision of 0.66 (Table~\ref{tab:tracker_ablation}). Next, we benchmark 2D traffic-sign tracking on the internal dataset using two tracker baselines~\cite{stanojevic2024boosttrack++,cao2023observationcentricsortrethinkingsort}, and measure how precision and recall change as we integrate each additional module into the pipeline. We use~\cite{stanojevic2024boosttrack++} as the starting point for our ablations.

As shown in Table~\ref{tab:tracker_ablation}, standard trackers like OC-SORT and BoostTrack++ achieve higher precision than the single-frame detector (0.70 and 0.72 vs.\ 0.66) by filtering out transient false positives. However, they suffer from a significant drop in recall (0.68 and 0.69 vs.\ 0.73). This suggests that conventional motion models struggle to associate detections of distant signs under perspective shifts. Our proposed tracking framework is designed to bridge this gap. The baseline version of our tracker achieves a balanced 0.71 Precision and 0.71 Recall. The addition of Interpolation further increases Recall to 0.72 by filling temporal gaps where the detector may have missed a frame, though at a slight cost to precision due to the potential for ``hallucinating'' tracks on short-lived noise. The most significant performance gains are observed with the introduction of higher-order kinematic models:
\begin{enumerate}
    \item \textbf{Second-Order Model:} By incorporating velocity and acceleration into the state transition, the tracker recovers the recall of the single-frame detector (0.73) while maintaining a higher precision (0.71).
    \item \textbf{Third-Order Model:} The integration of a third-order kinematic model (adding jerk to the state vector) yields the highest performance, with a Recall of 0.74 and Precision of 0.71.
\end{enumerate}
The third-order model achieves a recall that exceeds the single-frame detector (0.74 vs.\ 0.73). This shows that our temporal logic is not only filtering detections but also recovering true positives that the detector misses in individual frames due to motion blur or occlusion.
Although traffic signs are stationary in the 3D world, their motion in the 2D image plane is dictated entirely by the ego-vehicle's motion and the laws of perspective. This relationship is non-linear: as the vehicle approaches a sign, the rate of change in the sign's pixel coordinates and scale increases rapidly. A simple first-order (constant-velocity) model fails to approximate this curve at close ranges, and even a second-order (constant-acceleration) model is insufficient when scale changes are most pronounced. Our third-order model captures these higher-order derivatives of perspective shift, ensuring stable track association across the entire range from 200\,m down to the point of passing.
    
\begin{table*}[!t]
    \centering
    \caption{\textbf{Object tracker results and ablation study.} The baseline tracker is BoostTrack++~\cite{stanojevic2024boosttrack++} for which we have configured tracking parameters empirically; ``+Interpolation'' adds linear interpolation across tracking outputs; ``+Second-order kinematic model'' adds a constant-acceleration Kalman filter; ``+Third-order kinematic model'' additionally enables the constant-jerk filter. Best Recall and Precision are highlighted in bold; the last row is our final configuration.}    
    \renewcommand{\arraystretch}{1.15}
    \setlength{\tabcolsep}{22pt}
    \small
{%
    \begin{tabular}{lcc}
    \toprule
    \rowcolor{lightgreen}
    \textbf{Algorithm} & \textbf{Recall} & \textbf{Precision} \\
    \midrule
    Detector + OC-SORT~\cite{cao2023observationcentricsortrethinkingsort} & 0.68 & 0.70 \\
    Detector + BoostTrack++ (Default Parameters) ~\cite{stanojevic2024boosttrack++} & 0.69 & 0.72 \\
    \midrule
    Detector + Baseline Tracker (Tuned Parameters) & 0.71 & 0.71 \\
    + Interpolation & 0.72 & 0.70 \\
    + Second order kinematic model & 0.73 & 0.71 \\
    + Third order kinematic model & \textbf{0.74} & \textbf{0.71} \\
    \bottomrule
    \end{tabular}
}    
    \label{tab:tracker_ablation}
\end{table*}
    
\subsection{Occlusion Classification} This section provides a detailed analysis of the performance of various architectural backbones and training strategies for the task of occlusion classification. We evaluate four distinct approaches: frozen foundation model features (DINOv2 Frozen), standard Vision Transformers adapted with LoRA (ViT-Base~\cite{dosovitskiy2021imageworth16x16words} + LoRA), foundation models adapted with LoRA (DINOv2 + LoRA), and a transformer architecture trained from scratch (ViT-Base). \emph{LoRA} is applied with rank $r{=}16$ and scaling $\alpha{=}16$. The \emph{Frozen} setting keeps the DINOv2 backbone fixed and trains only the classification head.

\begin{table*}[!t]
    \centering
    \caption{\textbf{Binary Occlusion Classification attribute estimation results:}
             0~=~not occluded, 1~=~occluded.
             LoRA~\cite{hu2021loralowrankadaptationlarge} is applied to
             ViT-Base~\cite{dosovitskiy2021imageworth16x16words} and
             DINOv2~\cite{oquab2024dinov2learningrobustvisual};
             ViT-Base~\cite{dosovitskiy2021imageworth16x16words} is also trained from scratch. P represents Precision, while R represents Recall. Best results are in bold. }
    \label{tab:occlusion_attribute_2}
    \renewcommand{\arraystretch}{1.15}
    \setlength{\tabcolsep}{16pt}
    \small
    \begin{tabular}{ll | cc | cc}
    \toprule
    \rowcolor{lightgreen}
    \textbf{Image Backbone} & \textbf{Training Strategy}
        & \multicolumn{4}{c}{\textbf{Class}} \\
    \rowcolor{lightgreen}
        & & \multicolumn{2}{c}{\textbf{0}}
        & \multicolumn{2}{c}{\textbf{1}} \\
    \cline{3-6}
    \rowcolor{lightgreen}
        & & \textbf{P} & \textbf{R}
        & \textbf{P} & \textbf{R} \\
    \midrule
    ViT-Base~\cite{liu2022convnet}
        & From Scratch
        & 0.61 & \textbf{0.74}
        & \textbf{0.93} & 0.87 \\
    ViT-Base~\cite{dosovitskiy2021imageworth16x16words}
        & LoRA~\cite{hu2021loralowrankadaptationlarge}
        & 0.64 & 0.71
        & 0.92 & 0.89 \\
    ViT-DINOv2~\cite{oquab2024dinov2learningrobustvisual}
        & Frozen
        & 0.47 & 0.71
        & 0.78 & 0.91 \\
    ViT-DINOv2~\cite{oquab2024dinov2learningrobustvisual}
        & LoRA~\cite{hu2021loralowrankadaptationlarge}
        & \textbf{0.72} & \textbf{0.74}
        & \textbf{0.93} & \textbf{0.92} \\
    \bottomrule
    \end{tabular}
\end{table*}

Across transformer backbones, the learning-rate scale differs between \emph{Frozen} training ($4\times10^{-3}$) and \emph{LoRA} ($3\times10^{-4}$), whereas ViT-Base with LoRA uses a higher learning rate ($3\times10^{-3}$). We train the ViT/DINOv2 variants for 150 epochs with the LoRA adapters. The ViT-Base backbone from scratch is trained for 250 epochs. For all models we choose the best checkpoint based on validation precision and recall metrics. We report occlusion performance under two evaluation protocols: a binary protocol (occluded vs.\ non-occluded) and the full five-class protocol.
The classification of occlusion levels (ranging from 0\% to 100\%) presents a significant challenge due to the semantic ambiguity of partial obstructions. Our empirical results demonstrate that while foundation models provide a strong starting point, domain-specific adaptation is essential for reliable performance in automotive environments.

\begin{table*}[t]
    \centering
    \caption{\textbf{Occlusion classification attribute estimation results} across
             backbones (LoRA~\cite{hu2021loralowrankadaptationlarge} enabled
             only for DINOv2~\cite{oquab2024dinov2learningrobustvisual}).
             Class column reports the occlusion percentage;
             ViT-Base~\cite{dosovitskiy2021imageworth16x16words} is also trained from scratch.
             P represents Precision, while R represents Recall.
             Best results are in bold. }
    \label{tab:occlusion_attribute_5}
    \renewcommand{\arraystretch}{1.15}
    \setlength{\tabcolsep}{4pt}
    \small
    \begin{tabular}{ll | cc | cc | cc | cc | cc}
    \toprule
    \rowcolor{lightgreen}
    \textbf{Image Backbone} & \textbf{Training Strategy}
        & \multicolumn{10}{c}{\textbf{Class (\%)}} \\
    \rowcolor{lightgreen}
        & & \multicolumn{2}{c}{\textbf{0}}
        & \multicolumn{2}{c}{\textbf{25}}
        & \multicolumn{2}{c}{\textbf{50}}
        & \multicolumn{2}{c}{\textbf{75}}
        & \multicolumn{2}{c}{\textbf{100}} \\
    \cline{3-12}
    \rowcolor{lightgreen}
        & & \textbf{P} & \textbf{R}
        & \textbf{P} & \textbf{R}
        & \textbf{P} & \textbf{R}
        & \textbf{P} & \textbf{R}
        & \textbf{P} & \textbf{R} \\
    \midrule
    ViT-Base~\cite{liu2022convnet}
        & From Scratch
        & 0.61 & 0.74
        & 0.54 & 0.52
        & \textbf{0.52} & 0.52
        & \textbf{0.56} & 0.28
        & 0.54 & \textbf{0.68} \\
    ViT-Base~\cite{dosovitskiy2021imageworth16x16words}
        & LoRA~\cite{hu2021loralowrankadaptationlarge}
        & 0.64 & 0.71
        & 0.55 & 0.61
        & 0.50 & 0.54
        & 0.45 & \textbf{0.37}
        & \textbf{0.58} & 0.51 \\
    ViT-DINOv2~\cite{oquab2024dinov2learningrobustvisual}
        & Frozen
        & 0.47 & 0.71
        & 0.38 & 0.45
        & 0.39 & 0.31
        & 0.35 & 0.13
        & 0.49 & 0.52 \\
    ViT-DINOv2~\cite{oquab2024dinov2learningrobustvisual}
        & LoRA~\cite{hu2021loralowrankadaptationlarge}
        & \textbf{0.72} & \textbf{0.74}
        & \textbf{0.61} & \textbf{0.63}
        & \textbf{0.52} & \textbf{0.57}
        & 0.44 & 0.36
        & \textbf{0.58} & 0.59 \\
    \bottomrule
    \end{tabular}
\end{table*}

First, we cast occlusion as a \emph{binary} classification problem to verify that the attribute model reliably separates \emph{occluded} from \emph{non-occluded} instances. Results are reported in Table~\ref{tab:occlusion_attribute_2}. The binary classification results reveal the fundamental ability of each backbone to distinguish between clear and obstructed signs. The DINOv2 Frozen model serves as a baseline for the inherent ``zero-shot'' capabilities of self-supervised features. While it achieves a recall of 0.91 for occluded signs, its precision for non-occluded signs is low (0.47). This indicates that a static foundation model, without domain-specific adaptation, fails to reliably define the boundary of ``perfectly visible'' signs.
The integration of LoRA (Low-Rank Adaptation) stabilizes the models.
\begin{enumerate}
    \item \textbf{ViT-Base + LoRA}: Shows an improvement in precision for the non-occluded class (from 0.47 in the frozen model to 0.64) and a boost in recall for the occluded class (0.89).
    \item \textbf{DINOv2 + LoRA:} This configuration brings the highest results in our experiments. It achieves a recall of 0.92 and precision of 0.93 for occluded signs, while also reaching the highest recall of 0.74 and precision of 0.72 for the non-occluded class.
\end{enumerate}
The ViT-Base backbone, trained from scratch, matches the DINOv2 + LoRA model in Recall for the non-occluded class and Precision for the occluded class. However, it falls behind in Recall on occluded instances. 

In Table~\ref{tab:occlusion_attribute_5}, we present the results of occlusion classification on all five classes.

\textbf{Comparison of Training Strategies: Frozen vs.\ LoRA:} 
A comparison between DINOv2 Frozen and DINOv2 + LoRA highlights the limitations of using foundation models as static feature extractors for specialized tasks. The frozen model achieves a recall of 0.71 for non-occluded signs (0\%), but its performance degrades sharply as occlusion increases, reaching a lowest of 0.13 recall at the 75\% occlusion level. By introducing LoRA (Low-Rank Adaptation), we observe a significant recovery in performance across all categories. Specifically, at 75\% occlusion, recall improves from 0.13 to 0.36, while precision at 0\% occlusion jumps from 0.47 to 0.72. 
This suggests that while DINOv2 contains rich general features, the specific visual cues of obstructed traffic signs require the task-specific ``steering'' provided by parameter-efficient fine-tuning.

\textbf{Foundation Models vs.\ Standard Transformers:}
When comparing ViT-Base + LoRA and DINOv2 + LoRA, the benefits of DINOv2's self-supervised pretraining become evident. Both models utilize identical LoRA adaptation strategies, yet DINOv2 consistently outperforms ViT-Base in the 0\% (0.74 vs.\ 0.71 recall) and 100\% (0.59 vs.\ 0.51 recall) categories. DINOv2 + LoRA provides the most balanced precision-recall profile across the spectrum, suggesting that its pre-trained features are more robust to the varied lighting and perspective distortions found in the Qualcomm dataset.

\textbf{From-Scratch Training vs.\ Parameter-Efficient Adaptation:}
When comparing ViT-Base trained from scratch with DINOv2 + LoRA, In the binary setting, in Table~\ref{tab:occlusion_attribute_2}, both models share the same recall for non-occluded signs (0.74) and the same precision for occluded signs. However, DINOv2 + LoRA improves precision on the non-occluded class (0.72 vs.\ 0.61) and recall on the occluded class (0.92 vs.\ 0.87), indicating fewer misclassifications in both directions.
The gap widens in the five-class protocol as seen in Table~\ref{tab:occlusion_attribute_5}. DINOv2 + LoRA consistently
outperforms ViT-Base from scratch in both precision and recall for the low-to-moderate occlusion levels: at 25\% occlusion, recall rises from 0.52 to 0.63 and precision from 0.54 to 0.61; at 50\%, recall improves from 0.52 to 0.57 while precision remains tied at 0.52. ViT-Base from scratch retains an advantage only in precision at the 75\% level (0.56 vs.\ 0.44) and in recall at 100\% occlusion (0.68 vs.\ 0.59), suggesting that its longer training schedule (250 vs.\ 150 epochs) allows
it to memorize extreme-occlusion patterns at the expense of the intermediate classes. By contrast, DINOv2 + LoRA delivers a more balanced precision--recall profile across the full occlusion classes, confirming that the rich semantic features acquired during self-supervised pretraining, when steered by lightweight LoRA adapters, generalize more effectively than training a randomly initialized backbone---even when the latter is given significantly more training iterations.

\begin{figure}[!t]
\centering
\includegraphics[width=.48\columnwidth]{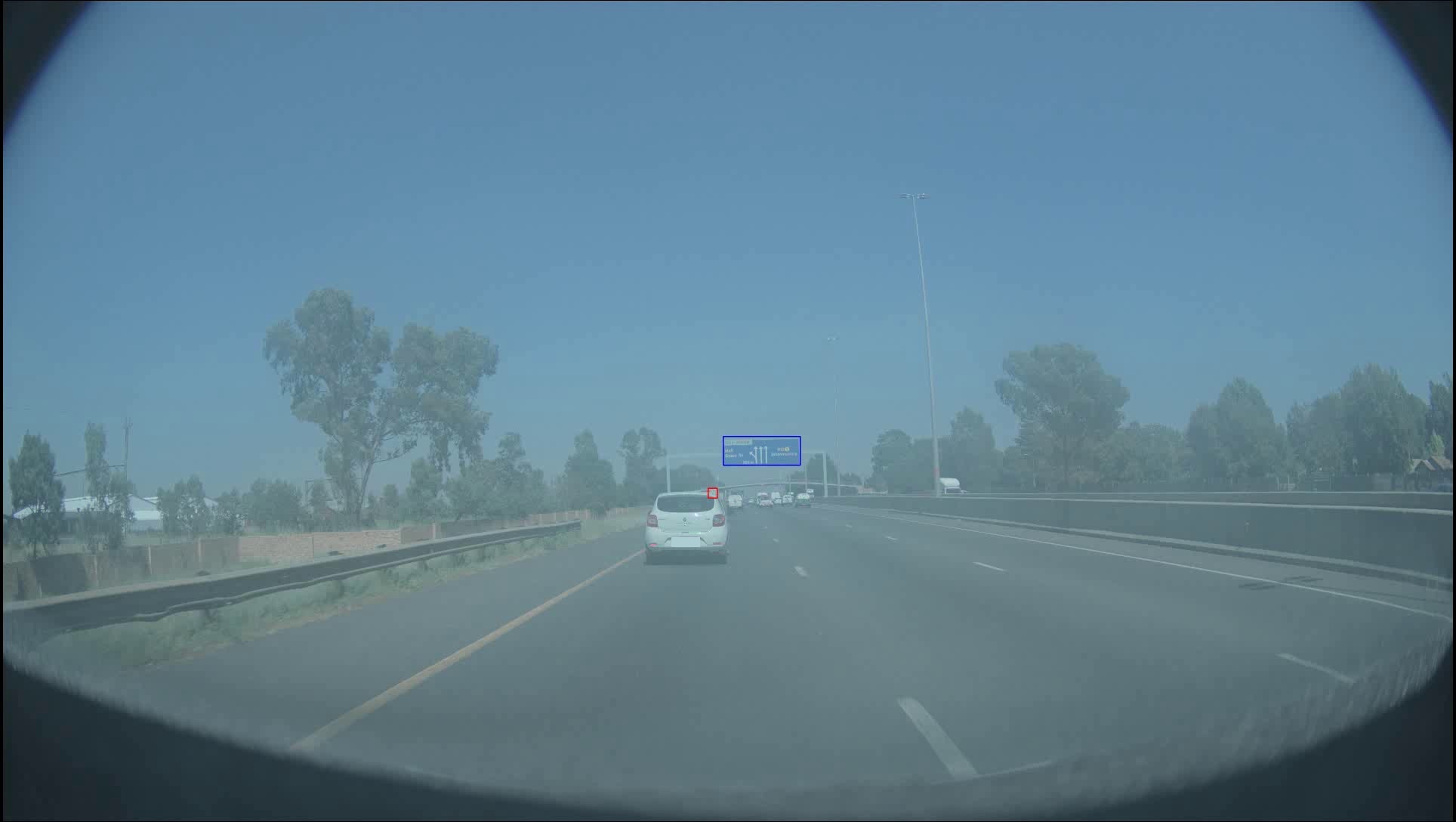}
\includegraphics[width=.48\columnwidth]{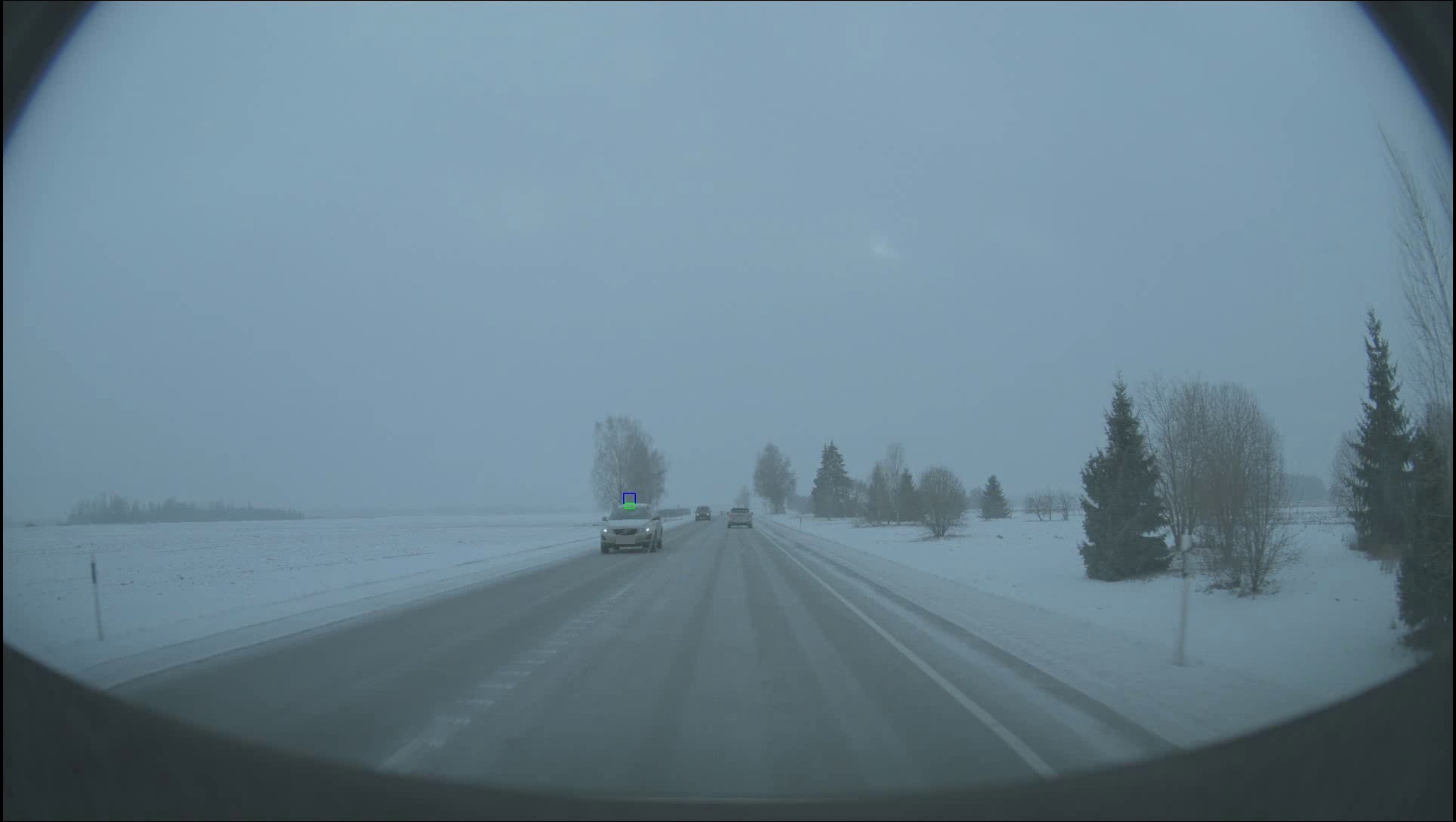}
\\[\smallskipamount]
\includegraphics[width=.48\columnwidth]{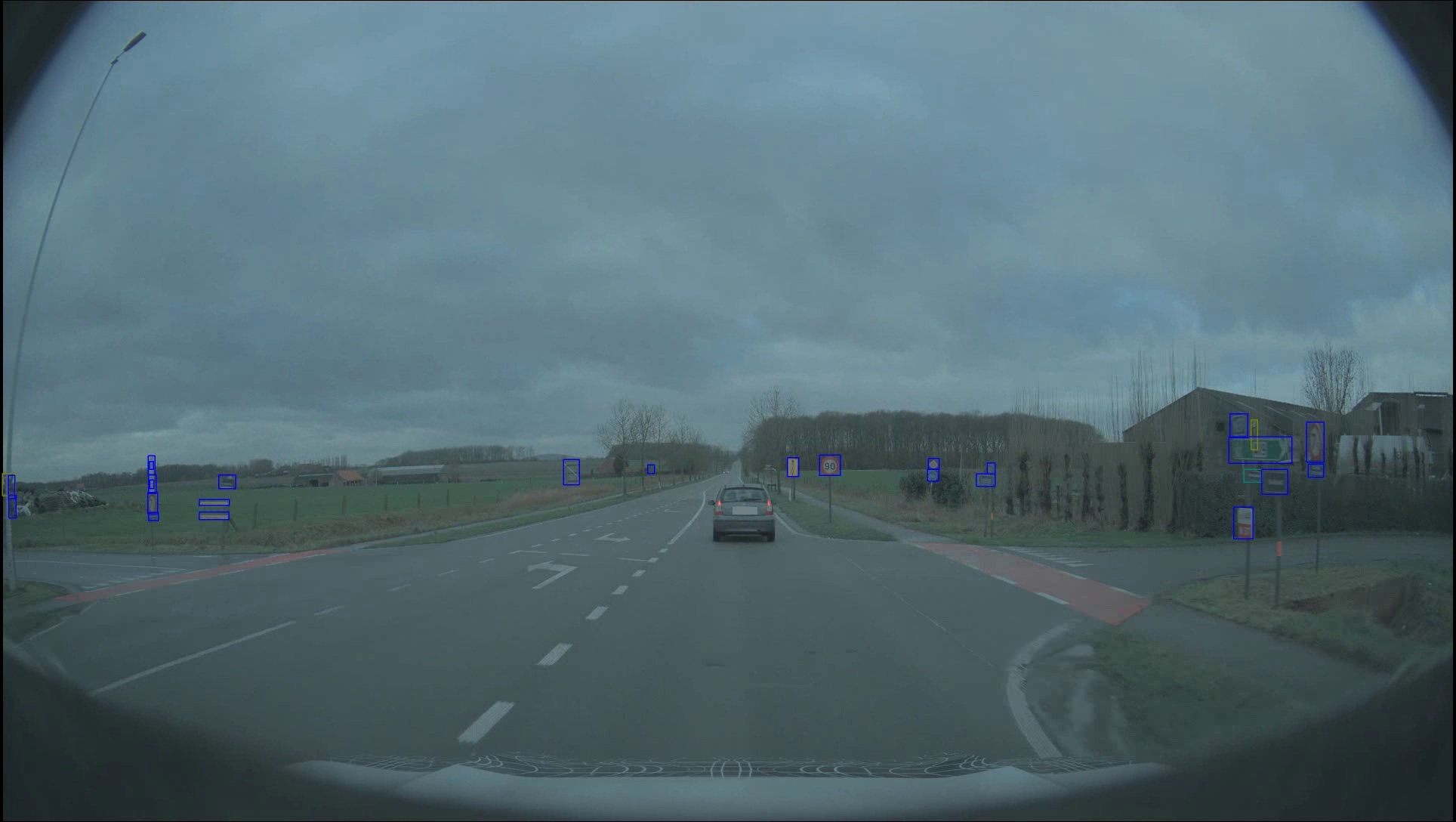}
\includegraphics[width=.48\columnwidth]{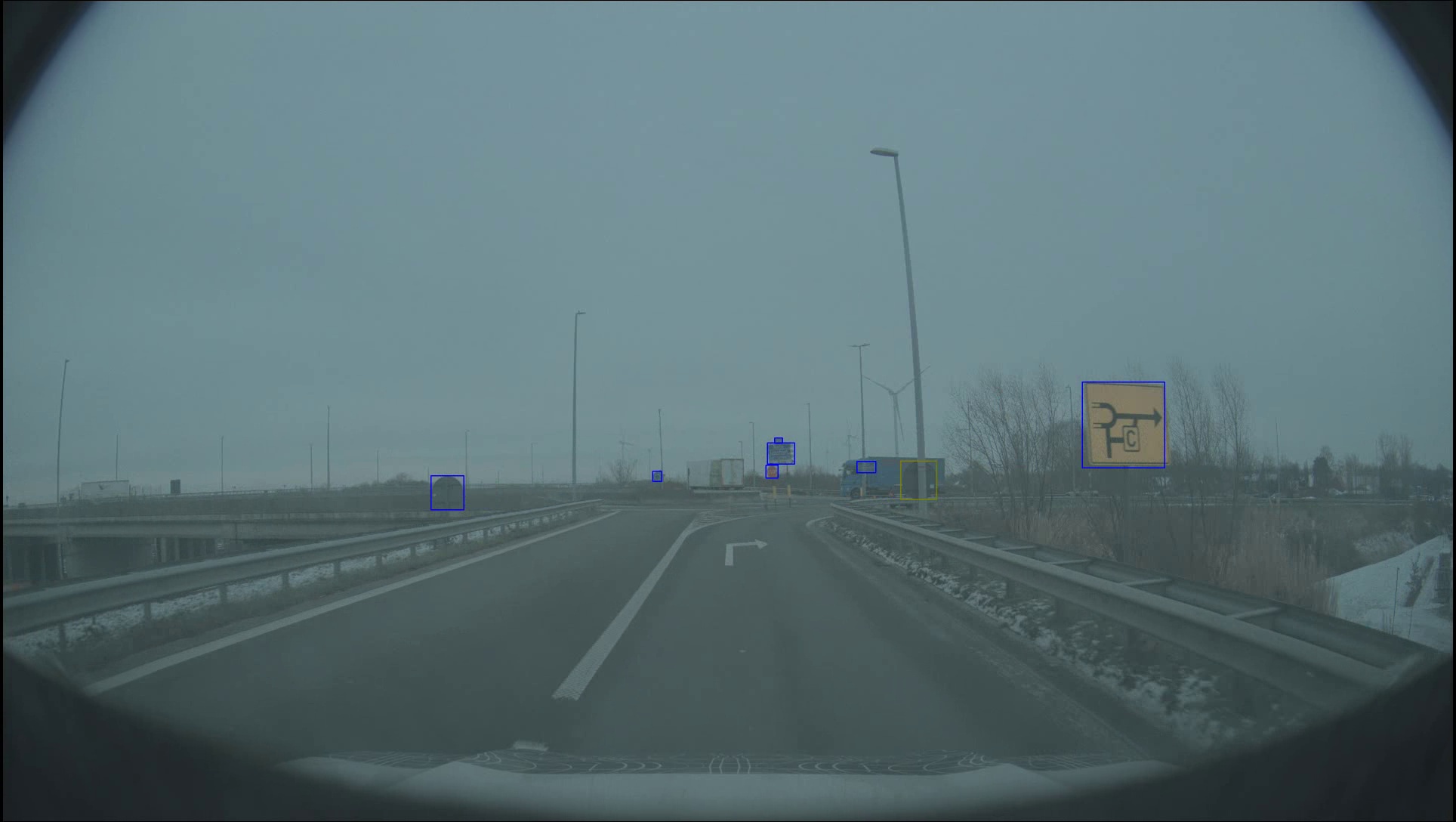}
\caption{Examples of occlusion classification. Bounding-box color encodes the predicted occlusion level: blue~=~0\%, turquoise~=~25\%, yellow~=~50\%, green~=~75\%, red~=~100\%.}\label{fig:occlusion_results}
\end{figure}

\textbf{Analysis of the 75\% Occlusion ``Hard Case'':}
Across all backbones, the 75\% occlusion class remains the most difficult to predict. We attribute this to the high degree of inter-class similarity between a sign that is ``mostly occluded'' (75\%) and one that is ``fully occluded'' (100\%) or ``half occluded'' (50\%). The precision values in this category remain relatively low (0.35 to 0.56), reflecting the subjective nature of these annotations. Nevertheless, the DINOv2 + LoRA configuration offers the best recall-oriented trade-off compared to the other models. Qualitative results are available in Figure ~\ref{fig:occlusion_results}.

\begin{table*}
    \centering
    \caption{\textbf{Readability attribute estimation results} with camera-only vs.\ camera+LiDAR input
             (DINOv2 backbone~\cite{oquab2024dinov2learningrobustvisual} + LoRA ~\cite{hu2021loralowrankadaptationlarge}).
             P represents Precision, while R represents Recall.
             Bold marks the better value per (class, metric) pair across the two input configurations.}
    \label{tab:readability_attribute}
    \renewcommand{\arraystretch}{1.15}
    \setlength{\tabcolsep}{10pt}
    \small
    \begin{tabular}{l | cc | cc | cc}
    \toprule
    \rowcolor{lightgreen}
    \textbf{Modality}
        & \multicolumn{6}{c}{\textbf{Class}} \\
    \rowcolor{lightgreen}
        & \multicolumn{2}{c}{\textbf{Readable}}
        & \multicolumn{2}{c}{\textbf{NotReadable}}
        & \multicolumn{2}{c}{\textbf{BackOfSign}} \\
    \cline{2-7}
    \rowcolor{lightgreen}
        & \textbf{P} & \textbf{R}
        & \textbf{P} & \textbf{R}
        & \textbf{P} & \textbf{R} \\
    \midrule
    Camera Only
        & 0.89 & 0.92
        & 0.96 & 0.89
        & 0.81 & 0.97 \\
    Camera + LiDAR
        & 0.89 & \textbf{0.93}
        & 0.96 & \textbf{0.90}
        & \textbf{0.83} & 0.97 \\
    \bottomrule
    \end{tabular}
\end{table*}

\subsection{Readability Classification} 
\begin{figure}[t]
\centering
\includegraphics[width=.48\columnwidth]{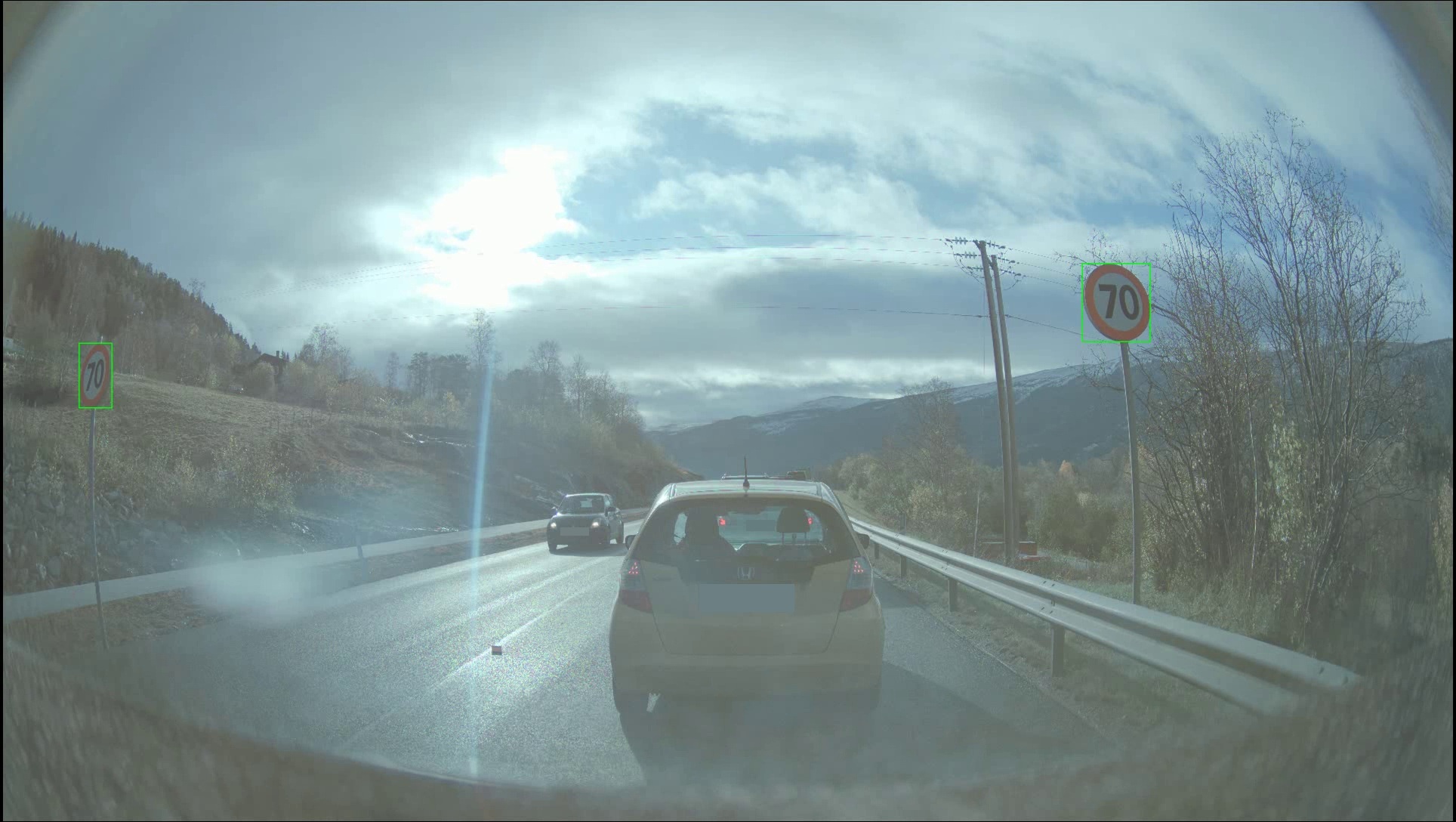}
\includegraphics[width=.48\columnwidth]{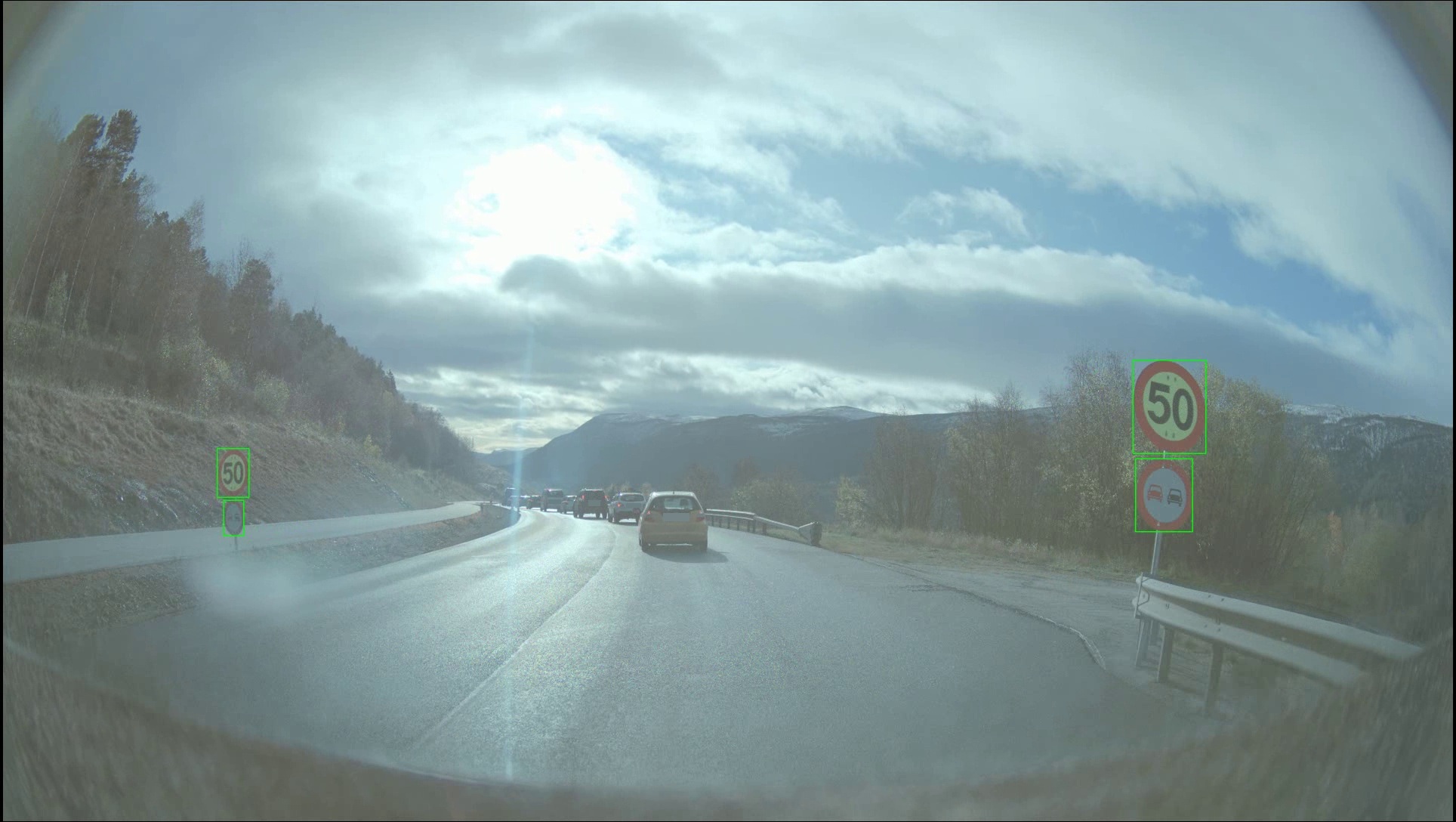}
\\[\smallskipamount]
\includegraphics[width=.48\columnwidth]{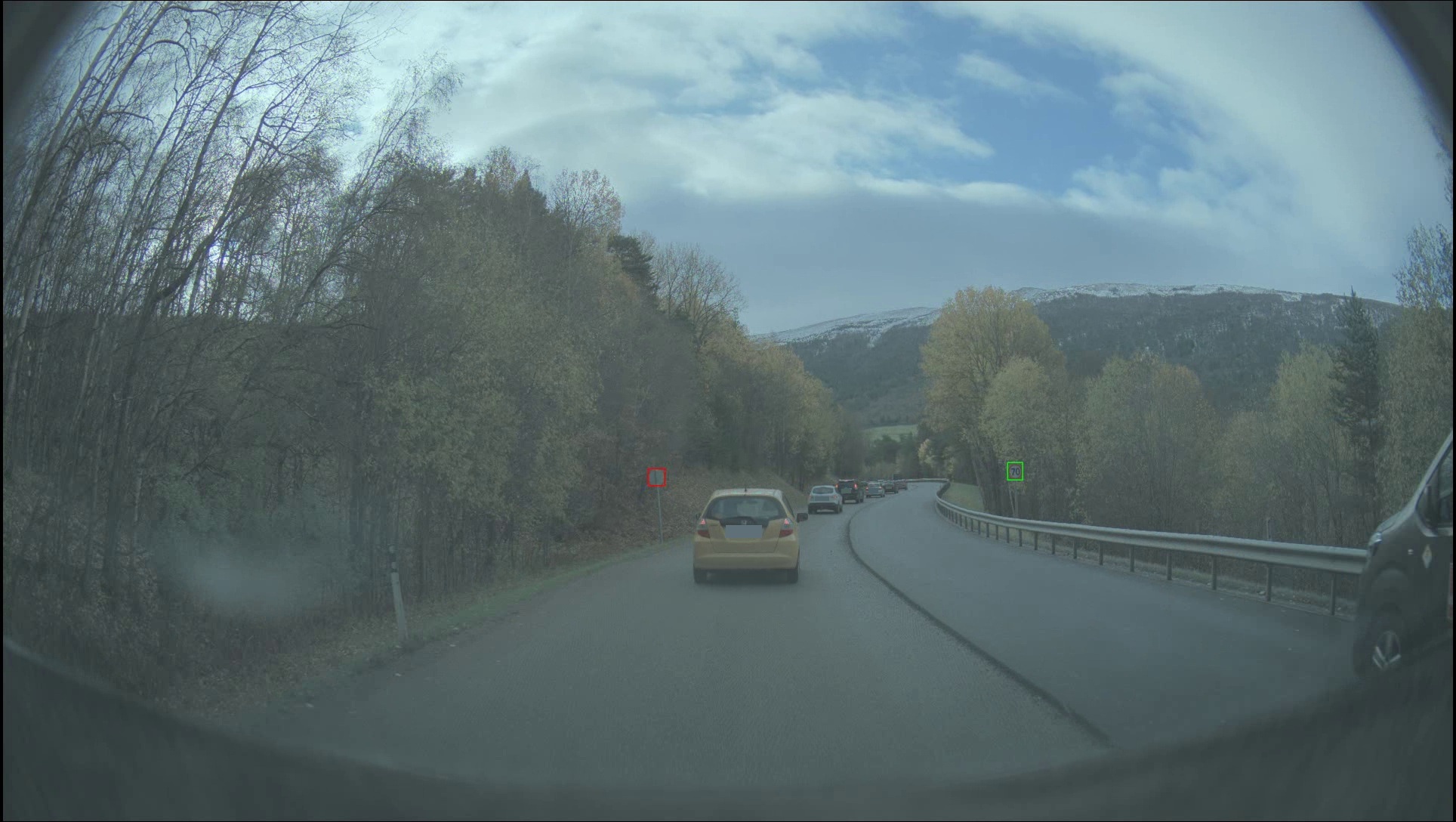}
\includegraphics[width=.48\columnwidth]{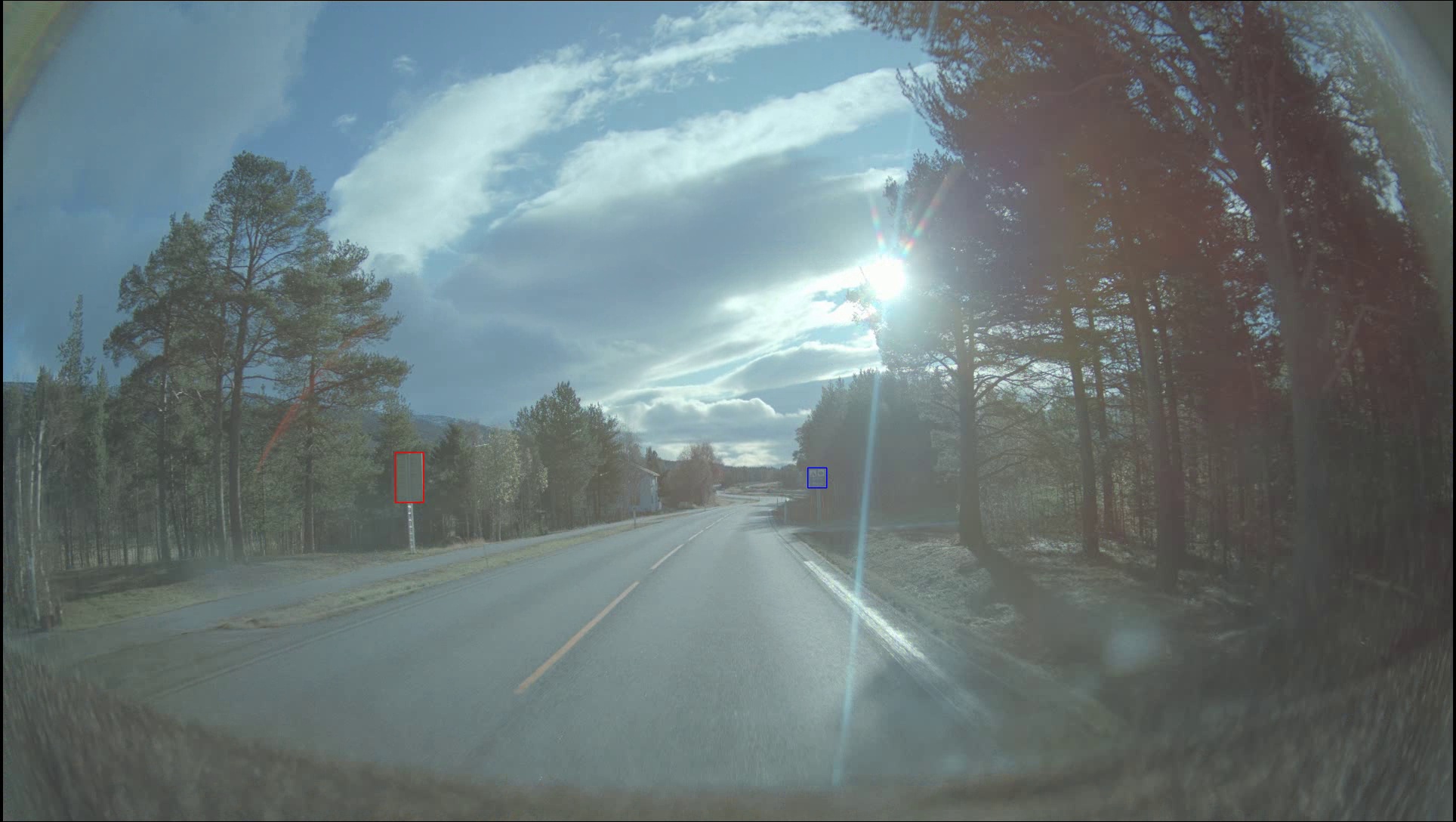}
\caption{Examples of readability classification. The color of the bounding box represents the predicted class: blue~=~NotReadable, red~=~BackOfSign, green~=~Readable.}\label{fig:readability}
\end{figure}

This section provides an analysis of the results for the readability attribute, comparing a vision-only approach with our multi-modal fusion strategy. The findings, summarized in Table~\ref{tab:readability_attribute}, demonstrate the benefits of integrating LiDAR reflectivity data to resolve visual ambiguities. 
The readability attribute, categorized into Readable, NotReadable, and BackOfSign, is vital for the system to determine which signs should influence the vehicle's ego-path. Our results show that while modern convolutional backbones are highly effective, the physical properties of sensors can be leveraged to overcome the limitations of pure vision.
For the readability task, we utilized DINOv2 and LoRA.
As shown in Table~\ref{tab:readability_attribute}, the ``Camera + LiDAR'' configuration improved the precision of the BackOfSign class from 0.81 to 0.83 and increased the recall of the Readable class to 0.93. The primary motivation for adding LiDAR was to address the false positives generated by the BackOfSign prediction.
In vision-only systems, we observed that a few classifications suffered from temporal instability, where the rear of a sign would intermittently be misclassified as a readable front face.
The front face of a traffic sign is designed to be retroreflective, resulting in a higher LiDAR intensity return. The back of a sign, which is usually matte or metallic, produces a much lower intensity. We choose the threshold value for intensity empirically. Qualitative results are presented in Figure ~\ref{fig:readability}.

\subsection{Embedded Classification} 
\begin{figure}[!t]
\centering
\includegraphics[width=.48\columnwidth]{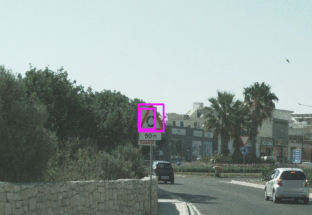}
\includegraphics[width=.48\columnwidth, trim={0 0 0 0.67cm}, clip]{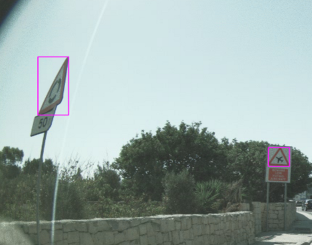}
\caption{Examples of edge cases that are correctly classified by the Embedded classifier as NotEmbedded.}\label{fig:embedded_results_edge_cases}
\end{figure}

\begin{figure}[h]
\centering
\includegraphics[width=.48\columnwidth]{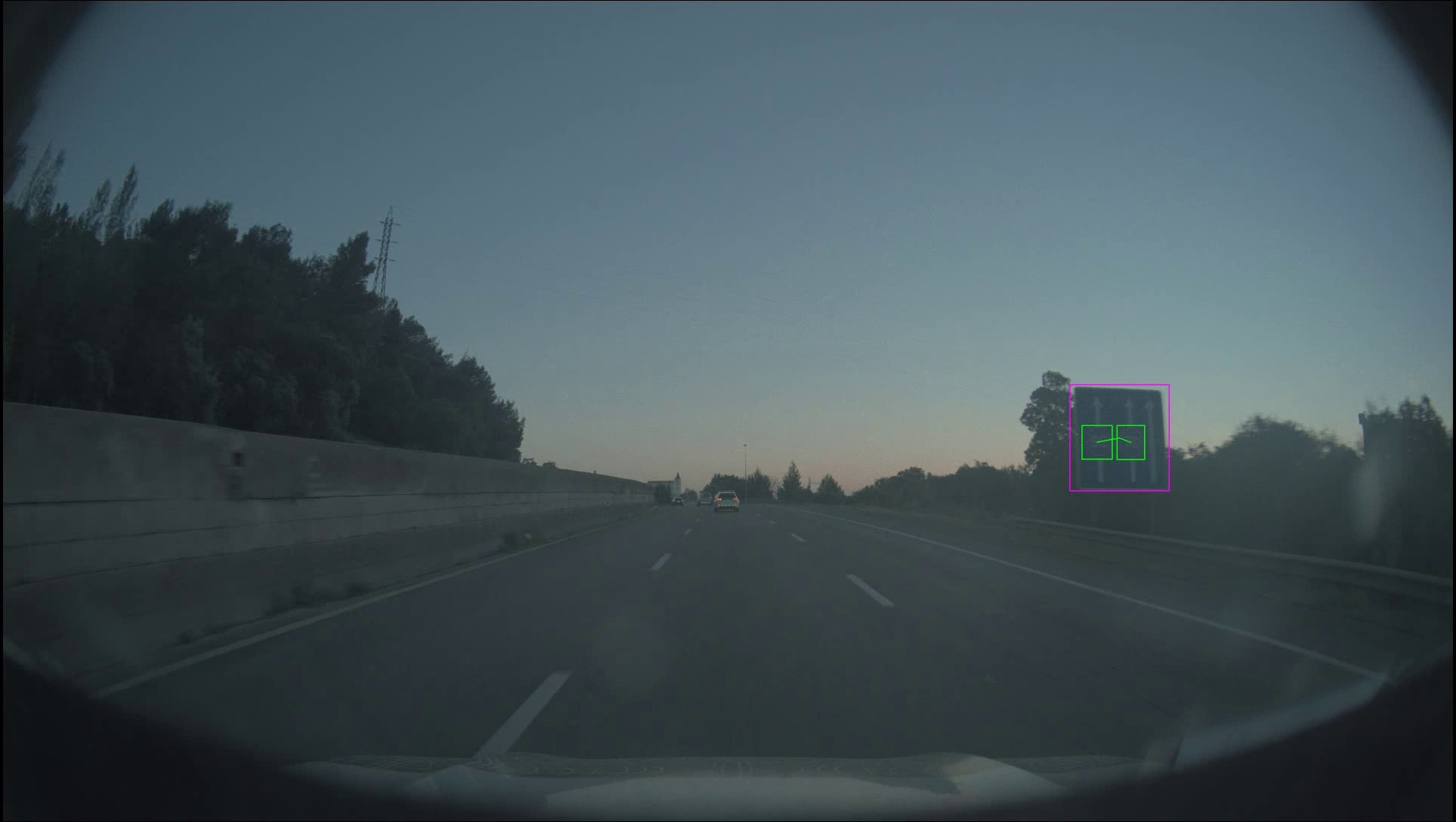}
\includegraphics[width=.48\columnwidth]{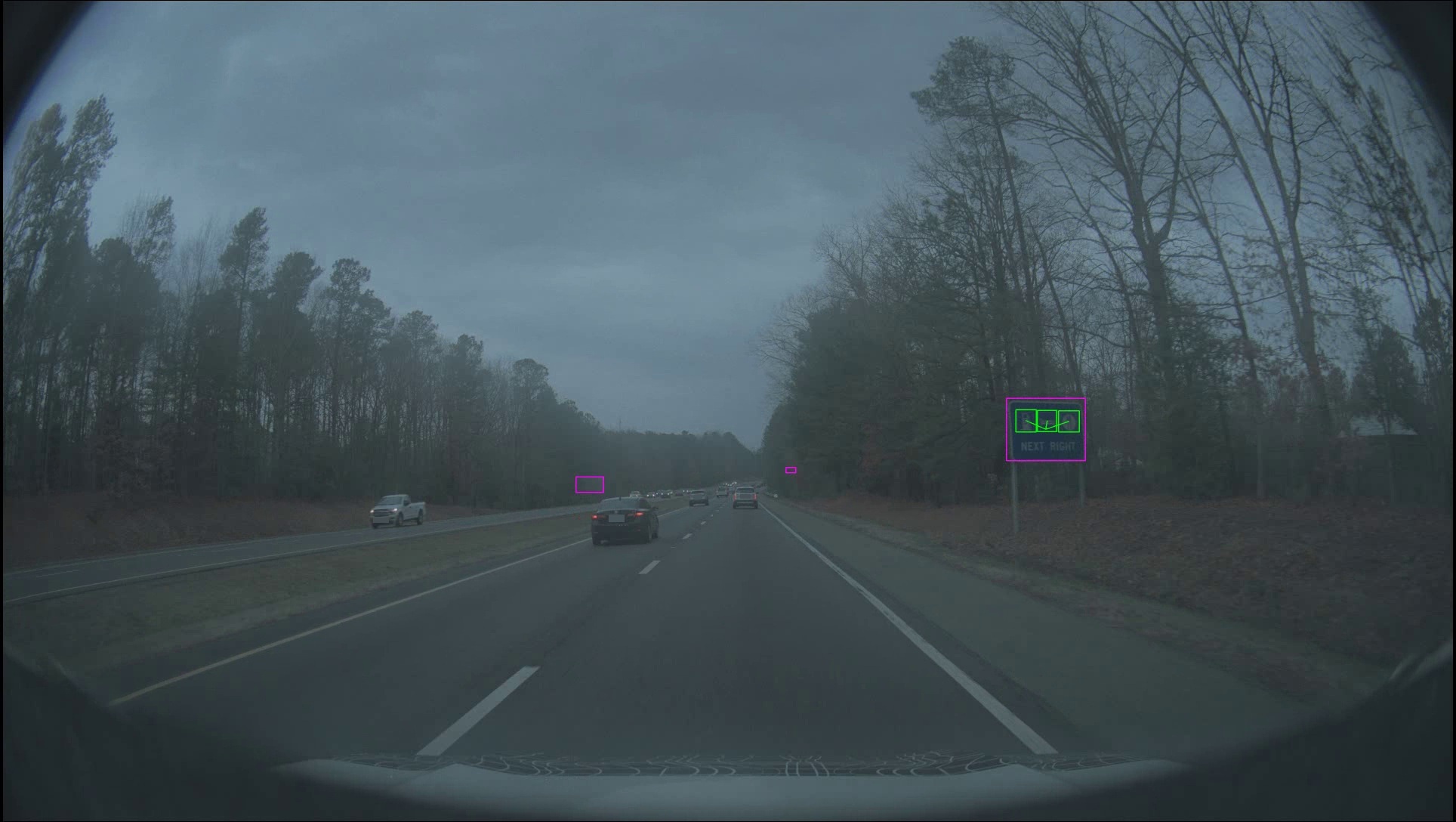}
\\[\smallskipamount]
\includegraphics[width=.48\columnwidth]{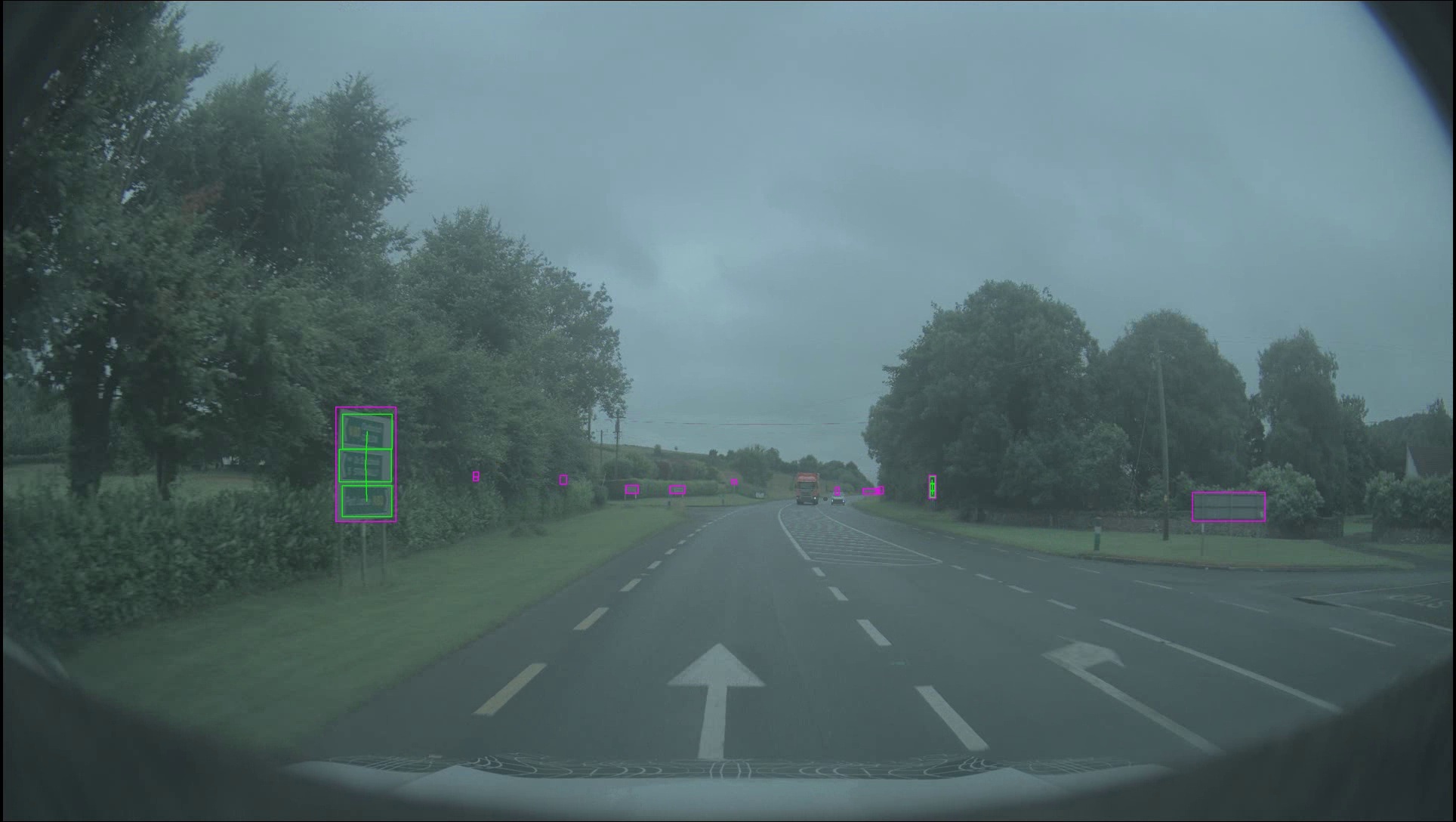}
\includegraphics[width=.48\columnwidth]{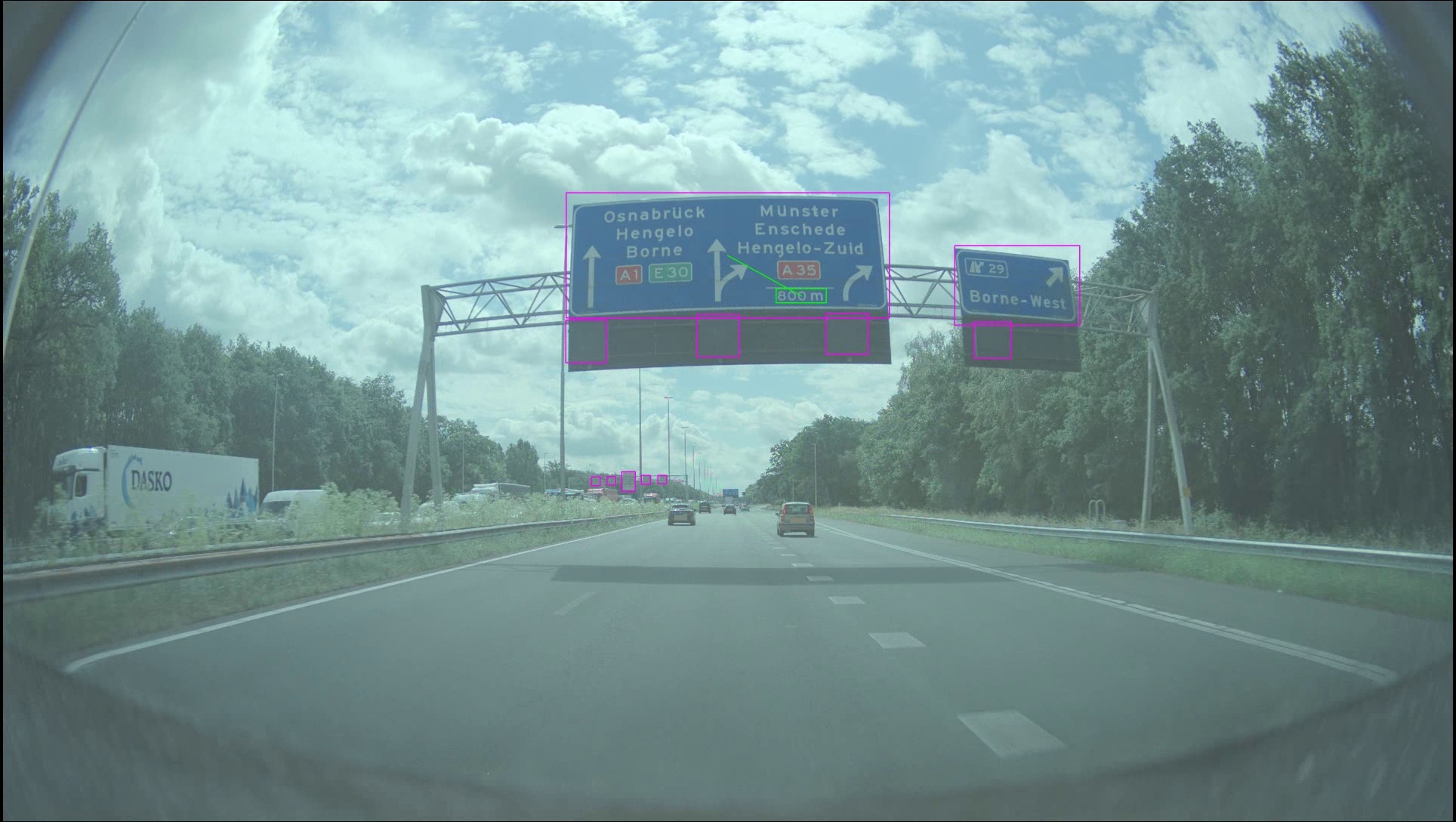}
\caption{Examples of embedded classifications. Green bounding boxes indicate signs classified as embedded; pink bounding boxes indicate the traffic sign that contains them.}\label{fig:embedded_results}
\end{figure}

On the full evaluation set, which contains both embedded and non-embedded traffic signs, the embedded-attribute classifier achieves 0.99 recall and 0.99 precision. Quantitative results are reported in Table~\ref{tab:attribute_accuracy_embedded}, and representative qualitative examples are shown in Figure ~\ref{fig:embedded_results}, while edge cases that the classifier correctly resolves as NotEmbedded (despite visually suggestive perspective overlap) are illustrated in Figure ~\ref{fig:embedded_results_edge_cases}. Given this reliability, we further use the classifier as an automated consistency check for manual annotations, helping identify and correct recurring labeling artifacts such as apparent ``Embedded'' configurations caused by perspective effects or sign placement.

\begin{table}[!t]
    \centering
    \caption{\textbf{Embedded attribute classifier results} using precision and recall metrics on the full evaluation set, including both embedded and non-embedded traffic signs.}    
    \renewcommand{\arraystretch}{1.15}
    \setlength{\tabcolsep}{20pt}
    \small
    \begin{tabular}{lcc}
    \toprule
    \rowcolor{lightgreen}
    \textbf{Attribute} & \textbf{Recall} & \textbf{Precision} \\
    \midrule
    Embedded  & 0.99 & 0.99 \\
    \bottomrule
    \end{tabular}
    \label{tab:attribute_accuracy_embedded}
\end{table}

\begin{table*}[!t]
\centering
    \caption{\textbf{Relevance attribute classifier results} using class-wise precision and recall metrics. \emph{Lane-distance}: a sign is Relevant if its distance to the nearest lane is below 5\,m. \emph{Point-inside~+~lane-distance}: same distance criterion plus the additional constraint that the sign must lie inside a lane-derived polygon. Bold marks the better value per (class, metric) cell across the two methods.}
\renewcommand{\arraystretch}{1.15}
\setlength{\tabcolsep}{18pt}
\small
\begin{tabular}{lccc}
\toprule
\rowcolor{lightgreen}
\textbf{Method} &
\textbf{Class} &
\textbf{Recall} &
\textbf{Precision} \\
\rowcolor{lightgray}
\multicolumn{4}{c}{\textbf{\textit{Lane-distance}}} \\
Lane-distance & Relevant & \textbf{0.89} & 0.54 \\
Lane-distance & NotRelevant & 0.62 & \textbf{0.92} \\
\midrule
\rowcolor{lightgray}
\multicolumn{4}{c}{\textbf{\textit{Lane-distance + point-inside}}} \\
Lane-distance + point-inside & Relevant & \textbf{0.89} & \textbf{0.64} \\
Lane-distance + point-inside & NotRelevant & \textbf{0.67} & 0.90 \\
\bottomrule
\end{tabular}
\label{tab:attribute_relevance}
\end{table*}

\begin{table*}[!t]
    \centering
    \caption{\textbf{3D detection results comparison on the Zenseact public dataset}. Zenseact baseline model results for all traffic sign types is obtained by averaging. Best AP is highlighted in bold.}
    \label{tab:results_zenseact2}
    \renewcommand{\arraystretch}{1.15}
    \setlength{\tabcolsep}{24pt}
    \small
    \begin{tabular}{lccc}
    \toprule
    \rowcolor{lightgreen}
    \textbf{Method} & \textbf{Traffic Sign Type} & \textbf{Modality} & \textbf{AP} \\
    \midrule

    \rowcolor{lightgray}
    \multicolumn{4}{c}{\textbf{\textit{Zenseact baseline model}}} \\
    Zenseact & Front & LiDAR & 0.42 \\
    Zenseact & Back & LiDAR & 0.15  \\
    Zenseact & All & LiDAR & 0.28  \\
    \rowcolor{lightgray}
    \multicolumn{4}{c}{\textbf{\textit{Proposed method}}} \\
    Ours & All & LiDAR & 0.60  \\
    Ours & All & Camera + LiDAR & \textbf{0.64}  \\
    \bottomrule
    \end{tabular}
\end{table*}

\subsection{Relevance Classification} 
\begin{figure}[t]
\centering
\includegraphics[width=.48\columnwidth]{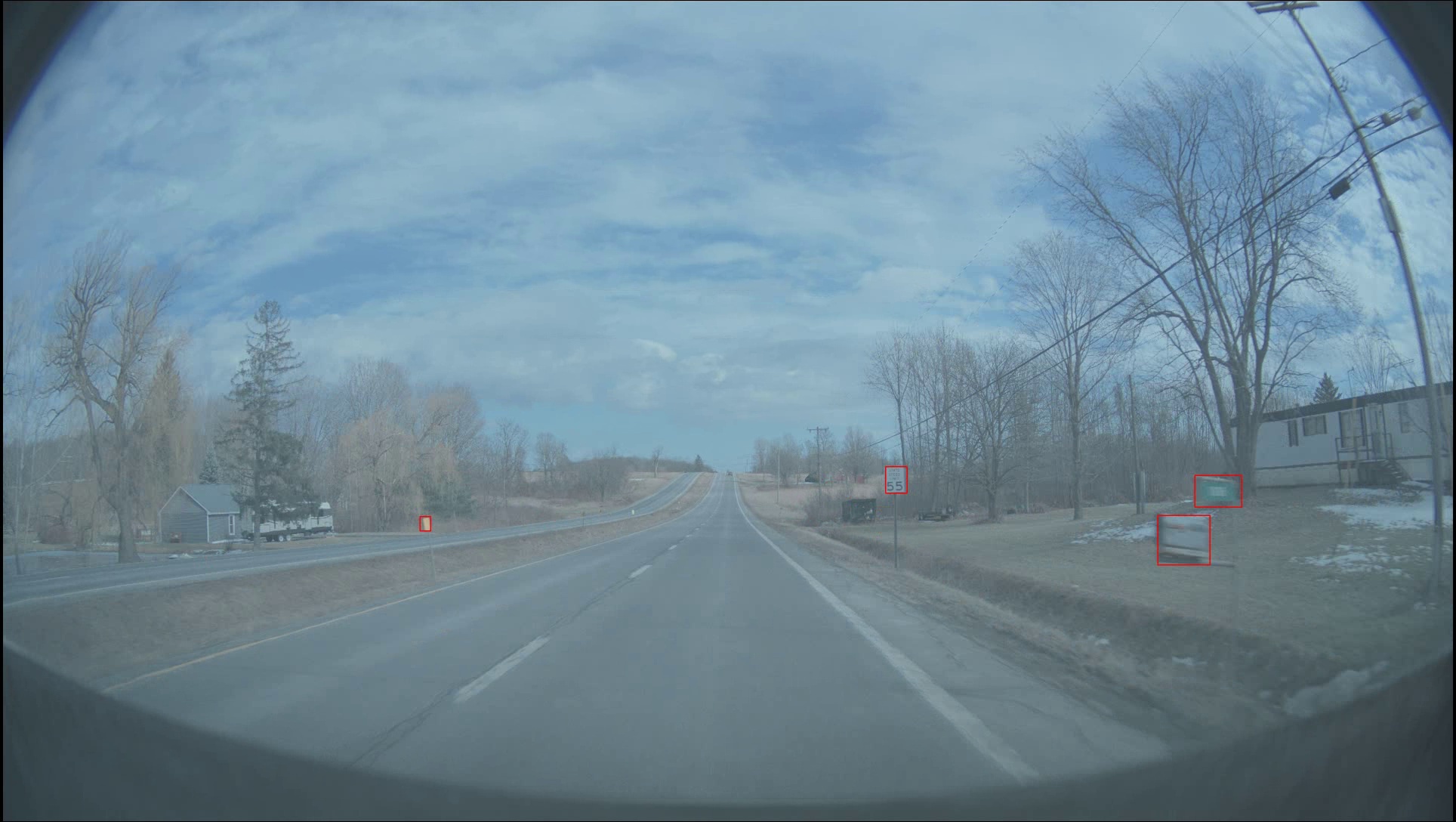}
\includegraphics[width=.48\columnwidth]{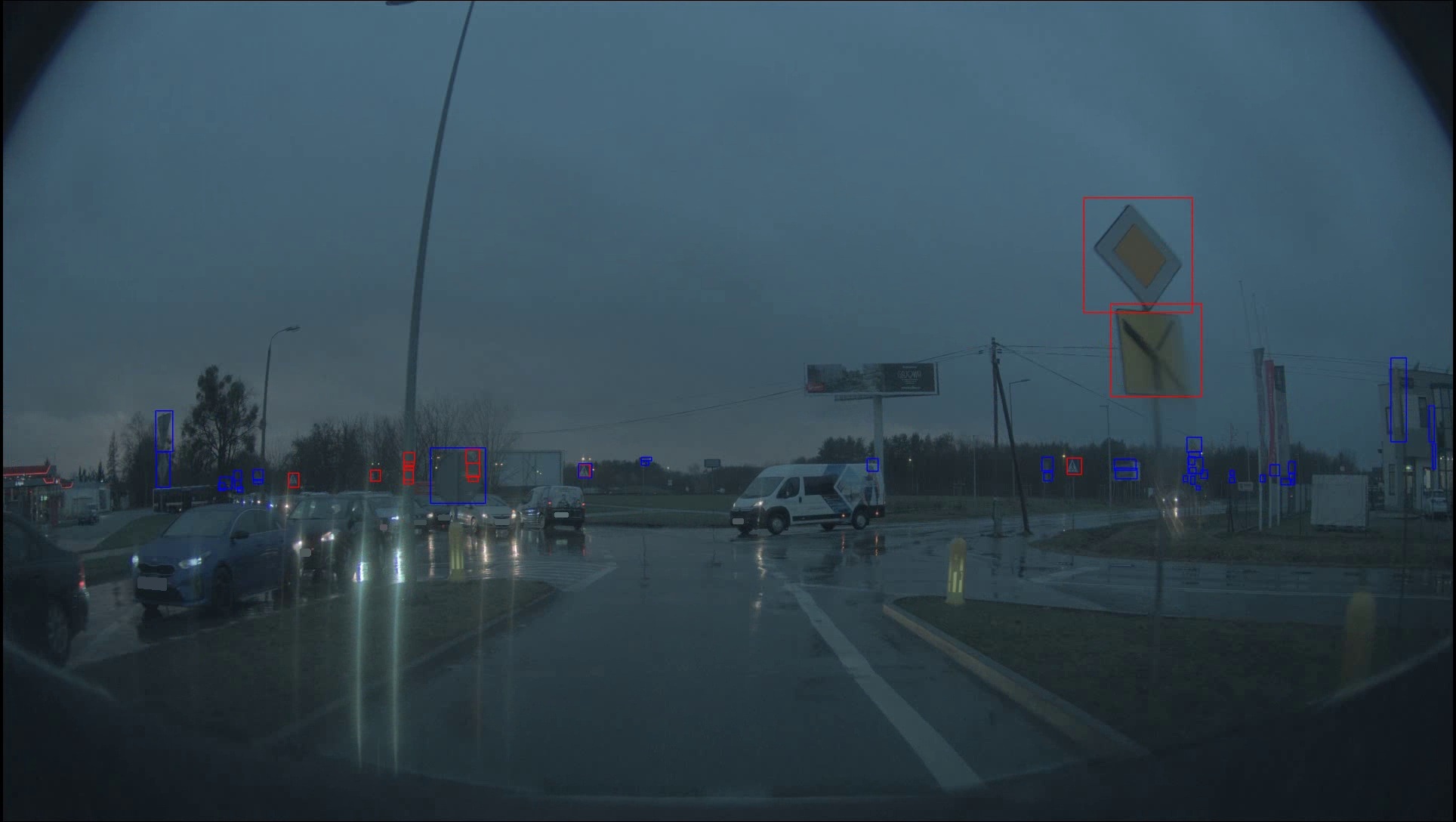}
\\[\smallskipamount]
\includegraphics[width=.48\columnwidth]{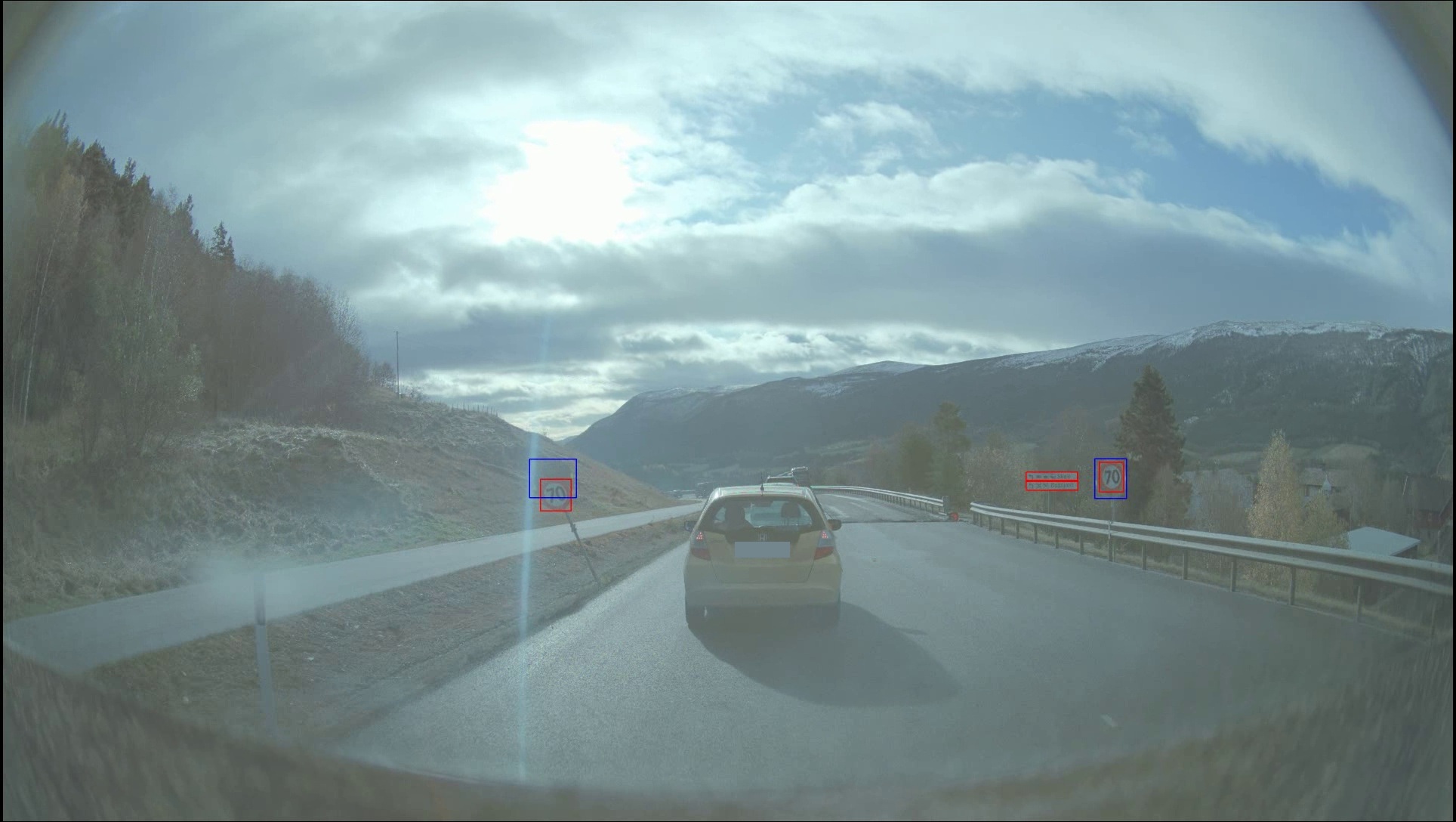}
\includegraphics[width=.48\columnwidth]{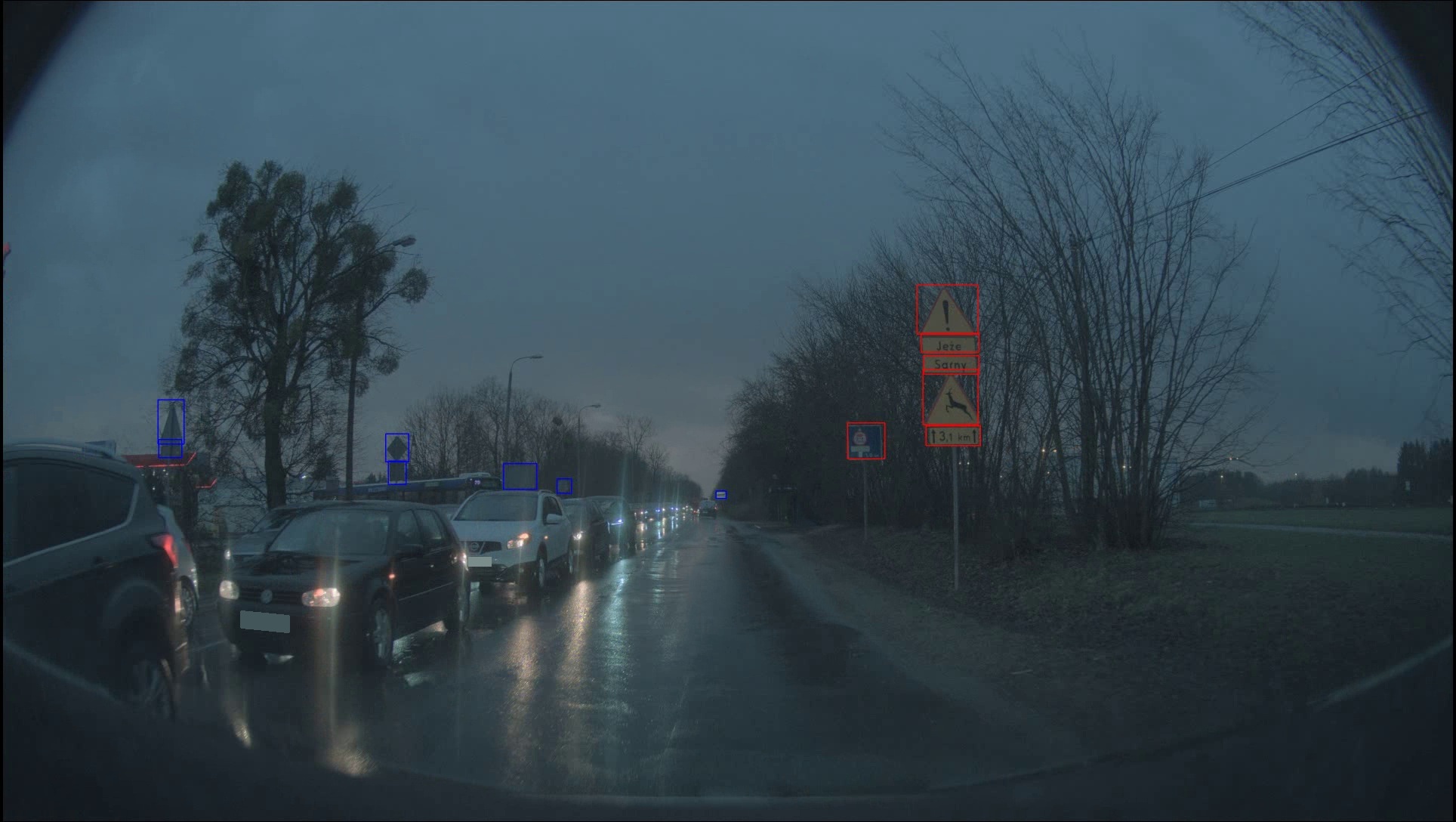}
\caption{Examples of relevance classification. Bounding-box color encodes the predicted class: blue~=~NotRelevant, red~=~Relevant.}\label{fig:relevance}
\end{figure}

We compare two approaches to verify whether a traffic sign is associated with a lane. The first approach checks whether the distance between the sign and the nearest lane is below a fixed threshold of 5\,m, set empirically. The second approach adds one more constraint: a polygon is generated from the lanes, and we additionally verify whether the sign lies inside this polygon. Results are presented in Table~\ref{tab:attribute_relevance}.
At the same 5\,m threshold, adding the point-inside constraint on top of the lane-distance check produces a substantial gain in precision for the positive predictions, denoted as ``Relevant'' in Table~\ref{tab:attribute_relevance}, increasing precision from 0.54 to 0.64. This indicates that the combined rule reduces false-positive ``Relevant'' assignments and is better at rejecting incorrect positives.
Recall remains unchanged, meaning the stricter criterion does not increase the miss rate on true positives.
Overall, accuracy increases from 0.71 to 0.76. We present qualitative results in Figure ~\ref{fig:relevance}.

\subsection{Comparative Evaluation of 3D Detector on Zenseact Dataset}
We compare our 3D detector results with the results reported in~\cite{alibeigi2023zenseact} in Table~\ref{tab:results_zenseact2}. The Zenseact baseline reports AP separately on front-facing and back-of-signs, whereas we report a single AP across all traffic sign types; for direct comparison, in Table~\ref{tab:results_zenseact2} we average the front-facing and back-facing AP values reported by the Zenseact baseline. Our method achieves an improvement of at least 0.22 AP over the LiDAR-only baseline reported in~\cite{alibeigi2023zenseact}, using Centerpoint~\cite{yin2021centerbased3dobjectdetection} as the underlying detector.

\begin{table*}[!t]
    \centering
    \caption{\textbf{3D detection results on the Qualcomm dataset}, comparing single-sweep and multi-sweep LiDAR accumulation. Best AP per range bin is highlighted in bold; the row in blue is our final configuration.}    
    \renewcommand{\arraystretch}{1.15}
    \setlength{\tabcolsep}{15pt}
    \small
    \begin{tabular}{lccccc}
    \toprule
    \rowcolor{lightgreen}
    \textbf{Method} & \textbf{AP} & \textbf{AP$_{{0}-{50}}$} & \textbf{AP$_{{50}-{100}}$}  & \textbf{AP$_{{100}-{150}}$}  & \textbf{AP$_{{150}-{200}}$} \\
    \midrule

    \rowcolor{lightgray}
    \multicolumn{6}{c}{\textbf{\textit{Qualcomm dataset}}} \\
    Single Sweep & 0.55 & 0.65 & 0.52 & 0.34 & 0.14 \\
    Multi Sweeps & \textbf{0.60} & \textbf{0.68} & \textbf{0.58} & \textbf{0.43} & \textbf{0.20} \\
    \bottomrule
    \end{tabular}
    \label{tab:results_qualcomm}
\end{table*}

\begin{table*}[!t]
    \centering
    \caption{\textbf{Ablation of the LiDAR contribution in the 2D detector on the Qualcomm dataset.} The Camera-only baseline is the same Qualcomm-trained backbone with no LiDAR input; ``+ LiDAR'' adds the depth and intensity maps fused via the Intensity-Aware Deformable Fusion module described in Section~\ref{sec:proposeddetectorandtracker}. Best value per metric is in bold.}    
    \renewcommand{\arraystretch}{1.15}
    \setlength{\tabcolsep}{13pt}
    \small
    \begin{tabular}{lcccccc}
    \toprule
    \rowcolor{lightgreen}
    \textbf{Method} & \textbf{AP} & \textbf{AP$_{50}$} & \textbf{AP$_{75}$} & \textbf{AP$_s$} & \textbf{AP$_m$} & \textbf{AP$_l$} \\
    \midrule
    \rowcolor{lightgray}
    \multicolumn{7}{c}{\textbf{\textit{Qualcomm dataset}}} \\
    \makecell[l]{Camera-only baseline}
    &  0.62 & 0.83 & 0.69 & 0.46 & 0.77 & 0.89 \\
    \makecell[l]{Camera and LiDAR Fusion}
    & \textbf{0.65} & \textbf{0.86} & \textbf{0.73} & \textbf{0.51} & \textbf{0.80} & \textbf{0.90} \\
    \bottomrule
    \end{tabular}
    \label{tab:results_2ddetector_ablations}
\end{table*}

\begin{figure*}[!t]
\centering
    \includegraphics[width=.32\textwidth]{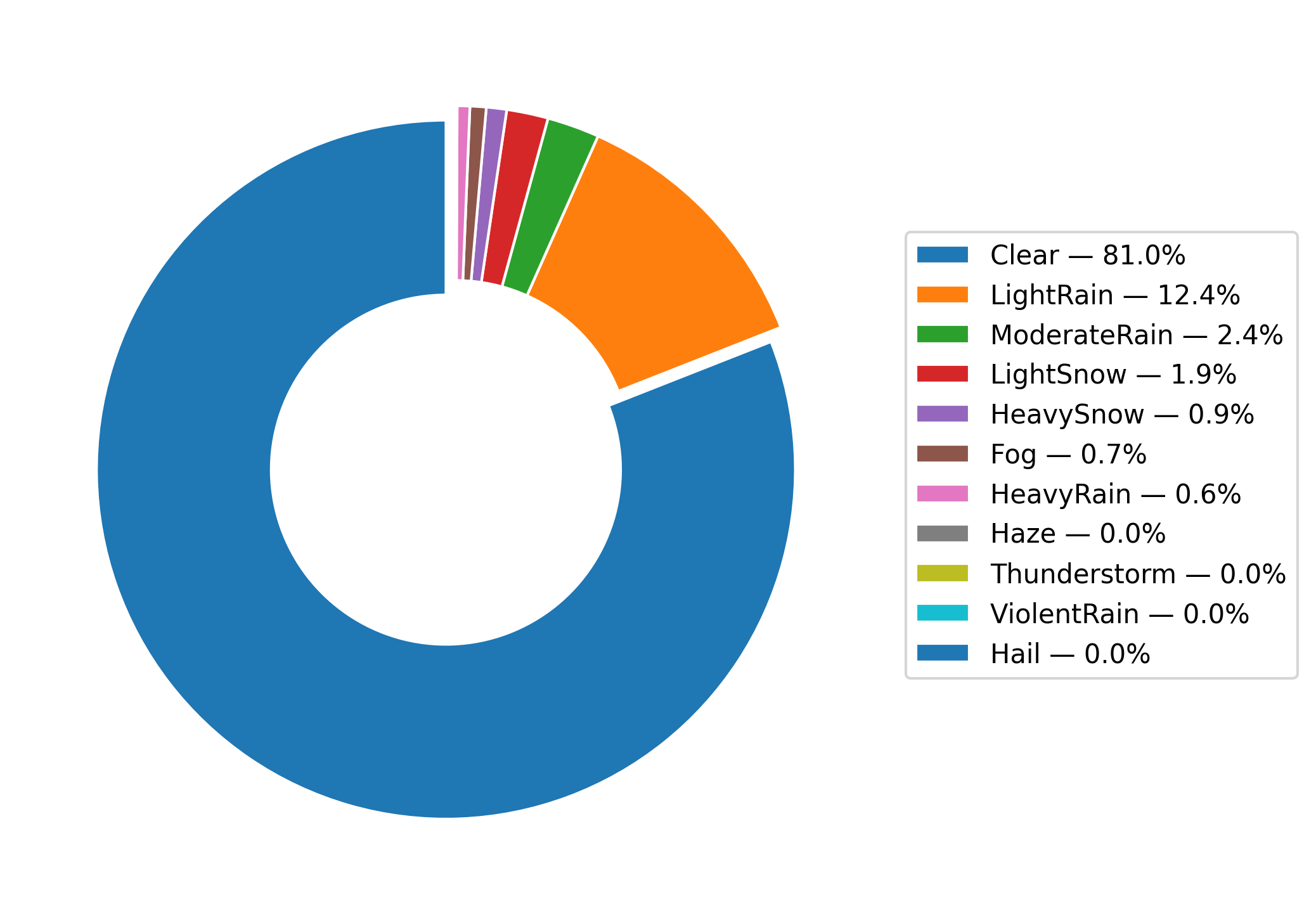}
    \includegraphics[width=.32\textwidth]{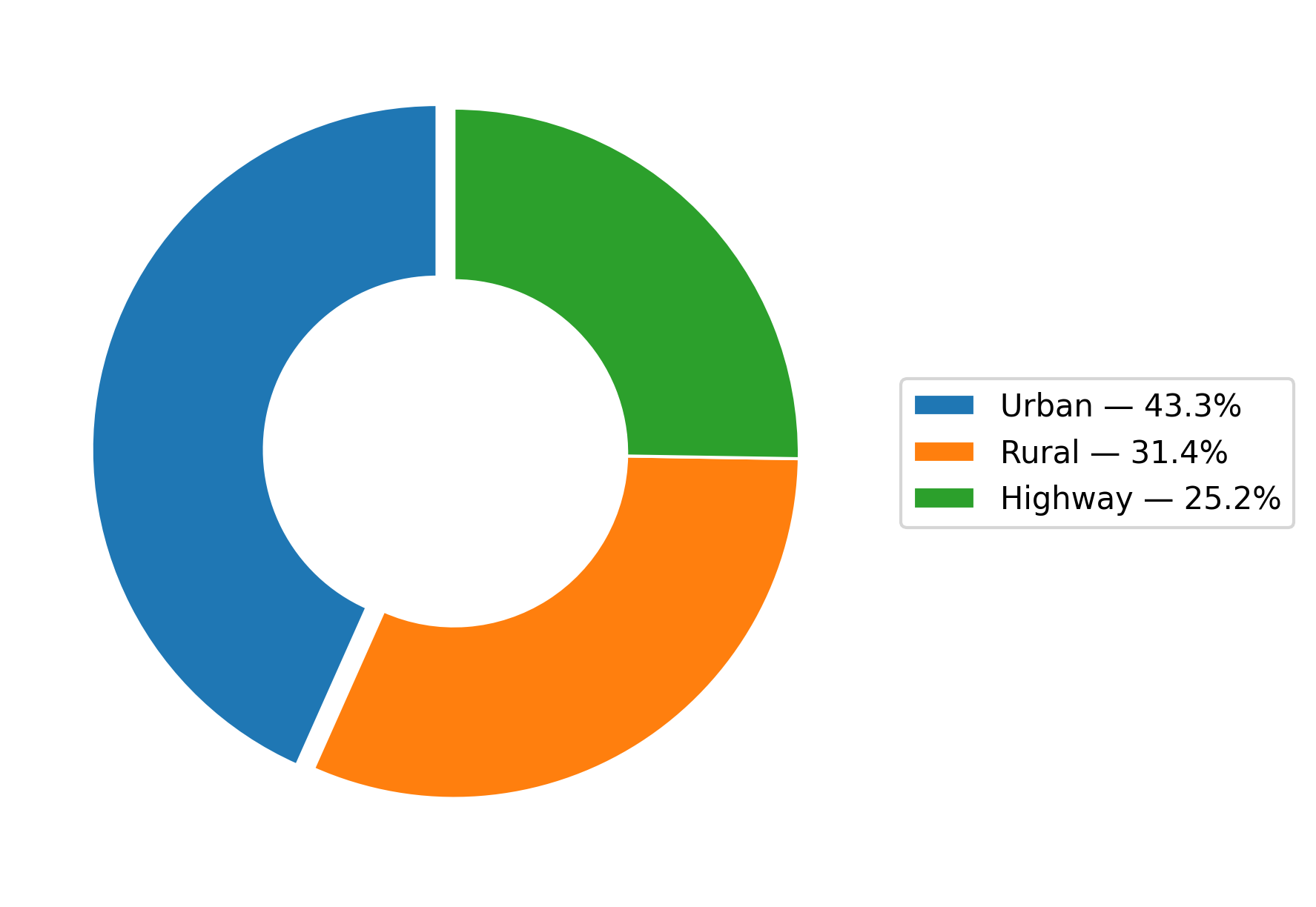}
    \includegraphics[width=.32\textwidth]{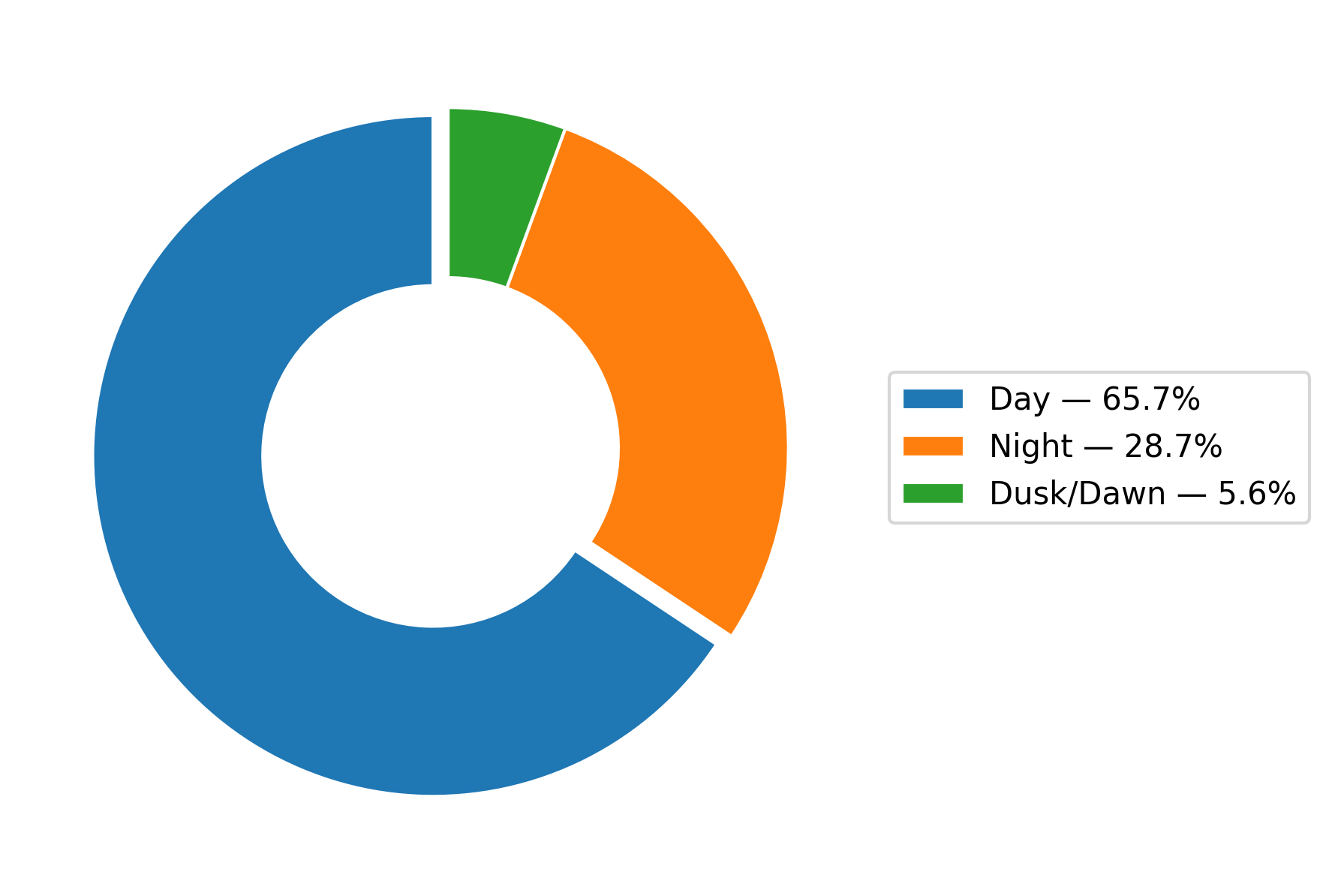}
\caption{Distribution of dataset metadata tags: (left) weather, (center) road type, (right) time of day.}\label{fig:image_dataset}
\end{figure*}

On the Qualcomm dataset, we train the 3D detector in two configurations: as a single-sweep configuration and a multi-sweep configuration. For the multi-sweep training we use 10 point cloud sweeps from the future and aggregate the point clouds to generate an aggregated point cloud. To further evaluate the impact of temporal density on long-range localization, Table~\ref{tab:results_qualcomm} presents an ablation study comparing single-sweep and multi-sweep LiDAR accumulation on the Qualcomm dataset. The results demonstrate that temporal aggregation is essential for distant objects, with the multi-sweep configuration improving the overall AP from 0.55 to 0.60. At the 150--200\,m range, the multi-sweep model achieves considerably higher AP (0.20 vs.\ 0.14). Despite these gains, the detector exhibits a performance decay as the distance increases, with AP dropping from 0.68 at close range to just 0.20 at the 200\,m boundary. This decay highlights the inherent difficulty of regressing precise bounding boxes from sparse point returns at extreme ranges when relying on 3D geometric features alone. The performance decay observed in the 3D distance bins of Table~\ref{tab:results_qualcomm} is directly correlated with the scale-based metrics in the 2D image plane of Table~\ref{tab:results_merged}. Physically, traffic signs at a 200\,m range typically occupy a footprint of only $10\times10$ pixels, falling within the ``Small'' (AP$_s$) category defined by the COCO protocol ($<32^2$ pixels).

We note that absolute AP values differ between Tables~\ref{tab:results_zenseact2} (Zenseact) and~\ref{tab:results_qualcomm} (Qualcomm) because the two datasets have different geographic coverage, sign distributions, and annotation density; the Zenseact-trained model is trained on a smaller, geographically narrower distribution, whereas the Qualcomm-trained model benefits from a substantially larger and more diverse training set.

\subsection{Comparison of  Camera-Only and Camera+LiDAR 2D Detector}
In our proposed 2D detector, incorporating LiDAR-derived depth and intensity maps improves detector performance. The detector continues to operate primarily in the image domain, where distant traffic signs remain represented by dense visual features, while LiDAR contributes complementary geometric and reflectivity cues. This is beneficial for traffic-sign detection since camera imagery preserves higher spatial detail at long range, whereas LiDAR point clouds become increasingly sparse with distance.

\begin{table}[!t]
    \centering
    \caption{\textbf{Object Miss Ratio (OMR) breakdown by time of day, road type,
             and weather condition} on the Qualcomm
             dataset (lower is better). Rare weather categories
             (Thunderstorm, ViolentRain, Hail) contain very few sequences
             and should be treated as indicative rather than statistically
             conclusive.}
    \label{tab:omr_breakdown}
    \renewcommand{\arraystretch}{1.15}
    \setlength{\tabcolsep}{16pt}
    \small
    \begin{tabular}{l r r}
    \toprule
    \rowcolor{lightgreen}
    \textbf{Category} &
    \textbf{Sequences} &
    \textbf{Object Miss Ratio} \\
    \midrule
    \rowcolor{lightgray}
    \multicolumn{3}{l}{\textbf{\textit{Time of Day}}} \\
    Day            & 145{,}412 & \textbf{0.67\%} \\
    Night          &  63{,}641 & \textbf{0.17\%} \\
    Dusk/Dawn      &  12{,}395 & \textbf{0.36\%} \\
    \midrule
    \rowcolor{lightgray}
    \multicolumn{3}{l}{\textbf{\textit{Road Type}}} \\
    Urban          &  95{,}806 & \textbf{0.32\%} \\
    Rural          &  69{,}443 & \textbf{0.66\%} \\
    Highway        &  55{,}819 & \textbf{1.16\%} \\
    \midrule
    \rowcolor{lightgray}
    \multicolumn{3}{l}{\textbf{\textit{Weather Condition}}} \\
    Clear          & 176{,}238 & \textbf{0.52\%} \\
    LightRain      &  26{,}898 & \textbf{0.38\%} \\
    ModerateRain   &   5{,}265 & \textbf{0.34\%} \\
    LightSnow      &   4{,}182 & \textbf{0.36\%} \\
    HeavySnow      &   2{,}046 & \textbf{0.27\%} \\
    Fog            &   1{,}629 & \textbf{0.72\%} \\
    HeavyRain      &   1{,}271 & \textbf{0.44\%} \\
    Haze           &        98 & \textbf{0.17\%} \\
    Thunderstorm   &        27 & \textbf{0.00\%} \\
    ViolentRain    &        11 & \textbf{0.29\%} \\
    Hail           &         1 & \textbf{0.00\%} \\
    \midrule
    \rowcolor{lightgray}
    \textbf{TOTAL} & \textbf{221{,}068} & \textbf{0.49\%} \\
    \bottomrule
    \end{tabular}
\end{table}

LiDAR depth provides a useful geometric prior for small and distant signs by constraining whether a candidate region is located at a plausible roadside distance and elevation, even when the LiDAR observations are too sparse to support stable 3D box estimation on their own. In addition, LiDAR intensity contributes reflectivity information that is especially relevant for traffic signs, which are designed to be reflective and therefore often produce distinctive return patterns relative to the background. As a result, in our 2D detector, depth and intensity act as lightweight auxiliary cues that improve feature discrimination without requiring the model to reconstruct precise long-range 3D structure. As seen in Table~\ref{tab:results_2ddetector_ablations}, the relative gain is most pronounced for small objects ($\text{AP}_s$: $+0.04$, from 0.47 to 0.51), with smaller absolute improvements for medium ($+0.02$) and large ($+0.01$) objects. LiDAR geometric and reflectivity cues are most beneficial for distant, pixel-sparse traffic signs that are challenging for the camera alone.

\subsection{Evaluation of effect of weather, lighting condition and road type}   
To evaluate the generalization capabilities of our model, we conduct extensive testing on our Qualcomm dataset to capture the ``long-tail'' variability of global driving. Unlike open-source benchmarks restricted to specific domains, our dataset accounts for the significant inter-country variability in traffic infrastructure, including varied sign shapes, LED-based displays, and multi-lingual content across the 60+ countries described in Section~\ref{sec:materialsandmethods}. As seen in Fig.~\ref{fig:image_dataset}, our dataset consists of different scenarios in diverse weather conditions, road types, and illumination.

We present our results on manually verified data. The verification process uses a human-in-the-loop methodology, in which each annotation is reviewed and corrected by a trained annotator. We report performance using the Object Miss Ratio defined in Section~\ref{sec:materialsandmethods}, which takes values in $[0\%, 100\%]$, where $0\%$ corresponds to no missed objects relative to the manually verified set and $100\%$ corresponds to a complete miss. Lower values are better.

In Table~\ref{tab:omr_breakdown} we evaluate the robustness of the system under varying illumination conditions. Assessing performance across these diverse conditions is critical to ensure that the labels used for training do not systematically favor particular lighting regimes. Overall, the pipeline exhibits strong temporal stability, achieving a total Object Miss Ratio of 0.49\% across the 221{,}068 evaluation sequences, which correspond to over 2{,}500 hours of driving data. The system reports its lowest miss ratio at Night (0.17\%), compared to 0.67\% during the Day. Two factors plausibly explain the favorable Night number. First, traffic signs are retroreflective; under headlight illumination they appear with high contrast against dark backgrounds, simplifying detection. Second, lower average ego-speeds at night yield more frames per sign, giving the tracker more opportunities to consolidate associations. Conversely, daytime sequences constitute the majority of the dataset (145{,}412 sequences) and exhibit the highest miss ratio (0.67\%); we attribute this primarily to increased visual complexity, where strong ambient lighting reveals detailed backgrounds that can partially camouflage signs, especially those affected by wear or placed within dense urban structures, combined with higher average driving speeds that shorten the per-sign observation window. We note that manually verifying the existence of far-away signs is also more difficult in night, dusk, and dawn scenes; this could in principle inflate true miss rates at low light, so the absolute Night number should be interpreted as a lower bound rather than a precise estimate.

In Table~\ref{tab:omr_breakdown}, we present the results separated by different road types. The pipeline achieves its best performance in \textbf{Urban} sequences, with a low Object Miss Ratio. We attribute this to the higher temporal redundancy typical of urban driving: lower ego-speeds increase the number of frames in which a traffic sign remains observable, giving the tracker more opportunities to maintain stable associations and recover detections that may be briefly degraded by occlusion or illumination changes. Additionally, Urban scenes contain more objects than rural and highway scenes, which means that one missing sign would not contribute as much to the metric. Performance degrades as driving speed and long-range viewing conditions become more dominant, with miss ratios increasing to 0.66\% in \textbf{Rural} and 1.16\% on \textbf{Highway}.
This trend is consistent with the known difficulties of high-speed perception, such as reduced observation time at long range, increased sensitivity to small-object appearance at distance, and higher-frequency vehicle dynamics that can momentarily destabilize temporal association and cross-sensor consistency.

We measured the Object Miss Ratio of the system on eleven weather categories, as shown in Table~\ref{tab:omr_breakdown}. The Moderate Rain (0.34\%) and Heavy Snow (0.27\%) conditions yield lower miss ratios than Clear weather (0.52\%). We attribute this to reduced ego-vehicle speeds in adverse weather, which increases the number of frames per object and improves temporal consolidation in the tracker. Fog is the most challenging condition, with the highest miss ratio (0.72\%). Unlike precipitation, fog degrades both sensing modalities: camera imagery suffers from range reduction and contrast loss, while LiDAR experiences backscatter from droplets, increasing point-cloud noise. The dataset is dominated by Clear-weather driving (176{,}238 sequences), providing a strong baseline for training and evaluation. We also include rare ``long-tail'' weather events such as thunderstorms, violent rain, and hail. We note that several of these rare categories contain only a small number of sequences (27 thunderstorm and 1 hail sequence), so the corresponding miss ratios should be interpreted as indicative rather than statistically conclusive. 

%% file: ieee_sections/sec_conclusions.tex
\section{Conclusions}
\label{sec:conclusions}

In this paper, we introduce a multi-modal framework for traffic-sign detection and tracking that addresses the challenges of regional generalization, long-range detection, and temporal stability. By fusing LiDAR depth and intensity with camera features, our system achieves region-invariant performance across the 60+ countries covered by our dataset. The dual motion-model Kalman filter explicitly handles the non-linear perspective transformations encountered during vehicle approach, yielding incremental Recall and Precision improvements over a constant-velocity baseline. Beyond localization, we developed a semantic attribute classification pipeline that extracts operational context, namely occlusion, readability, embeddedness, and relevance, providing a filtering mechanism for downstream autonomous driving tasks. Validation on the Qualcomm dataset, spanning 60+ countries and over 2{,}500 hours of driving data, demonstrates the scalability and robustness of our approach: the proposed pipeline achieves a total Object Miss Ratio (OMR) of 0.49\% across 221{,}068 evaluation sequences, with strong stability across illumination, road type, and weather regimes. These results establish a foundation for globally generalizable perception in autonomous driving systems.



\vspace{6pt} 

%% file: root.bib
@book{joseph2021autonomous,
  title={Autonomous driving and advanced driver-assistance systems (ADAS): applications, development, legal issues, and testing},
  author={Joseph, Lentin and Mondal, Amit Kumar},
  year={2021},
  publisher={CRC Press}
}

@inproceedings{sistu2019neurall,
  title={Neurall: Towards a unified visual perception model for automated driving},
  author={Sistu, Ganesh and Leang, Isabelle and Chennupati, Sumanth and Yogamani, Senthil and Hughes, Ciar{\'a}n and Milz, Stefan and Rawashdeh, Samir},
  booktitle={2019 IEEE Intelligent Transportation Systems Conference (ITSC)},
  pages={796--803},
  year={2019},
  organization={IEEE}
}

@conference{Chennupativisapp19,
    author={Chennupati, S. and Ganesh Sistu. and Senthil Yogamani. and Samir Rawashdeh.},
    title={AuxNet: Auxiliary Tasks Enhanced Semantic Segmentation for Automated Driving},
    booktitle={Proceedings of the 14th International Joint Conference on Computer Vision, Imaging and Computer Graphics Theory and Applications - Volume 5 VISAPP,},
    year={2019},
    pages={645-652},
    publisher={SciTePress}
}

@conference{uricar2019challenges,
  title={Challenges in designing datasets and validation for autonomous driving},
  author={Uric{\'a}r, Michal and Hurych, David and Krizek, Pavel and Yogamani, Senthil},
booktitle={Proceedings of the International Joint Conference on Computer Vision, Imaging and Computer Graphics Theory and Applications (VISAPP)},
year={2019}
}

@article{yahiaoui2019overview,
  title={Overview and empirical analysis of isp parameter tuning for visual perception in autonomous driving},
  author={Yahiaoui, Lucie and Horgan, Jonathan and Deegan, Brian and Yogamani, Senthil and Hughes, Ciar{\'a}n and Denny, Patrick},
  journal={Journal of Imaging},
  volume={5},
  number={10},
  pages={78},
  year={2019},
  publisher={MDPI}
}

@inproceedings{kumar2018near,
  title={Near-field depth estimation using monocular fisheye camera: A semi-supervised learning approach using sparse LiDAR data},
  author={Kumar, Varun Ravi and Milz, Stefan and Witt, Christian and Simon, Martin and Amende, Karl and Petzold, Johannes and Yogamani, Senthil and Pech, Timo},
  booktitle={CVPR Workshop},
  volume={7},
  pages={2},
  year={2018}
}

@article{wang2023improved,
  title={Improved YOLOv5 network for real-time multi-scale traffic sign detection},
  author={Wang, Junfan and Chen, Yi and Dong, Zhekang and Gao, Mingyu},
  journal={Neural Computing and Applications},
  volume={35},
  number={10},
  pages={7853--7865},
  year={2023},
  publisher={Springer}
}

@article{zhang2025lr,
  title={LR-DETR: a lightweight real-time traffic sign detection model based on improved RT-DETR},
  author={Zhang, Longzhen and Wang, Mingyang and Zhao, Xianhao and Wang, Xianjie},
  journal={Journal of Real-Time Image Processing},
  volume={22},
  number={2},
  pages={1--16},
  year={2025},
  publisher={Springer}
}

@article{zhang2024learning,
  title={Learning multi-layer interactive residual feature fusion network for real-time traffic sign detection with stage routing attention},
  author={Zhang, Jianming and Yi, Yao and Wang, Zulou and Alqahtani, Fayez and Wang, Jin},
  journal={Journal of Real-Time Image Processing},
  volume={21},
  number={5},
  pages={176},
  year={2024},
  publisher={Springer}
}

@article{liu2024clfnet,
  title={CLFNet: a multi-modal data fusion network for traffic sign extraction},
  author={Liu, Rufei and Su, Zhanwen and Zhang, Yi and Li, Ming},
  journal={Measurement Science and Technology},
  volume={36},
  number={1},
  pages={015131},
  year={2024},
  publisher={IOP Publishing}
}

@article{zhu2022traffic,
  title={Traffic sign recognition based on deep learning},
  author={Zhu, Yanzhao and Yan, Wei Qi},
  journal={Multimedia Tools and Applications},
  volume={81},
  number={13},
  pages={17779--17791},
  year={2022},
  publisher={Springer}
}

@article{wali2019vision,
  title={Vision-based traffic sign detection and recognition systems: Current trends and challenges},
  author={Wali, Safat B and Abdullah, Majid A and Hannan, Mahammad A and Hussain, Aini and Samad, Salina A and Ker, Pin J and Mansor, Muhamad Bin},
  journal={Sensors},
  volume={19},
  number={9},
  pages={2093},
  year={2019},
  publisher={MDPI}
}

@article{temel2019traffic,
  title={Traffic sign detection under challenging conditions: A deeper look into performance variations and spectral characteristics},
  author={Temel, Dogancan and Chen, Min-Hung and AlRegib, Ghassan},
  journal={IEEE Transactions on Intelligent Transportation Systems},
  volume={21},
  number={9},
  pages={3663--3673},
  year={2019},
  publisher={IEEE}
}

@misc{michaelis2020benchmarkingrobustnessobjectdetection,
      title={Benchmarking Robustness in Object Detection: Autonomous Driving when Winter is Coming}, 
      author={Claudio Michaelis and Benjamin Mitzkus and Robert Geirhos and Evgenia Rusak and Oliver Bringmann and Alexander S. Ecker and Matthias Bethge and Wieland Brendel},
      year={2020},
      eprint={1907.07484},
      archivePrefix={arXiv},
      primaryClass={cs.CV},
      url={https://arxiv.org/abs/1907.07484}, 
}

@article{wang2023vehicle,
  title={Vehicle-mounted adaptive traffic sign detector for small-sized signs in multiple working conditions},
  author={Wang, Junfan and Chen, Yi and Ji, Xiaoyue and Dong, Zhekang and Gao, Mingyu and Lai, Chun Sing},
  journal={IEEE Transactions on Intelligent Transportation Systems},
  volume={25},
  number={1},
  pages={710--724},
  year={2023},
  publisher={IEEE}
}

@article{suresha2024recent,
  title={Recent advancement in small traffic sign detection: approaches and dataset},
  author={Suresha, R and Manohar, N and Kumar, G Ajay and Singh, M Rohit},
  journal={IEEE Access},
  volume={12},
  pages={192840--192859},
  year={2024},
  publisher={IEEE}
}

@inproceedings{Stallkamp-IJCNN-2011,
    author = {Johannes Stallkamp and Marc Schlipsing and Jan Salmen and Christian Igel},
    booktitle = {IEEE International Joint Conference on Neural Networks},
    title = {The {G}erman {T}raffic {S}ign {R}ecognition {B}enchmark: A multi-class classification competition},
    year = {2011},
    pages = {1453--1460}
}

@InProceedings{Zhe_2016_CVPR,
    author = {Zhu, Zhe and Liang, Dun and Zhang, Songhai and Huang, Xiaolei and Li, Baoli and Hu, Shimin},
    title = {Traffic-Sign Detection and Classification in the Wild},
    booktitle = {The IEEE Conference on Computer Vision and Pattern Recognition (CVPR)},
    year = {2016}
}

@InProceedings{Timofte-WACV-2009,
	author =       {Timofte, Radu and Zimmermann, Karel and van Gool, Luc},
	title =        {Multi-view traffic sign detection, recognition, and 3D localisation},
	booktitle =    {Ninth IEEE Computer Society Workshop on Application of Computer Vision},
	project =      {IBBT-URBAN},
	address =      {Snowbird, Utah, USA},
	pages =        {1-8},
	month =        {December},
	day =          {7-8},
	year =         {2009},
	isbn =         {978-1-4244-5496-9},
	issn =         {1550-5790},
	book_pages =   {69-76},
}

@article{babic2021analysis,
  title={Analysis of market-ready traffic sign recognition systems in cars: a test field study},
  author={Babi{\'c}, Darko and Babi{\'c}, Dario and Fioli{\'c}, Mario and {\v{S}}ari{\'c}, {\v{Z}}eljko},
  journal={Energies},
  volume={14},
  number={12},
  pages={3697},
  year={2021},
  publisher={MDPI}
}

@InProceedings{bdd100k,
    author = {Yu, Fisher and Chen, Haofeng and Wang, Xin and Xian, Wenqi and Chen,
              Yingying and Liu, Fangchen and Madhavan, Vashisht and Darrell, Trevor},
    title = {BDD100K: A Diverse Driving Dataset for Heterogeneous Multitask Learning},
    booktitle = {IEEE/CVF Conference on Computer Vision and Pattern Recognition (CVPR)},
    month = {June},
    year = {2020}
}

@misc{ertler2020mapillarytrafficsigndataset,
      title={The Mapillary Traffic Sign Dataset for Detection and Classification on a Global Scale}, 
      author={Christian Ertler and Jerneja Mislej and Tobias Ollmann and Lorenzo Porzi and Gerhard Neuhold and Yubin Kuang},
      year={2020},
      eprint={1909.04422},
      archivePrefix={arXiv},
      primaryClass={cs.CV},
      url={https://arxiv.org/abs/1909.04422}, 
}

@inproceedings{alibeigi2023zenseact,
  title={Zenseact open dataset: A large-scale and diverse multimodal dataset for autonomous driving},
  author={Alibeigi, Mina and Ljungbergh, William and Tonderski, Adam and Hess, Georg and Lilja, Adam and Lindstr{\"o}m, Carl and Motorniuk, Daria and Fu, Junsheng and Widahl, Jenny and Petersson, Christoffer},
  booktitle={Proceedings of the IEEE/CVF International Conference on Computer Vision},
  pages={20178--20188},
  year={2023}
}

@article{sobh2021adversarial,
  title={Adversarial attacks on multi-task visual perception for autonomous driving},
  author={Sobh, Ibrahim and Hamed, Ahmed and Kumar, Varun Ravi and Yogamani, Senthil},
  journal={arXiv preprint arXiv:2107.07449},
  year={2021}
}

@inproceedings{schramm2024bevcar,
  title={Bevcar: Camera-radar fusion for bev map and object segmentation},
  author={Schramm, Jonas and V{\"o}disch, Niclas and Petek, K{\"u}rsat and Kiran, B Ravi and Yogamani, Senthil and Burgard, Wolfram and Valada, Abhinav},
  booktitle={2024 IEEE/RSJ International Conference on Intelligent Robots and Systems (IROS)},
  pages={1435--1442},
  year={2024},
  organization={IEEE}
}

@ARTICLE{9000872,
  author={Feng, Di and Haase-Schütz, Christian and Rosenbaum, Lars and Hertlein, Heinz and Gläser, Claudius and Timm, Fabian and Wiesbeck, Werner and Dietmayer, Klaus},
  journal={IEEE Transactions on Intelligent Transportation Systems}, 
  title={Deep Multi-Modal Object Detection and Semantic Segmentation for Autonomous Driving: Datasets, Methods, and Challenges}, 
  year={2021},
  volume={22},
  number={3},
  pages={1341-1360},
  doi={10.1109/TITS.2020.2972974}
}

@misc{vora2020pointpaintingsequentialfusion3d,
      title={PointPainting: Sequential Fusion for 3D Object Detection}, 
      author={Sourabh Vora and Alex H. Lang and Bassam Helou and Oscar Beijbom},
      year={2020},
      eprint={1911.10150},
      archivePrefix={arXiv},
      primaryClass={cs.CV},
      url={https://arxiv.org/abs/1911.10150}, 
}

@InProceedings{Sun_2017_ICCV,
author = {Sun, Chen and Shrivastava, Abhinav and Singh, Saurabh and Gupta, Abhinav},
title = {Revisiting Unreasonable Effectiveness of Data in Deep Learning Era},
booktitle = {Proceedings of the IEEE International Conference on Computer Vision (ICCV)},
month = {Oct},
year = {2017}
}

@ARTICLE{6335478,
  author={Mogelmose, Andreas and Trivedi, Mohan Manubhai and Moeslund, Thomas B.},
  journal={IEEE Transactions on Intelligent Transportation Systems}, 
  title={Vision-Based Traffic Sign Detection and Analysis for Intelligent Driver Assistance Systems: Perspectives and Survey}, 
  year={2012},
  volume={13},
  number={4},
  pages={1484-1497},
  doi={10.1109/TITS.2012.2209421}
}

@misc{cao2023observationcentricsortrethinkingsort,
      title={Observation-Centric SORT: Rethinking SORT for Robust Multi-Object Tracking}, 
      author={Jinkun Cao and Jiangmiao Pang and Xinshuo Weng and Rawal Khirodkar and Kris Kitani},
      year={2023},
      eprint={2203.14360},
      archivePrefix={arXiv},
      primaryClass={cs.CV},
      url={https://arxiv.org/abs/2203.14360}, 
}

@INPROCEEDINGS{7533003,
  author={Bewley, Alex and Ge, Zongyuan and Ott, Lionel and Ramos, Fabio and Upcroft, Ben},
  booktitle={2016 IEEE International Conference on Image Processing (ICIP)}, 
  title={Simple online and realtime tracking}, 
  year={2016},
  volume={},
  number={},
  pages={3464-3468},
  doi={10.1109/ICIP.2016.7533003}
}

@INPROCEEDINGS{8296962,
  author={Wojke, Nicolai and Bewley, Alex and Paulus, Dietrich},
  booktitle={2017 IEEE International Conference on Image Processing (ICIP)}, 
  title={Simple online and realtime tracking with a deep association metric}, 
  year={2017},
  volume={},
  number={},
  pages={3645-3649},
  doi={10.1109/ICIP.2017.8296962}
}

@article{stanojevic2024boosttrack,
  title={BoostTrack: Boosting the similarity measure and detection confidence for improved multiple object tracking},
  author={Stanojevic, Vukasin D and Todorovic, Branimir T},
  journal={Machine Vision and Applications},
  volume={35},
  number={3},
  pages={53},
  year={2024},
  publisher={Springer}
}

@article{stanojevic2024boosttrack++,
  title={Boosttrack++: using tracklet information to detect more objects in multiple object tracking},
  author={Stanojevi{\'c}, Vuka{\v{s}}in and Todorovi{\'c}, Branimir},
  journal={arXiv preprint arXiv:2408.13003},
  year={2024}
}

@article{DBLP:journals/corr/abs-2110-06864,
  author       = {Yifu Zhang and
                  Peize Sun and
                  Yi Jiang and
                  Dongdong Yu and
                  Zehuan Yuan and
                  Ping Luo and
                  Wenyu Liu and
                  Xinggang Wang},
  title        = {ByteTrack: Multi-Object Tracking by Associating Every Detection Box},
  journal      = {CoRR},
  volume       = {abs/2110.06864},
  year         = {2021},
  url          = {https://arxiv.org/abs/2110.06864},
  eprinttype   = {arXiv},
  eprint       = {2110.06864},
  bibsource    = {dblp computer science bibliography, https://dblp.org}
}

@misc{du2023strongsortmakedeepsortgreat,
      title={StrongSORT: Make DeepSORT Great Again}, 
      author={Yunhao Du and Zhicheng Zhao and Yang Song and Yanyun Zhao and Fei Su and Tao Gong and Hongying Meng},
      year={2023},
      eprint={2202.13514},
      archivePrefix={arXiv},
      primaryClass={cs.CV},
      url={https://arxiv.org/abs/2202.13514}, 
}

@misc{lin2015microsoftcococommonobjects,
      title={Microsoft COCO: Common Objects in Context}, 
      author={Tsung-Yi Lin and Michael Maire and Serge Belongie and Lubomir Bourdev and Ross Girshick and James Hays and Pietro Perona and Deva Ramanan and C. Lawrence Zitnick and Piotr Dollár},
      year={2015},
      eprint={1405.0312},
      archivePrefix={arXiv},
      primaryClass={cs.CV},
      url={https://arxiv.org/abs/1405.0312}, 
}

@misc{liu2021swintransformerhierarchicalvision,
      title={Swin Transformer: Hierarchical Vision Transformer using Shifted Windows}, 
      author={Ze Liu and Yutong Lin and Yue Cao and Han Hu and Yixuan Wei and Zheng Zhang and Stephen Lin and Baining Guo},
      year={2021},
      eprint={2103.14030},
      archivePrefix={arXiv},
      primaryClass={cs.CV},
      url={https://arxiv.org/abs/2103.14030}, 
}

@misc{zong2023detrscollaborativehybridassignments,
      title={DETRs with Collaborative Hybrid Assignments Training}, 
      author={Zhuofan Zong and Guanglu Song and Yu Liu},
      year={2023},
      eprint={2211.12860},
      archivePrefix={arXiv},
      primaryClass={cs.CV},
      url={https://arxiv.org/abs/2211.12860}, 
}

@Article{liu2022convnet,
  author  = {Zhuang Liu and Hanzi Mao and Chao-Yuan Wu and Christoph Feichtenhofer and Trevor Darrell and Saining Xie},
  title   = {A ConvNet for the 2020s},
  journal = {arXiv preprint arXiv:2201.03545},
  year    = {2022},
}

@misc{oquab2024dinov2learningrobustvisual,
      title={DINOv2: Learning Robust Visual Features without Supervision}, 
      author={Maxime Oquab and Timothée Darcet and Théo Moutakanni and Huy Vo and Marc Szafraniec and Vasil Khalidov and Pierre Fernandez and Daniel Haziza and Francisco Massa and Alaaeldin El-Nouby and Mahmoud Assran and Nicolas Ballas and Wojciech Galuba and Russell Howes and Po-Yao Huang and Shang-Wen Li and Ishan Misra and Michael Rabbat and Vasu Sharma and Gabriel Synnaeve and Hu Xu and Hervé Jegou and Julien Mairal and Patrick Labatut and Armand Joulin and Piotr Bojanowski},
      year={2024},
      eprint={2304.07193},
      archivePrefix={arXiv},
      primaryClass={cs.CV},
      url={https://arxiv.org/abs/2304.07193}, 
}

@misc{hu2021loralowrankadaptationlarge,
      title={LoRA: Low-Rank Adaptation of Large Language Models}, 
      author={Edward J. Hu and Yelong Shen and Phillip Wallis and Zeyuan Allen-Zhu and Yuanzhi Li and Shean Wang and Lu Wang and Weizhu Chen},
      year={2021},
      eprint={2106.09685},
      archivePrefix={arXiv},
      primaryClass={cs.CL},
      url={https://arxiv.org/abs/2106.09685}, 
}

@misc{liu2024bevfusionmultitaskmultisensorfusion,
      title={BEVFusion: Multi-Task Multi-Sensor Fusion with Unified Bird's-Eye View Representation}, 
      author={Zhijian Liu and Haotian Tang and Alexander Amini and Xinyu Yang and Huizi Mao and Daniela Rus and Song Han},
      year={2024},
      eprint={2205.13542},
      archivePrefix={arXiv},
      primaryClass={cs.CV},
      url={https://arxiv.org/abs/2205.13542}, 
}

@misc{bai2022transfusionrobustlidarcamerafusion,
      title={TransFusion: Robust LiDAR-Camera Fusion for 3D Object Detection with Transformers}, 
      author={Xuyang Bai and Zeyu Hu and Xinge Zhu and Qingqiu Huang and Yilun Chen and Hongbo Fu and Chiew-Lan Tai},
      year={2022},
      eprint={2203.11496},
      archivePrefix={arXiv},
      primaryClass={cs.CV},
      url={https://arxiv.org/abs/2203.11496}, 
}

@misc{huang2023detectinglabelingrethinkinglidarcamera,
      title={Detecting As Labeling: Rethinking LiDAR-camera Fusion in 3D Object Detection}, 
      author={Junjie Huang and Yun Ye and Zhujin Liang and Yi Shan and Dalong Du},
      year={2023},
      eprint={2311.07152},
      archivePrefix={arXiv},
      primaryClass={cs.CV},
      url={https://arxiv.org/abs/2311.07152}, 
}

@misc{he2015deepresiduallearningimage,
      title={Deep Residual Learning for Image Recognition}, 
      author={Kaiming He and Xiangyu Zhang and Shaoqing Ren and Jian Sun},
      year={2015},
      eprint={1512.03385},
      archivePrefix={arXiv},
      primaryClass={cs.CV},
      url={https://arxiv.org/abs/1512.03385}, 
}

@misc{chen2023largekernel3dscalingkernels3d,
      title={LargeKernel3D: Scaling up Kernels in 3D Sparse CNNs}, 
      author={Yukang Chen and Jianhui Liu and Xiangyu Zhang and Xiaojuan Qi and Jiaya Jia},
      year={2023},
      eprint={2206.10555},
      archivePrefix={arXiv},
      primaryClass={cs.CV},
      url={https://arxiv.org/abs/2206.10555}, 
}

@misc{yin2021centerbased3dobjectdetection,
      title={Center-based 3D Object Detection and Tracking}, 
      author={Tianwei Yin and Xingyi Zhou and Philipp Krähenbühl},
      year={2021},
      eprint={2006.11275},
      archivePrefix={arXiv},
      primaryClass={cs.CV},
      url={https://arxiv.org/abs/2006.11275}, 
}

@misc{dosovitskiy2021imageworth16x16words,
      title={An Image is Worth 16x16 Words: Transformers for Image Recognition at Scale}, 
      author={Alexey Dosovitskiy and Lucas Beyer and Alexander Kolesnikov and Dirk Weissenborn and Xiaohua Zhai and Thomas Unterthiner and Mostafa Dehghani and Matthias Minderer and Georg Heigold and Sylvain Gelly and Jakob Uszkoreit and Neil Houlsby},
      year={2021},
      eprint={2010.11929},
      archivePrefix={arXiv},
      primaryClass={cs.CV},
      url={https://arxiv.org/abs/2010.11929}, 
}

@misc{loshchilov2019decoupledweightdecayregularization,
      title={Decoupled Weight Decay Regularization}, 
      author={Ilya Loshchilov and Frank Hutter},
      year={2019},
      eprint={1711.05101},
      archivePrefix={arXiv},
      primaryClass={cs.LG},
      url={https://arxiv.org/abs/1711.05101}, 
}
